\documentclass[letterpaper]{article} 
\usepackage{aaai2026}  
\usepackage{times}  
\usepackage{helvet}  
\usepackage{courier}  
\usepackage[hyphens]{url}  
\usepackage{graphicx} 
\usepackage{natbib}  
\usepackage{caption} 
\usepackage{algorithm}
\usepackage[noend]{algpseudocode}
\algnotext{EndIf}  
\usepackage{xcolor}
\usepackage{amsmath}
\usepackage{amsfonts}
\usepackage[justification=centering]{subfig}  

\usepackage{tabularx}
\usepackage{appendix}
\usepackage{booktabs}
\usepackage{multirow}
\usepackage{makecell}

\usepackage{listings}
\definecolor{codegreen}{rgb}{0,0.6,0}
\definecolor{codegray}{rgb}{0.5,0.5,0.5}
\definecolor{codeblue}{rgb}{0.0,0.0,0.6}
\definecolor{backcolour}{rgb}{0.95,0.95,0.95}
\lstdefinestyle{mystyle}{
    backgroundcolor=\color{backcolour},
    commentstyle=\color{codegreen},
    keywordstyle=\color{codeblue},
    numberstyle=\tiny\color{codegray},
    stringstyle=\color{codegreen},
    basicstyle=\ttfamily\footnotesize,
    breakatwhitespace=false,
    breaklines=true,
    captionpos=t,
    keepspaces=true,
    numbers=left,
    numbersep=5pt,
    showspaces=false,
    showstringspaces=false,
    showtabs=false,
    tabsize=2,
    frame=single, 
    rulecolor=\color{black},
    title=\lstname
}

\usepackage{newfloat}
\usepackage{listings}
\DeclareCaptionStyle{ruled}{labelfont=normalfont,labelsep=colon,strut=off} 
\floatstyle{ruled}
\newfloat{listing}{tb}{lst}{}
\floatname{listing}{Listing}
\usepackage{etoolbox}

\makeatletter
\apptocmd{\@maketitle}{%
  \begin{center}
    \includegraphics[width=0.95\textwidth]{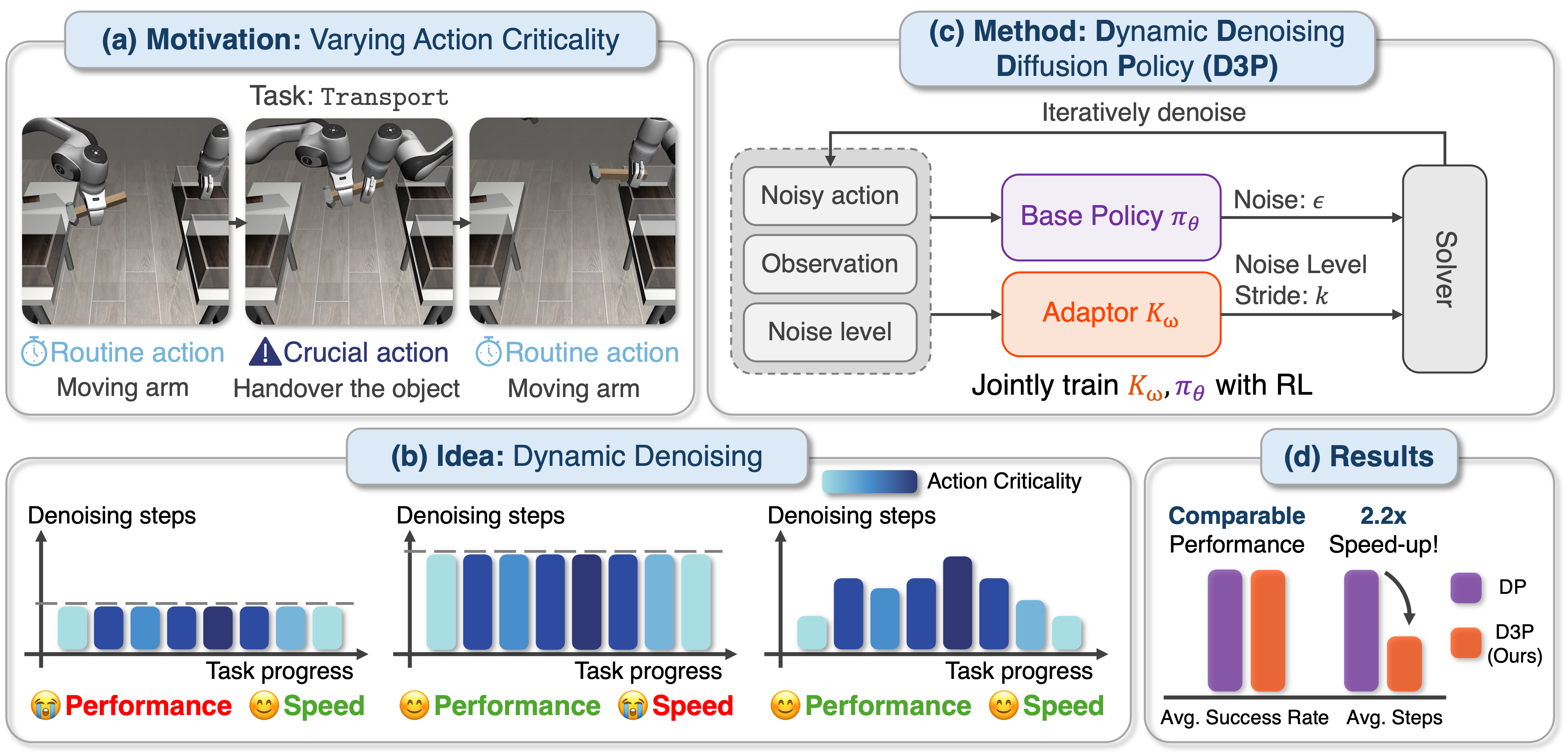}
    \captionof{figure}{An overview of \textbf{D}ynamic \textbf{D}enoising \textbf{D}iffusion \textbf{P}olicy (D3P). \textbf{(a) Motivation:} Robotic tasks involve actions of varying criticality. Crucial actions, like object handover, have a greater impact on task success than routine actions. \textbf{(b) Idea:} Instead of using a fixed number of denoising steps, D3P dynamically allocates more denoising steps to crucial actions. \textbf{(c) Method:} D3P uses a base policy $\pi_\theta$ and a lightweight adaptor $K_\omega$. The adaptor predicts the noise-level strides, which determines total denoising steps for current action. \textbf{(d) Results:} D3P achieves a comparable success rate, while providing a 2.2$\times$ speed-up to a normal fixed-step diffusion policy.}
    
    \label{fig:overview}
  \end{center}
  \vspace{1mm}
}{}{\PackageError{aaai2026}{Could not patch @maketitle}{}}
\makeatother

\usepackage{cleveref}
\Crefname{figure}{Fig.}{Figs.}
\Crefname{equation}{Eq.}{Eqs.}
\Crefname{section}{Sec.}{Secs.}
\Crefname{algorithm}{Alg.}{Algs.}
\Crefname{table}{Tab.}{Tabs.}

\title{D3P: Dynamic Denoising Diffusion Policy via Reinforcement Learning}
\author {
    Shu'ang Yu\textsuperscript{\rm 12},
    Feng Gao\textsuperscript{\rm 1},
    Yi Wu\textsuperscript{\rm 13},
    Chao Yu\textsuperscript{\rm 1$\dagger$},
    Yu Wang\textsuperscript{\rm 1$\dagger$},
}
\affiliations {
    \textsuperscript{\rm 1}Tsinghua University\quad
    \textsuperscript{\rm 2}Shanghai AI Laboratory\quad
    \textsuperscript{\rm 3}Shanghai Qi Zhi Institute \\
    \textsuperscript{\rm $\dagger$} Corresponding Authors \\
    \texttt{\{yuchao, yu-wang\}@mail.tsinghua.edu.cn}
}

\begin{document}

\maketitle

\begin{abstract}
Diffusion policies excel at learning complex action distributions for robotic visuomotor tasks, yet their iterative denoising process poses a major bottleneck for real-time deployment. Existing acceleration methods apply a fixed number of denoising steps per action, implicitly treating all actions as equally important.
However, our experiments reveal that robotic tasks often contain a mix of \emph{crucial} and \emph{routine} actions, which differ in their impact on task success. Motivated by this finding, we propose \textbf{D}ynamic \textbf{D}enoising \textbf{D}iffusion \textbf{P}olicy \textbf{(D3P)}, a diffusion-based policy that adaptively allocates denoising steps across actions at test time.
D3P uses a lightweight, state-aware adaptor to allocate the optimal number of denoising steps for each action. We jointly optimize the adaptor and base diffusion policy via reinforcement learning to balance task performance and inference efficiency.
On simulated tasks, D3P achieves an averaged 2.2$\times$ inference speed-up over baselines without degrading success. Furthermore, we demonstrate D3P's effectiveness on a physical robot, achieving a 1.9$\times$ acceleration over the baseline.
\end{abstract}
\section{Introduction}
\label{sec:intro}
Diffusion policies have demonstrated remarkable promise in robotic visuomotor tasks~\cite{2022jannerdiffuser, pearce2023imitatinghumanbehaviourdiffusion, chi2023diffusion, ze20243d, ma2024hierarchical}. 
By casting action generation as a conditional denoising diffusion process, they naturally model the full, often highly multimodal distribution of feasible actions while retaining stable optimization dynamics \cite{chi2023diffusion}.
Concretely, action sampling integrates the reverse-time stochastic differential equation (SDE) that refines Gaussian noise into clean actions, following \citet{ho2020denoising} and \citet{song2020score}.
This procedure entails dozens of denoising steps, so diffusion policies typically run more slowly at inference than one-shot generators based on GANs~\cite{goodfellow2020generative}, VAEs~\cite{kingma2013auto}, or autoregressive models~\cite{shafiullah2022behavior, zhao2023learning}, which limits their deployment in real-time control.

To address these challenges, previous work has sought to accelerate inference through various strategies, such as reformulating the process as an ordinary differential equation (ODE) to allow fewer sampling steps~\cite{song2020denoising, lu2022dpm, lu2022dpmp}, distilling the policy into a single-step model~\cite{prasad2024consistency, wang2024one}, or employing streaming techniques that use action history as a prior~\cite{hoeg2024streaming, chen2025falcon}. These acceleration methods for diffusion policies typically use a uniform number of denoising steps per action, which implicitly assumes that all actions are equally important for task success.

However, our analysis of robotic tasks provides evidence that contradicts this assumption. 
We observe a stark non-uniformity in action criticality: task executions typically consist of pivotal \textbf{crucial actions} that largely determine success, and more forgiving \textbf{routine actions} with only a marginal impact.
For instance, in a \texttt{Transport} task, the precise moment of object handover is crucial, whereas the arm movements before and after have less impact on task success. Treating all actions uniformly, therefore, incurs unnecessary computational overhead and restricts the ability to balance decision quality with inference efficiency.
This insight motivates a new design principle: \emph{Allocate denoising steps adaptively: spend more computation on crucial actions, and less on routine ones—to balance precision and efficiency.}

In this paper, we propose \textbf{D}ynamic \textbf{D}enoising \textbf{D}iffusion \textbf{P}olicy (D3P), a diffusion policy that dynamically adjusts the number of denoising steps during task execution. The D3P architecture consists of two components: a standard noise predicting network that serves as the base diffusion policy, and a lightweight adaptor (\Cref{fig:overview}). The adaptor is designed to predict the \emph{noise-level strides} based on the current observation, which eventually adapts the number of denoising steps for the current action.
We mathematically formulate the dynamic denoising problem as a two-layer partially observable Markov decision process (POMDP). Then we use reinforcement learning (RL) with this two-layer POMDP to jointly train both the base diffusion policy and the adaptor. 
Specifically, a base policy is fine-tuned with DPPO~\cite{ren2024dppo} to maximize task success. Concurrently, the lightweight adaptor is trained from scratch with PPO~\cite{schulman2017proximal}, where its reward function incentivizes minimizing denoising steps without compromising performance. A key challenge in this joint training process is to balance the contradictory objectives of task performance and inference efficiency. To address this, we introduce a three-stage training strategy that ensures stable convergence and robust performance.

We evaluate D3P on a range of simulated manipulation tasks. Using the same amount of training data, D3P achieves an average 2.2$\times$ inference speed-up over baseline methods while maintaining comparable performance. These results demonstrate that D3P effectively addresses the trade-off between performance and efficiency. Furthermore, we deploy D3P on the physical robot. It achieves a 1.9$\times$ acceleration in inference speed against a normal fixed-step diffusion policy. Overall, our main contributions are as follows:

\begin{enumerate}
\item Through an empirical study, we reveal that different actions contribute unequally to manipulation tasks. There exist some crucial actions significantly influencing task completion.
\item Building on this observation, we propose \textbf{D}ynamic \textbf{D}enoising \textbf{D}iffusion \textbf{P}olicy (\textbf{D3P}). Trained via RL, D3P features a noise-predicting network as the base diffusion policy and an adaptor to dynamically adjust its denoising steps during task execution.
\item We conduct experiments in eight simulated manipulation tasks, demonstrating that D3P achieves the best performance with an averaged 2.2$\times$ speed-up over baselines.
\item We demonstrate that D3P can be successfully deployed on a physical robot. It achieves a 1.9$\times$ acceleration against a standard fixed-step diffusion policy.
\end{enumerate}
\section{Preliminaries}
\label{sec:preliminary}
\begin{figure*}
  \centering
    \includegraphics[width=1.00\textwidth]{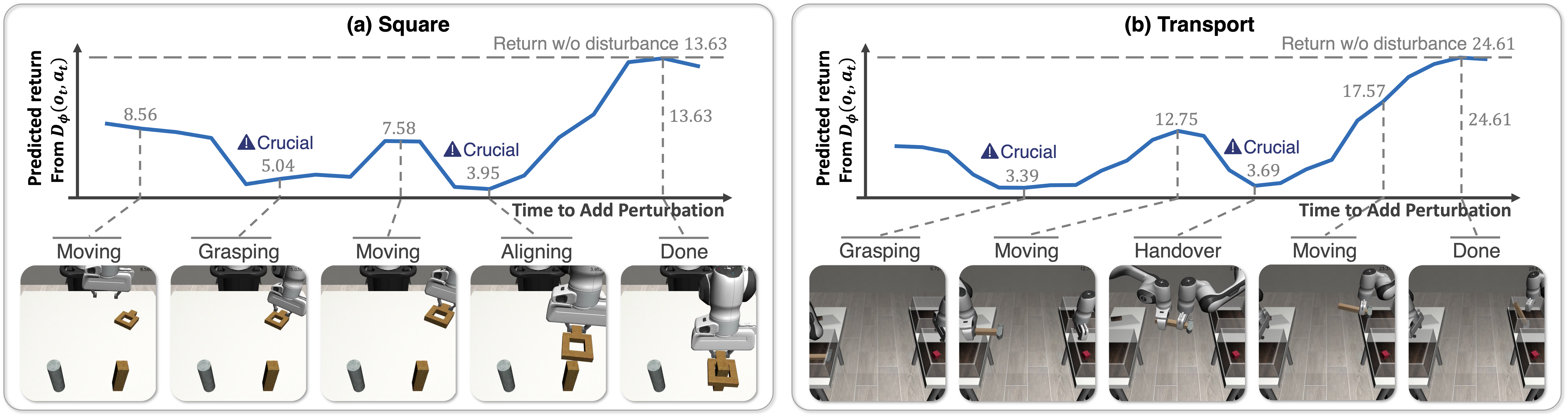}
  \caption{\textbf{Visualizing Action Criticality via Perturbed Returns.} The plots show the predicted perturbed return from $D_\phi(o_t, a_t)$ at different time of the task. A lower return indicates the action more crucial, as a perturbation is more likely to lead to task failure.}
  \label{fig:empirical_study}
\end{figure*}

\subsubsection{Partially Observable Markov Decision Process (POMDP)}
We formulate the robot environment within the framework of a partially observable Markov decision Process (POMDP), defined by the tuple $\mathcal{M}_\text{ENV}:=\left(\mathcal{S}_\text{ENV}, \mathcal{A}_\text{ENV}, \mathcal{O}_\text{ENV}, P_{\text{init}, \text{ENV}}, P_\text{ENV}, R_\text{ENV}\right)$. Here, $\mathcal{S}_\text{ENV}$, $\mathcal{A}_\text{ENV}$, and $\mathcal{O}_\text{ENV}$ are the state, action, and observation space separately. The process begins with an initial state $s_0 \sim P_{\text{init}, \text{ENV}}$. At each environment timestep $t$, an agent receives an observation $o_t \in \mathcal{O}_\text{ENV}$, and takes an action $a_t \in \mathcal{A}_\text{ENV}$ according to a policy $\pi(a_t\mid o_t)$. The environment responds by transitioning to a new state $s_{t+1} \sim P_\text{ENV}(s_{t+1}\mid s_t, a_t)$ and providing a reward $r_t = R_\text{ENV}(s_t, a_t)$. The objective in RL is to learn a policy $\pi_\theta$ that maximizes the expected return, $\mathbb{E}_{\pi_\theta}\left[J_t(s_t, a_t)\right] := \mathbb{E}_{\pi_\theta}\left[\sum_{\tau=t}^{T-1} \gamma_\text{ENV}^{\tau-t} r_\tau \mid s_t, a_t\right]$, where $\gamma_\text{ENV} \in (0, 1)$ is a discount factor and $T$ is the the episode horizon.

\subsubsection{Diffusion Models}
Denoising diffusion probabilistic models (DDPMs)~\cite{sohl2015deep, ho2020denoising} frame sample generation as an iterative denoising procedure, reversing a length-$N$
diffusion chain $\{x^{i}\}_{i=0}^{N}$, where $x^{0}$ is a clean sample,
$x^{N} \sim \mathcal{N}(\mathbf 0,\mathbf I)$ is pure Gaussian noise, and $N$ is the number of denoising steps.
The denoising is parameterized by a neural network, $\epsilon_\theta(x^i, i)$, trained to predict the noise component within a noisy sample $x^i$ at noise-level $i$. At inference, this process starts with a sample of pure noise $x^N$ and progressively refines it over $N$ steps, generating a clean sample $x^0$.

Despite great performance, inference in DDPMs is slow because it must execute all $N$ reverse steps.  
Denoising diffusion implicit models (DDIMs)~\cite{song2020denoising} accelerate sampling by replacing the stochastic reverse process with a deterministic mapping that allows larger strides in the noise schedule.
Using the same training loss as DDPMs, DDIMs traverse only a
sparse set of noise levels $\tau_0 > \tau_1 > \dots > \tau_S=0$ where $S \ll N$.
At a given level $i$, the sampler can skip $k>1$ noise-levels and go directly to a
less–noisy point $x^{i-k}$:
\begin{equation}
    \begin{aligned}
        x^{i-k} \sim 
            \mathcal{N}\!\bigl(\mu(x^i, \epsilon_i, i, k),\;
            \eta\,\sigma_i^{2}\mathbf I\bigr),
    \end{aligned}
    \label{eq:ddim}
\end{equation}  
where $\sigma_i$ comes from the predefined schedule.
Setting $\eta=0$ removes the stochastic term, making the process fully deterministic and much faster for inference

\subsubsection{Diffusion Policies}
Diffusion policy (DP)~\cite{chi2023diffusion} treats a diffusion model as the
control policy $\pi_\theta$.  
To preserve temporal consistency, DP predicts an action chunk $X_t = \{a_t,\dots,a_{t+T_a-1}\}$ conditioned on the current observation $o_t$. With the DDIM sampler, the denoising update at noise level $i$ is \looseness=-1
\begin{equation}
    X_t^{\,i-1}\sim
\mathcal{N}\left(
        \mu\left(X_t^{\,i},
                   \epsilon_\theta(X_t^{\,i},o_t,i),
                   i,k\right),\;
        \eta\,\sigma_i^{2}\mathbf I\right),
    \label{eq:dp}
\end{equation}
with $k$ the stride. 
This iterative procedure forms a POMDP
$\mathcal{M}_{\text{DN}}
    :=(\mathcal{S}_{\text{DN}},
        \mathcal{A}_{\text{DN}},
        \mathcal{O}_{\text{DN}},
        P_{\text{init},\text{DN}},
        P_{\text{DN}},
        R_{\text{DN}})$.
This process starts with a state of pure noise $X_t^N \sim \mathcal{N}(\mathbf{0}, \mathbf{I})$ and terminates at $i=0$ to produce the final action chunk.

\section{Empirical Study: Identifying Crucial Actions}
\label{sec:empirical}
We conduct an empirical study to validate that not all actions in a task are equally important. An action is deemed crucial if a disturbance on it significantly degrades task performance. Otherwise, it is considered routine.

\subsubsection{Experiment Design}
We start from a pre-trained expert policy $\pi_{\text{expert}}$ that achieves over 90\% success in the environment $\mathcal{M}_\text{ENV}$. At a randomly selected timestep $t\in\{0,\dots,T-1\}$ we add Gaussian noise $\xi$ to the expert action $a_t$, yielding a perturbed action $a'_t = a_t + \xi$. The episode then resumes under $\pi_{\text{expert}}$. The perturbation’s impact is measured by the episode return, $J(s_t,a'_t)=\mathbb{E}_{\pi_{\text{expert}}}\!\left[\sum_{\tau=0}^{T}\gamma_{\text{ENV}}^{\tau-t} r_\tau \bigm| s_t,a'_t\right].$ A low $J$ signals that the original $a_t$ was crucial.
We fit a lightweight predictor $D_\phi:\mathcal{O}_\text{ENV} \times \mathcal{A}_\text{ENV} \to \mathbb{R}$ that predicts $J$ from the unperturbed pair $(o_t,a_t)$.
More implementation details of our empirical study appear in \textbf{Appendix A}.

\begin{figure*}
    \centering
    \includegraphics[width=0.85\linewidth]{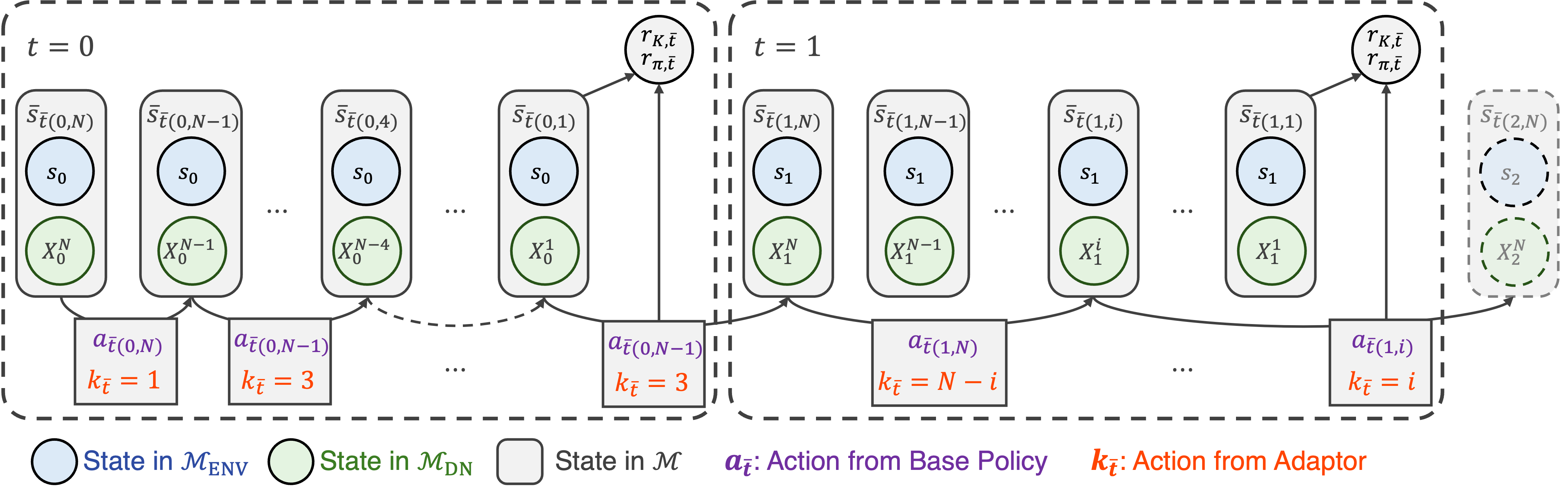}
    \caption{We formulate the dynamic denoising problem as a two-layer POMDP, where a denoising process ($\mathcal{M}_{\text{DN}}$) is nested within the environment ($\mathcal{M}_{\text{ENV}}$). At each step, the adaptor $K_{\omega}$ predicts the noise-level strides. The base diffusion policy and the adaptor are jointly updated via RL.}
    \label{fig:method}
\end{figure*}

\subsubsection{Key Findings}
\Cref{fig:empirical_study} plots $D_\phi$'s predicted return on the \texttt{Square} and \texttt{Transport} tasks from Robomimic~\cite{robomimic2021}. The results demonstrate that the criticality of actions throughout a task is highly non-uniform. For \texttt{Square}, fine interactions, such as grasping the block and aligning it with the peg, receive the lowest predicted returns (5.04, 3.95), marking them as crucial. Whereas broad arm motions score higher (7.58, 8.56) and are routine. After completion, the return peaks at 13.63 because later actions no longer influence success.
For \texttt{Transport}, grasping the hammer (3.39) and the bimanual hand-over (3.69) dominate task success, while transit motions are routine.
Overall, crucial actions are concentrated in direct physical interactions, such as grasping and handover, while broader movements have a relatively minor impact on the results. This observation motivates our approach: adaptively allocating denoising steps to different actions, focusing more capacity on the crucial actions.\looseness=-1

\section{Method}
In this section, we introduce \textbf{D}ynamic \textbf{D}enoising \textbf{D}iffusion \textbf{P}olicy (\textbf{D3P}), a method designed to dynamically adjust the number of denoising steps during task execution. D3P utilizes the noise-level stride scheme in DDIM solver~\cite{song2020denoising}, and augments a base diffusion policy $\pi_\theta$ with an adaptor $K_\omega$ trained to pick strides, as shown in \Cref{fig:overview}(c). The adaptor observes $\left(o_t,\ X_t^i\right)$ and outputs a stride: large strides skip more noise levels to speed up routine actions, while small strides keep precise denoising for crucial actions. We frame adaptive denoising as a two-layer POMDP and train the base policy $\pi_\theta$ and adaptor $K_\omega$ jointly with RL. During the training, $\pi_\theta$ is fine-tuned to maximize task reward, while $K_\omega$ learns to cut denoising steps without degrading success. A three-stage training strategy is used to stabilize training. The full algorithm is summarized in \Cref{alg:d3p}.

\subsection{Problem Formulation}
Following \citet{psenka2023learning,ren2024dppo}, we define a two-layer POMDP $\mathcal{M}=(\mathcal{S},\mathcal{A},\mathcal{O},P_{\text{init}},P,R)$ that nests the denoising process $\mathcal{M}_{\text{DN}}$ inside the environment $\mathcal{M}_{\text{ENV}}$ (\Cref{fig:method}). In this two-layer PODMP $\mathcal{M}$, we denote a time index as $\bar{t}(t,i)=tN + (N-i-1)$, where $t$ is the environment step and $i\in[0,N]$ is the index of noise level. The states and observations in $\mathcal{M}$ are all tuples, denoted as 
$
\bar{s}_{\bar{t}}=(s_t,X_t^{i}),
\bar{o}_{\bar{t}}=(o_t,X_t^{i}),
$
with $s_t\in\mathcal{S}_{\text{ENV}}$ and $X_t^{i}\in\mathcal{S}_{\text{DN}}$. 
An episode starts at $\bar{t}(0,N)$ with $\bar{s}_{\bar{t}(0, N)}\sim P_\text{init}$, and $P_\text{init}$ is defined as
\begin{equation}
P_{\text{init}} = \left(P_{\text{init},\text{ENV}},\  P_{\text{init},\text{DN}}\right) .
\label{eq:pf1}
\end{equation}
At each time step $\bar{t}$, the action tuple $(a_{\bar{t}}, k_{\bar{t}})$ is generated by the base policy $\pi_\theta$ and the adaptor $K_\omega$ following \Cref{eq:pf2}, where $\lfloor\cdot\rfloor$ indicates rounding down.
\begin{equation}
    \begin{aligned}
        & k_{\bar{t}(t, i)} \sim K_\omega(k_{\bar{t}}\mid \bar{o}_{\bar{t}}),\ j_{\bar{t}}=\max \left(i-\lfloor k_{\bar{t}(t, i)} \rfloor ,\ 0\right), \\
        & a_{\bar{t}} \sim \pi_\theta(a_{\bar{t}}\mid \bar{o}_{\bar{t}}, i)\!:=\!\mathcal{N}\!\left(\mu(X_{t}^i, \epsilon_\theta(\bar{o}_{\bar{t}}, i), i, k_{\bar{t}}), \eta\sigma_i^2\mathbf{I}\right).
    \end{aligned}
    \label{eq:pf2}
\end{equation}
\Cref{eq:pf2} indicates that $K_\omega$ predicts the noise-level strides $k_{\bar{t}}$ to further control the total denoising steps. The $\eta$ is set to $1$ during training, making the Gaussian likelihood of $\pi_\theta$ computable, and is set to $0$ when inference. After taking actions, $\mathcal{M}$ makes transition following \Cref{eq:pf3}.
\begin{equation}
    \bar{s}_{\bar{t}+1}\sim \left\{\begin{aligned}
        & \left( \delta_{s_t}, \delta_{a_{\bar{t}}} \right) &,\ j>0,\\
        & \left(P_\text{ENV}(s_{t+1}\mid s_t, X_t^0), P_{\text{init}, \text{DN}} \right)&,\ j=0.
    \end{aligned}\right.
    \label{eq:pf3}
\end{equation}
In \Cref{eq:pf3}, $j>0$ signifies an incomplete denoising process, and both the environment state $s_t$ and observastion $o_t$ remain unchanged. When $j=0$, the denoising process concludes and the resulting action is executed in the environment. Subsequently, $\mathcal{M}_{\text{ENV}}$ transitions to a new state according to $P_\text{ENV}$, and a new pure noise, $X_{t+1}^N$, is resampled from Gaussian distribution.

\begin{algorithm*}
    \caption{Dynamic Denoising Diffusion Policy (D3P)}
    \label{alg:d3p}
    \begin{algorithmic}[1]
        \State \textbf{Input:} Pretrained base policy $\pi_{\theta}$, dynamic denoising environment $\mathcal{M}$, discount factors $\gamma_{\text{ENV}}, \gamma_s$, environment horizon $T$, max denoise steps $N$, warm-up denoising steps $c$, update epochs $e, e_\text{slow}$, stage threshold $\zeta_1, \zeta_2$.
        
        \State \textbf{Initialize:} Adaptor $K_\omega$ with mean of $c$, empty buffer $\mathcal{B}$
        .
        \While{Average $r_{s, bar{t}} < \zeta_1$} \textcolor{blue}{\Comment{Stage 1: Warm-up}}
            \State Warm-up $\pi_\theta$ using DPPO~\cite{ren2024dppo} with fixed denoising steps $c$ 
        \EndWhile
        
        \While{not converged} \textcolor{blue}{\Comment{Stage 2: Joint training}}
            \State Reset environment to $\bar{t}(0, N)$ as \Cref{eq:pf1}
            \While{$t<T$} \textcolor{gray}{\Comment{Rollout data}}
                \State Sample action $(k_{\bar{t}}, \bar{X}_{\bar{t}})$ following \Cref{eq:pf2}. Get $\log K_{\bar{t}}, \log\pi_{\bar{t}}$
                \State Execute the actions. Perform transition as \Cref{eq:pf3}. Get $\bar{o}_{\bar{t}+1}$ and $r_{\pi, \bar{t}}$ following~\cite{ren2024dppo}.
                \State Add $\left(\bar{o}_{\bar{t}}, \left(k_{\bar{t}}, a_{\bar{t}}\right), \left(\log K_\omega (k_{\bar{t}}), \log \pi_\theta(a_{\bar{t}})\right), r_{\pi, \bar{t}} \right)$ to buffer $\mathcal{B}$.
            \EndWhile
            \State Calculate the advantage $\hat{A}_\Theta$ using \Cref{eq:adv}. Get success flag $r_{s, \bar{t}}$ refer to task results. \textcolor{gray}{\Comment{Prepare for update}}
            \State Set $r_{K, \bar{t}}$ as \Cref{eq:rew1}, using $\hat{A}_\Theta$ and $\text{r}_{s, \bar{t}}$.
    
            \For{$\text{epoch}=0, 1, \dots, e-1$} \textcolor{gray}{\Comment{Policy optimization}}
                \State Update parameter $\theta$ and $\Theta$ using DPPO~\cite{ren2024dppo}
                \State Update parameter $\omega$ with PPO loss in \Cref{eq:ppo0}
            \EndFor
    
            \If{Average $\text{stp} < \zeta_2$} 
            $e\leftarrow e_\text{slow}$ \textcolor{blue}{\Comment{Stage 3: Conservative fine-tuning}}
            \EndIf
        \EndWhile
        \State \textbf{Return:} Trained base policy $\pi_\theta$ and adaptor $K_\omega$.
    \end{algorithmic}
\end{algorithm*}

\subsection{Dynamic Denoising}
Following the problem formulation, we jointly train $\pi_\theta$ and $K_\omega$ on the same batch of rollout data.

To fine-tune the base policy, we apply DPPO~\cite{ren2024dppo} to optimize $\pi_\theta$ and its value critic $V_\Theta$ on the dynamic-denoising POMDP $\mathcal{M}$. The critic $V_\Theta(o_t)$, conditioned on the environment observation $o_t$, later serves as a proxy for task performance when training the adaptor.

The adaptor aims to reduce the total denoising steps without degrading performance. The most direct performance signal is the binary success flag $r_{s,\bar{t}}\in\{0,1\}$, yet its sparsity and high variance impede stable learning. Instead, we employ the advantage of each state–action pair in $\mathcal{M}_{\text{ENV}}$ as a dense, low-variance, but slightly biased metric, computed as
\begin{equation}
\hat{A}_{\Theta}(X_t^0,o_t)
= J_{\pi_\theta}(s_t,X_t^0)-V_\Theta(o_t),
\label{eq:adv}
\end{equation}
where $J_{\pi_\theta}(s_t,X_t^0)$ is the discounted return defined in the \textbf{Preliminaries}. A larger $\hat{A}_{\Theta}$ indicates that the fully denoised action $X_t^0$ is beneficial for future returns.

To encourage shorter denoising chains, we introduce a discount factor $\gamma_s\in(0,1)$ that penalizes longer denoising process. Balancing task performance against computational cost, we define the reward for adaptor as
\begin{equation}
    r_{K, \bar{t}(t, i)} = \left\{
    \begin{aligned}
        & 0 & ,\ j>0 \\
        &\begin{aligned}
        \alpha\ &\hat{A}_{\Theta}\left(X_t^0, o_t\right)\ \gamma_s^{\text{sgn}_t\times\text{stp}_t} \\
        +& \beta\ r_{s, \bar{t}}\ \gamma_s^{\text{stp}_t}
        \end{aligned} & ,\ j=0
    \end{aligned}
    \right.
    \label{eq:rew1}
\end{equation}
where $\alpha$ and $\beta$ are weighting coefficients, $\text{sgn}_t$ is the sign of the advantage $\hat{A}_{\Theta}(X_t^0,o_t)$, and $\text{stp}_t$ denotes the total denoising steps for generating the clean action $X_t^0$. This formulation rewards advantageous actions and successful episodes while exponentially penalizing long denoising sequences.

We update $K_\omega$ with PPO~\cite{schulman2017proximal}, minimizing the clipped PPO loss in \Cref{eq:ppo0}.
\begin{equation}\begin{aligned}
    - L^{\text{CLIP}}(\omega) = -\hat{\mathbb{E}}_{K_{\omega_\text{old}}}\left[\min \left( \rho(\omega) \hat{A}_{\omega_\text{old}}(k_{\bar{t}}, \bar{o}_{\bar{t}}),\right.\right. \\
    \left.\left.\text{clip}\left(\rho(\omega), 1-\epsilon_{\text{clip}}, 1+\epsilon_{\text{clip}}\right) \hat{A}_{\omega_\text{old}}(k_{\bar{t}}, \bar{o}_{\bar{t}}) \right) \right].
\end{aligned}\label{eq:ppo0}\end{equation}
Here, $\rho(\omega) = \frac{K_\omega(k_{\bar{t}} \mid \bar{o}_{\bar{t}})}{K_{\omega_\text{old}}(k_{\bar{t}} \mid \bar{o}_{\bar{t}})}$ is the probability ratio between the current and old adaptor. The advantage $\hat{A}_{\omega_{\text{old}}}$ is computed using Generalized Advantage Estimation (GAE)~\cite{schulman2015high} based on the reward $r_{K, \bar{t}}$.

\subsection{Training Strategy}
Directly training the base policy and adaptor from scratch often causes instability and even collapse. To address this, D3P adopts a three-stage training strategy preceded by behavior cloning of the base policy $\pi_\theta$ on pre-collected datasets.

\subsubsection{Stage 1: Base DP Warm-up.} 
We first warm up $\pi_\theta$ with DPPO~\cite{ren2024dppo} while keeping the denoising steps fixed at $c$. The warming up proceeds until the task success rate exceeds a preset threshold $\zeta_1$. This stage can be skipped if a sufficiently strong base policy is already available.

\subsubsection{Stage 2: Joint Training.} 
Next, we initialize the adaptor $K_\omega$ as a Gaussian policy with mean of $c$ and variance of $v^{2}$. Then we train $K_\omega$ and $\pi_\theta$ jointly. Each iteration collects a batch of trajectories followed by $e$ PPO update epochs for both modules.

\subsubsection{Stage 3: Conservative Fine-tuning.} 
When the average denoising steps satisfies $\mathbb{E}[\text{stp}_t]<\zeta_2$, we switch to a conservative stage. This stage mirrors Stage 2 but uses fewer update epochs per iteration. This precaution prevents the adaptor from shrinking the denoising steps so aggressively that it destabilizes the base policy and leads to a collapse.

The full algorithm of D3P, including the three-stage schedule, is summarized in \Cref{alg:d3p}.
\section{Experiments}
To evaluate D3P, we conduct comprehensive robot manipulation experiments in both simulation and the real world. We first detail the experimental setups, then benchmark D3P against several baselines. Subsequently, we present ablation studies to analyze the contribution of each component in our framework. Finally, we demonstrate the deployment of D3P onto a physical robot, highlighting its practical applicability.

\begin{figure*}[!tb]
  \centering 

  \subfloat[Lift (State)]{%
    \includegraphics[width=0.245\textwidth]{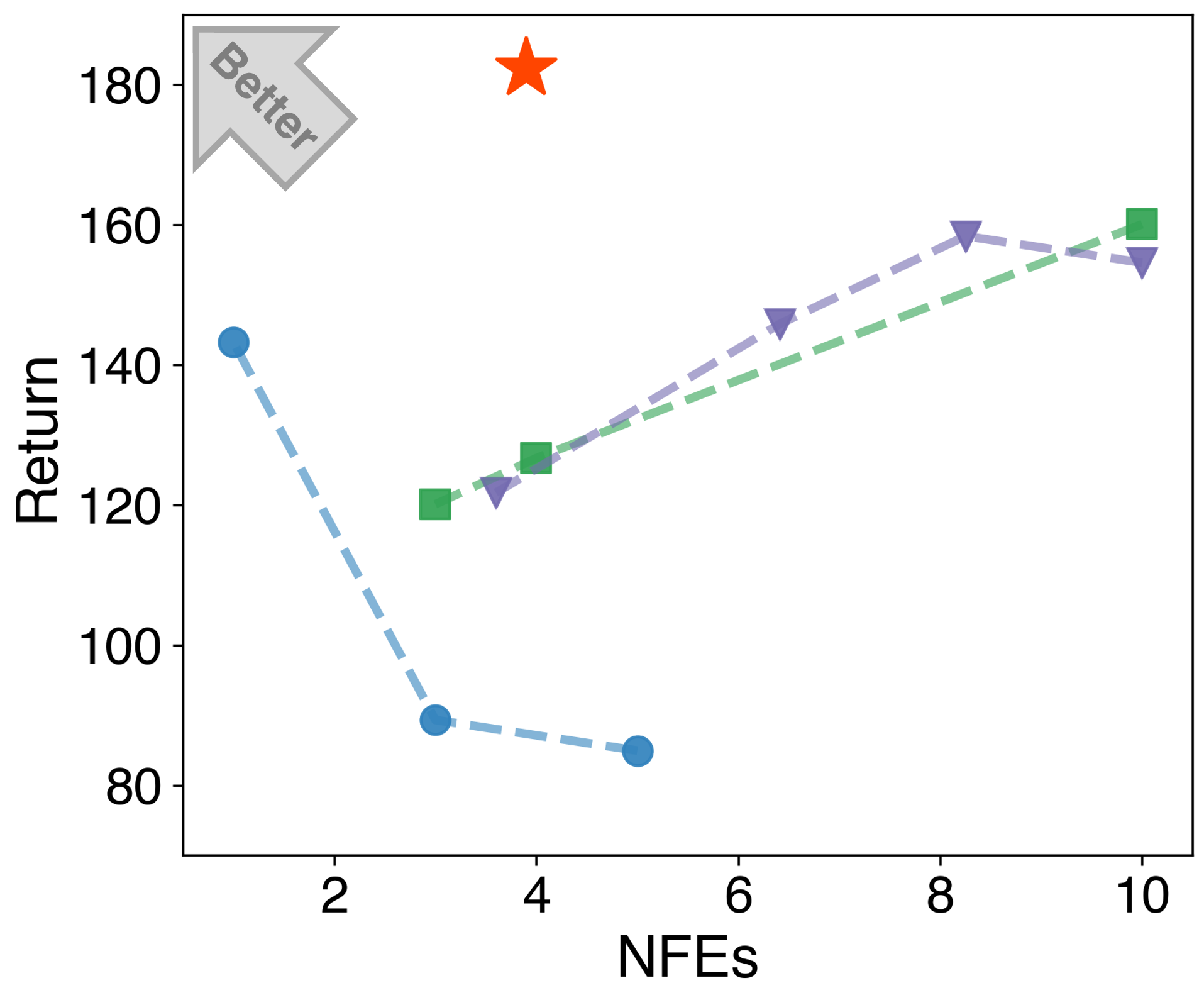}%
    \label{fig:lift} 
  }
  \subfloat[Can (State)]{%
    \includegraphics[width=0.245\textwidth]{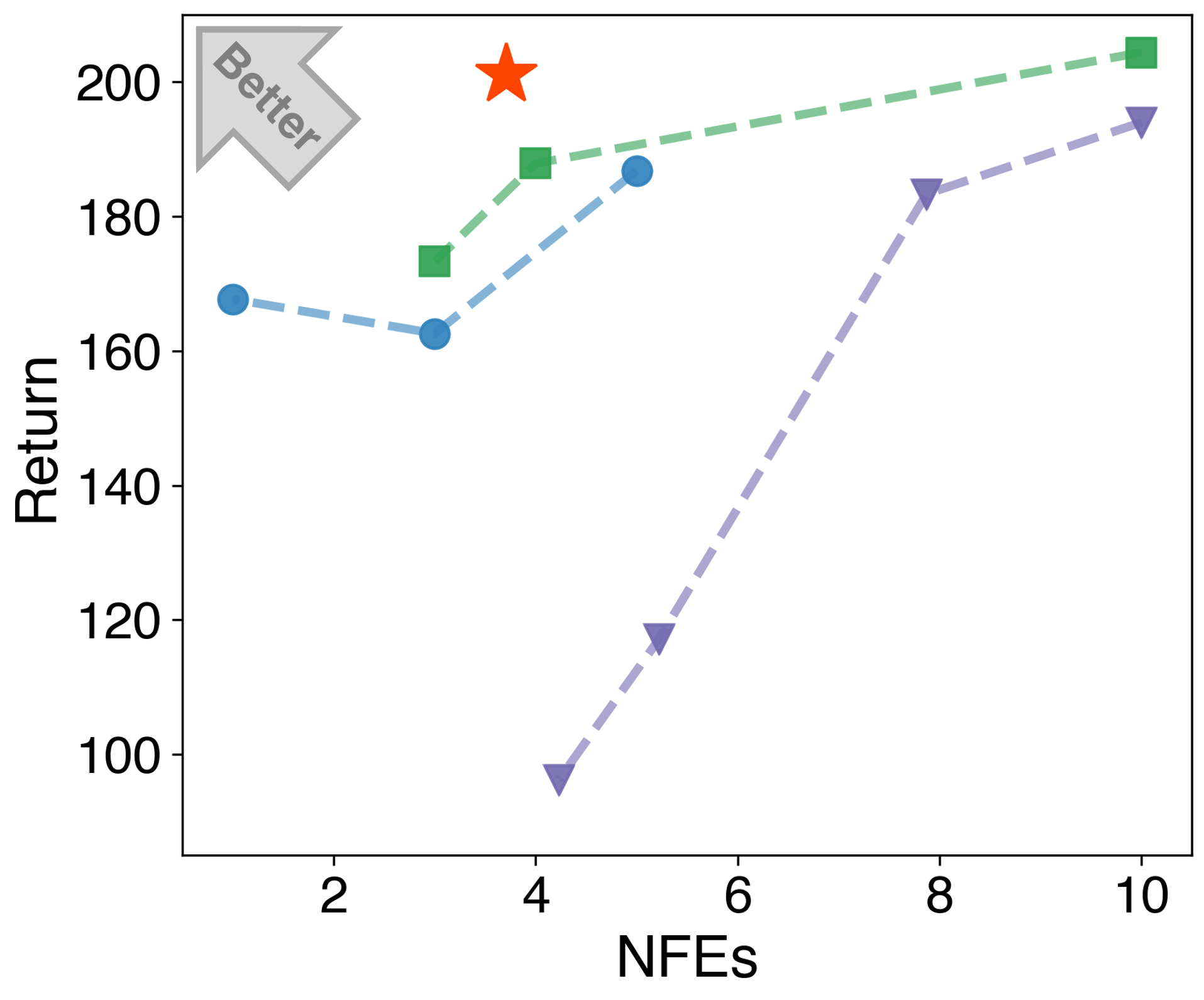}%
    \label{fig:can} 
  }
  \subfloat[Square (State)]{%
    \includegraphics[width=0.245\textwidth]{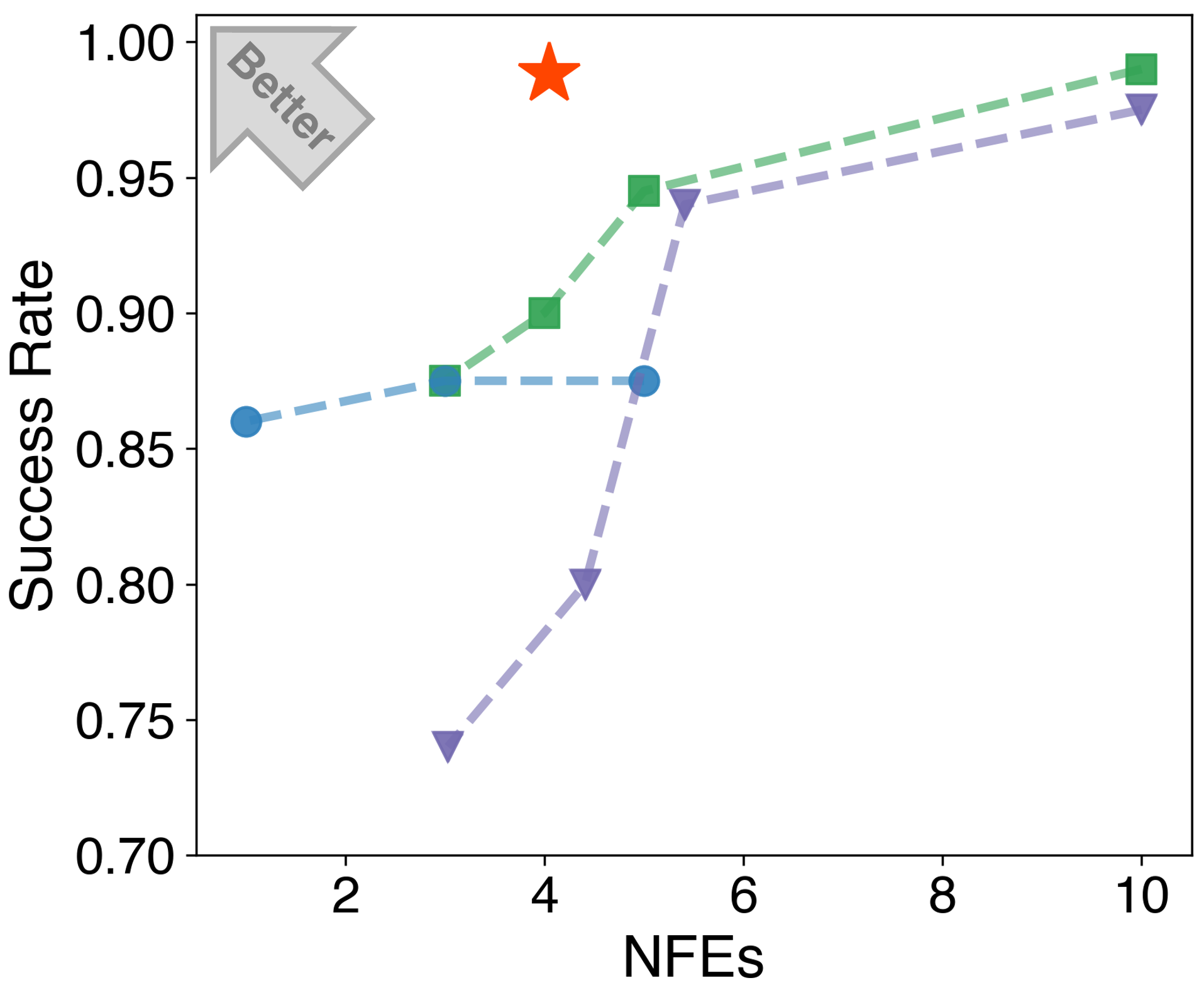}%
    \label{fig:square} 
  }
  \subfloat[Square (Pixel)]{%
    \includegraphics[width=0.245\textwidth]{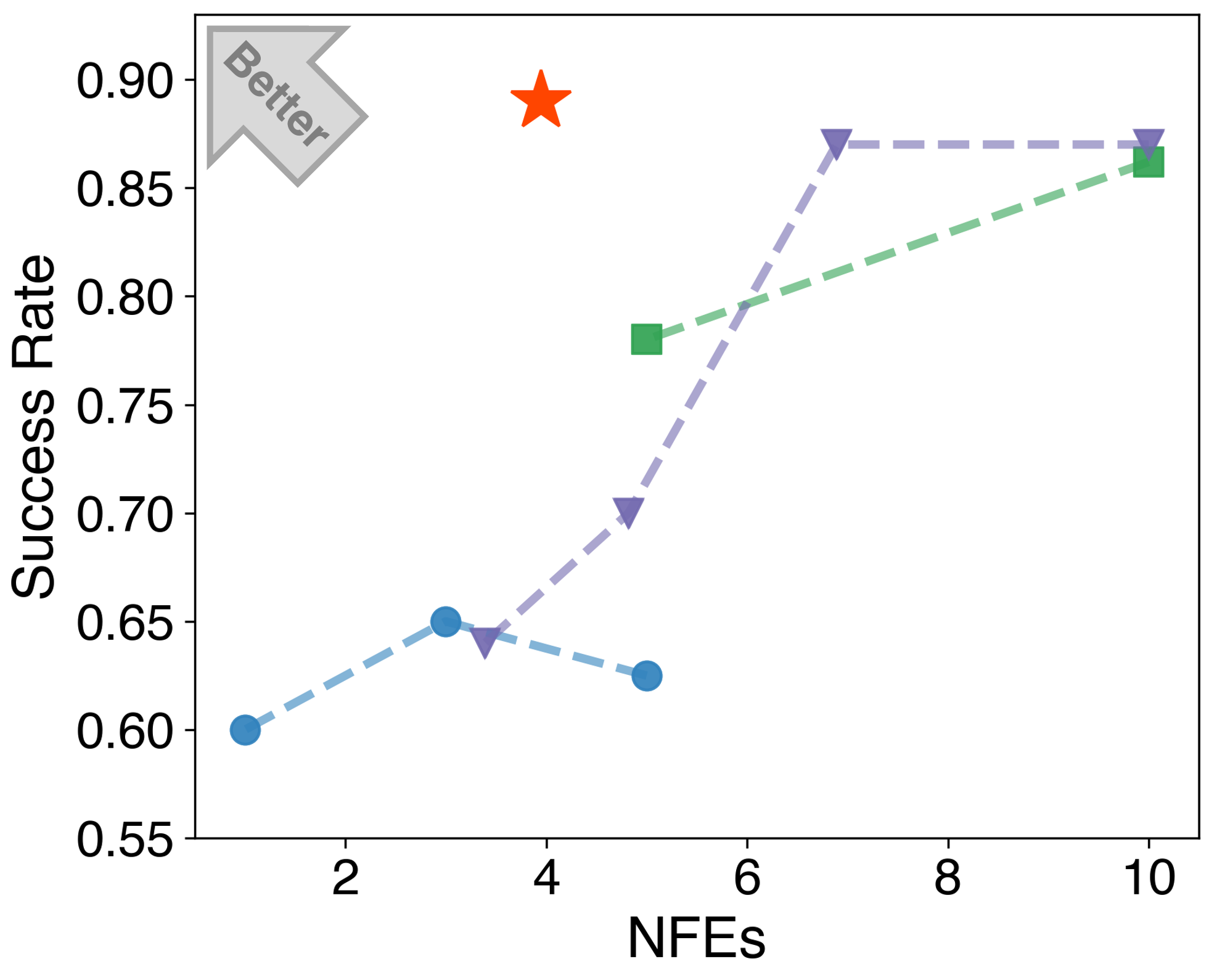}%
    \label{fig:square-img} 
  }
  \\
  \subfloat[Transport (State)]{%
    \includegraphics[width=0.245\textwidth]{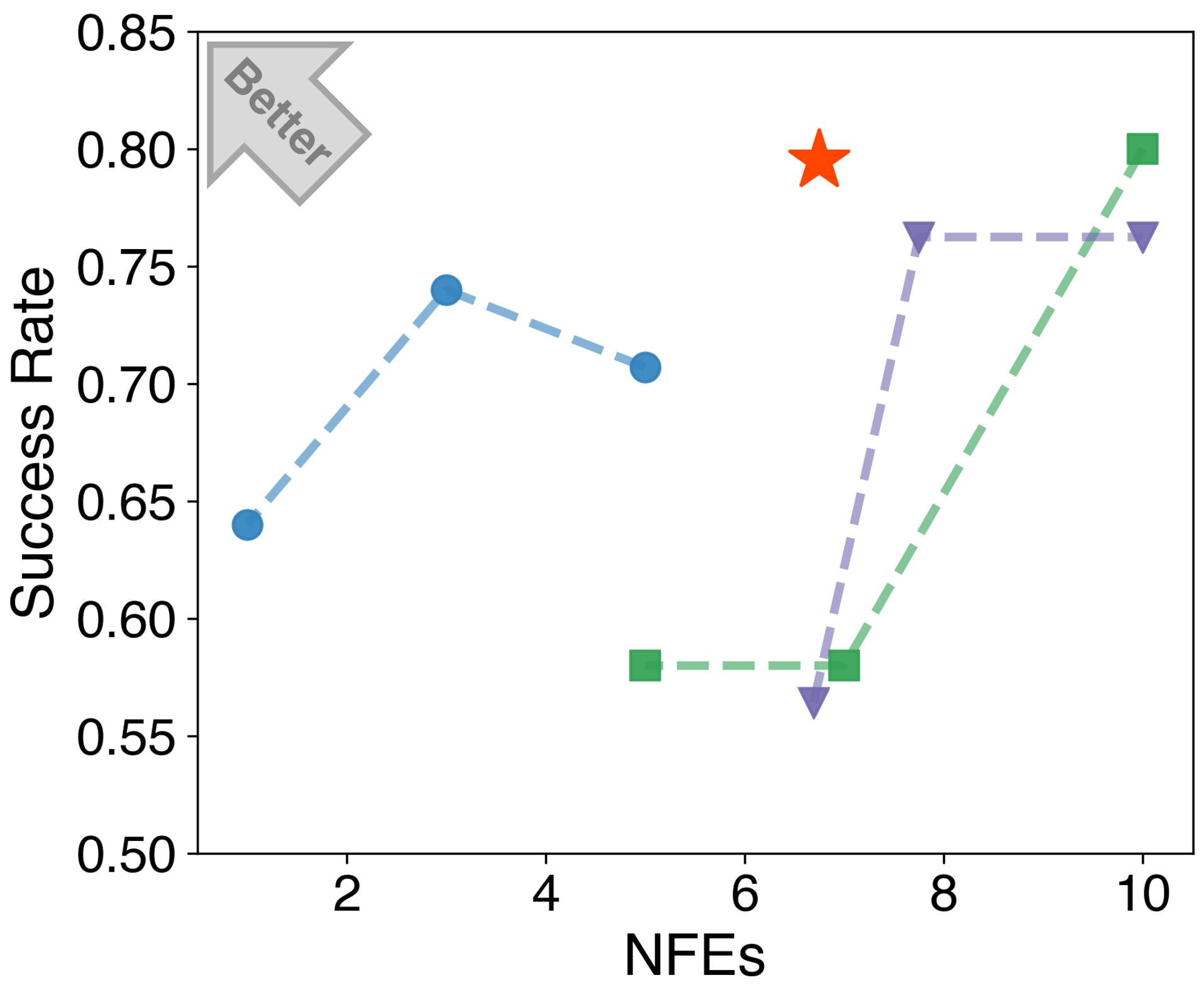}%
    \label{fig:transport} 
  }
  \subfloat[Transport (Pixel)]{%
    \includegraphics[width=0.245\textwidth]{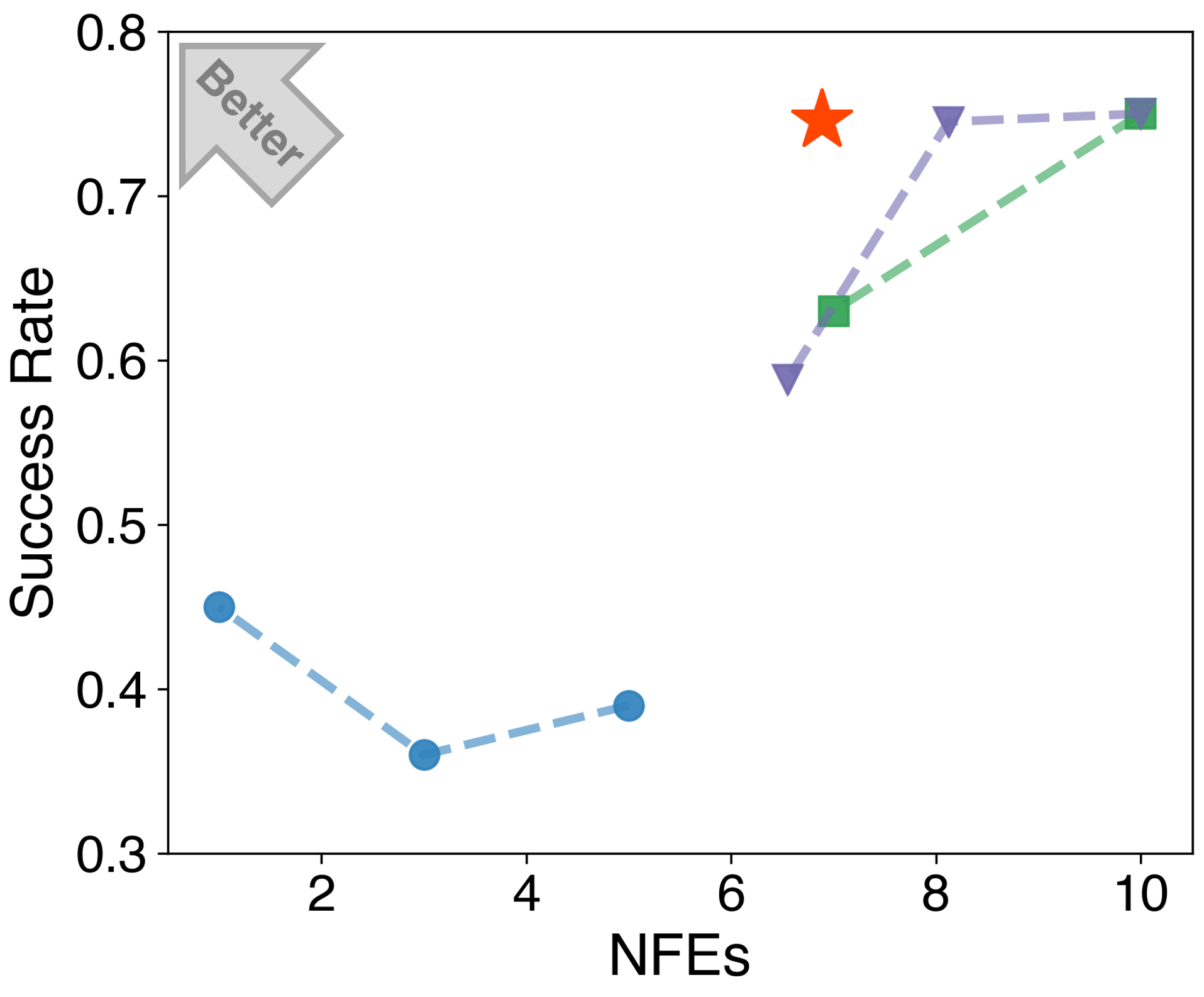}%
    \label{fig:transport-img} 
  }
  \subfloat[Kitchen-complete-v0]{%
    \includegraphics[width=0.245\textwidth]{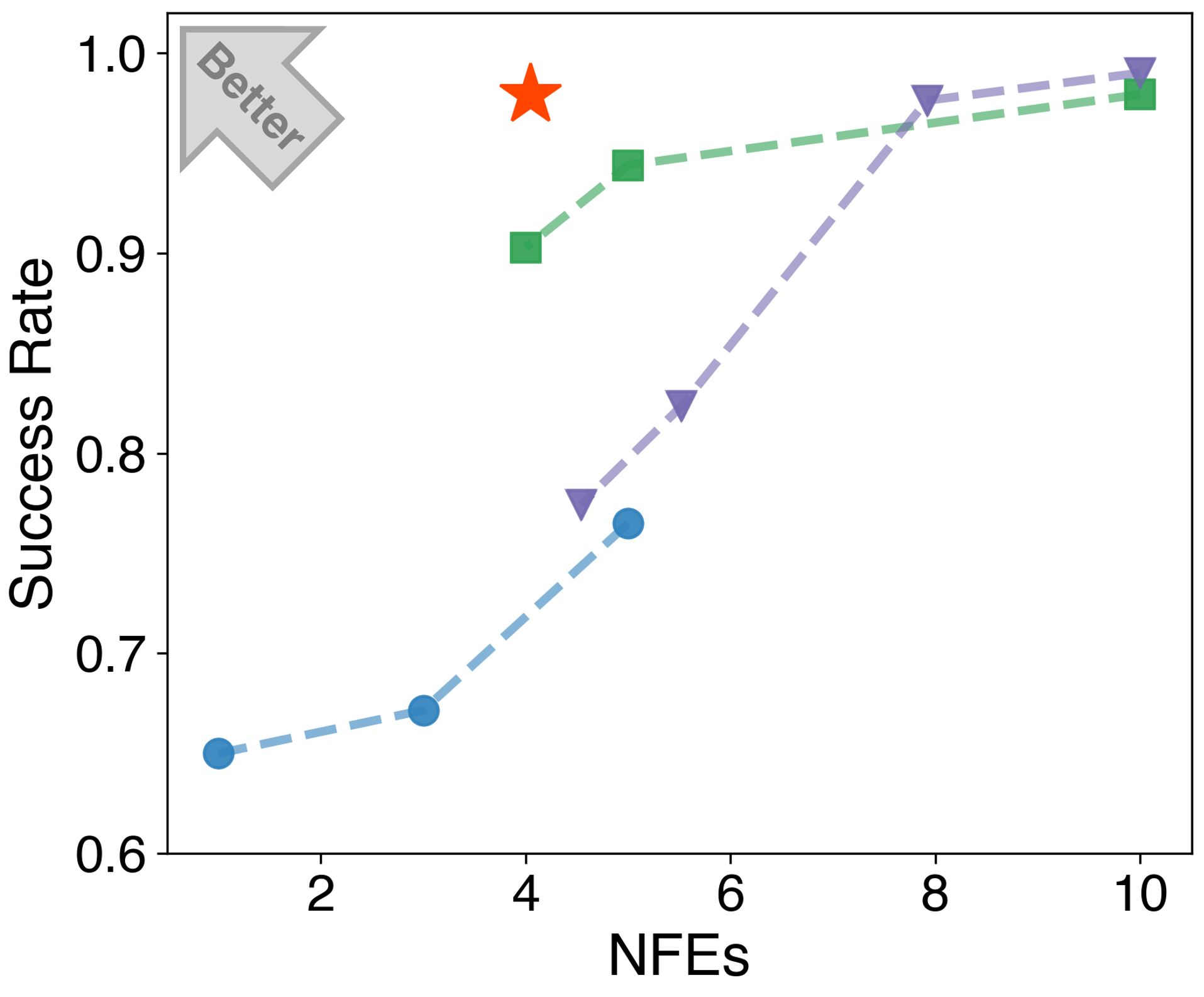}%
    \label{fig:kitchen-complete} 
  }
  \subfloat[Kitchen-mixed-v0]{%
    \includegraphics[width=0.245\textwidth]{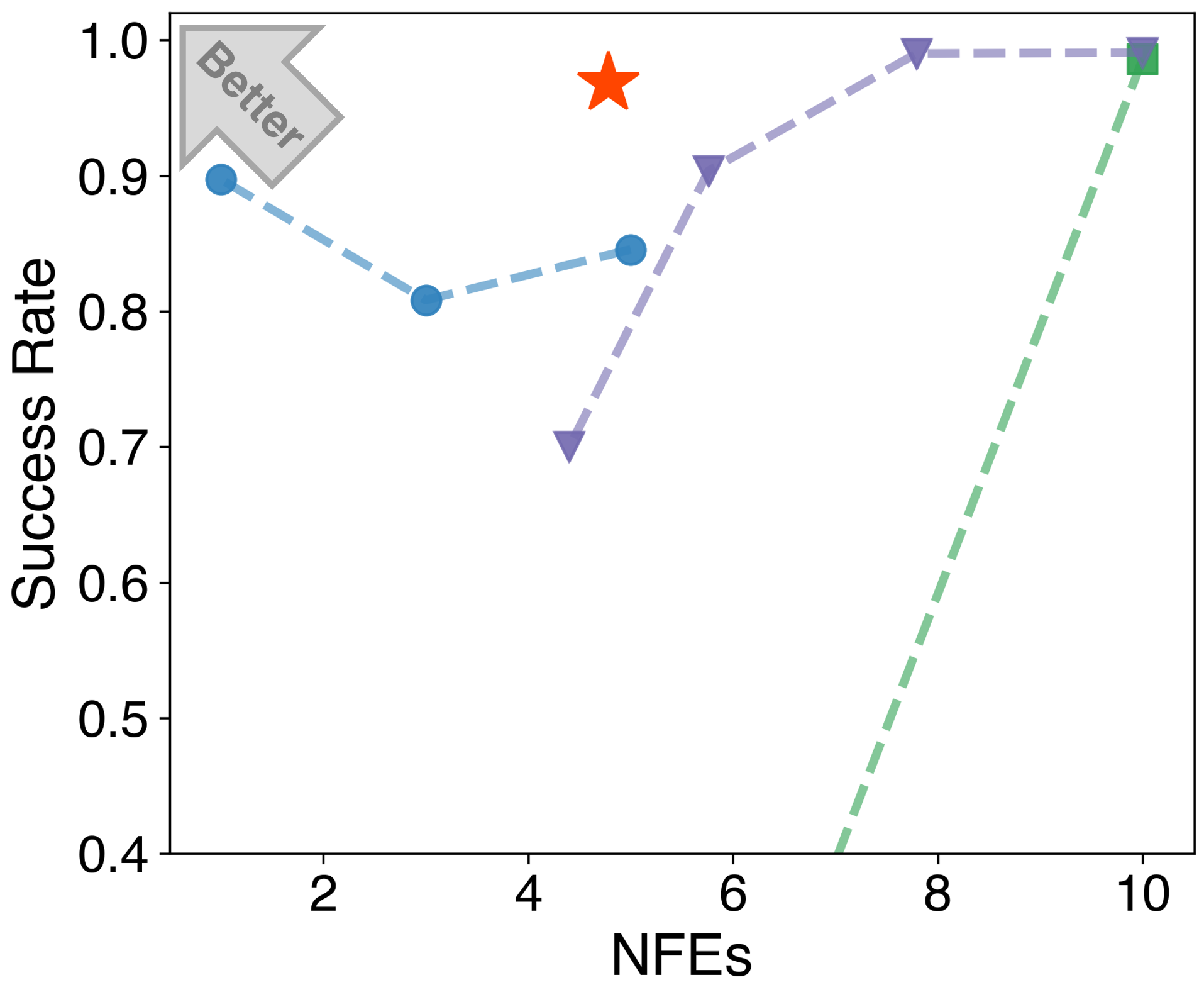}%
    \label{fig:kitchen-mixed} 
  }
  \\
  \subfloat{
  \includegraphics[width = 0.5\textwidth]{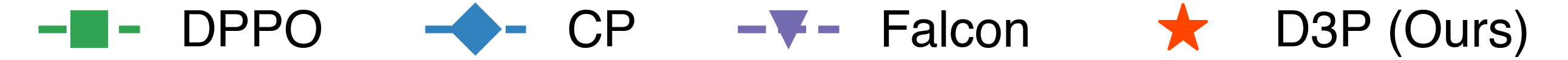}
  }
  \caption{We plot success rate / episodic return against the Number of Function Evaluations (NFE) per action. We use episodic return for the simpler \texttt{Lift} and \texttt{Can} tasks because success rates for most methods saturate above 90\%. D3P achieves the best in performance and efficiency across all tasks. D3P matches or surpasses the peak performance of the 10-step DPPO baseline while achieving an average 2.2$\times$ speed-up. All results are averaged over 3 seeds with 100 evaluation episodes per seed.}
  \label{fig:main_res} 
\end{figure*}

\subsection{Setups}
\subsubsection{Environments}
We evaluate our method on eight manipulation tasks from two benchmarks: Robomimic~\cite{robomimic2021} and Franka Kitchen~\cite{gupta2019relay}. These environments include simple pick-and-place tasks and challenging long-horizon, multi-stage assembly tasks.

\subsubsection{Baselines}
We compare D3P against three representative baselines covering different paradigms: (1) DPPO~\cite{ren2024dppo}, a state-of-the-art (SOTA) algorithm for \textit{online fine-tuning} DPs, (2) consistency policy (CP)~\cite{prasad2024consistency}, a \textit{distillation-based acceleration} method, and (3) Falcon~\cite{chen2025falcon}, a training-free \textit{streaming} approach. To ensure a fair comparison, all policies are pre-trained on the same dataset. Using the pre-trained policy, we train D3P and the DPPO policy with an identical amount of online RL data. The Falcon and consistency policy baselines are subsequently derived from the DPPO fine-tuned policy.

\subsubsection{Metrics}
We evaluate task performance using success rate and episodic return ($J = \sum_{\tau=0}^T r_\tau$) where higher values indicate better performance.
To assess computational efficiency, we follow \citet{prasad2024consistency, chen2025falcon} and adopt the Number of Function Evaluations (NFE) per action as our primary metric. NFE provides a fair comparison of inference cost because all methods use the same base network architecture. For DPPO and CP, the NFE count is equivalent to the number of denoising steps. For Falcon, its selection mechanism requires an additional base policy inference, adding one NFE per action. For our method, we only count the NFE from the base policy, as our lightweight adaptor has fewer than $1/15$ the parameters of the base policy. All results are averaged over three random seeds, with evaluations conducted on 100 episodes per seed.

For additional details on the experimental setup, please refer to \textbf{Appendix B}.

\subsection{Performance Evaluation}
\Cref{fig:main_res} compares our D3P against the baselines by plotting their task performance versus inference cost. On the simpler \texttt{Lift} and \texttt{Can} tasks, the success rates for most methods saturate above 90\% and we use episodic return as a more fine-grained performance metric. In this figure, an ideal method would occupy the upper-left corner, signifying high performance achieved at a low inference cost.

As expected, the performance of the DPPO fine-tuned policy correlates with its inference cost. Reducing the NFEs improves training and inference speed, yet leads to a clear performance drop in success rate and return. Similarly, while Falcon accelerates the 10-step DPPO policy, its performance consistently decreases as the acceleration ratio increases. The success rate (or return) of Falcon degrades sharply under 6 NFE. Consistency Policy, through distillation, enables few-step or even single-step inference. However, the distillation process creates a performance ceiling that prevents it from matching the optimal, multi-step teacher policy. 

In contrast, D3P dynamically adapts its denoising effort, using more denoising steps only when necessary. This allows D3P to match or even exceed the peak performance of the 10-step DPPO policy while achieving an average inference speed-up of 2.2 times. The results unequivocally demonstrate that D3P establishes a better Pareto frontier, consistently achieving an optimal performance-efficiency trade-off than all baselines across all tasks. Due to space constraints, detailed training curves are provided in \textbf{Appendix C}.

\subsection{Ablation study}
We perform ablation studies on the \texttt{Square} and \texttt{Kitchen-complete-v0} tasks to isolate the contrinbutions of D3P's key design choices.

\begin{figure}[!tb]
  \centering 
  \subfloat[Square]{%
    \includegraphics[width=0.492\linewidth]{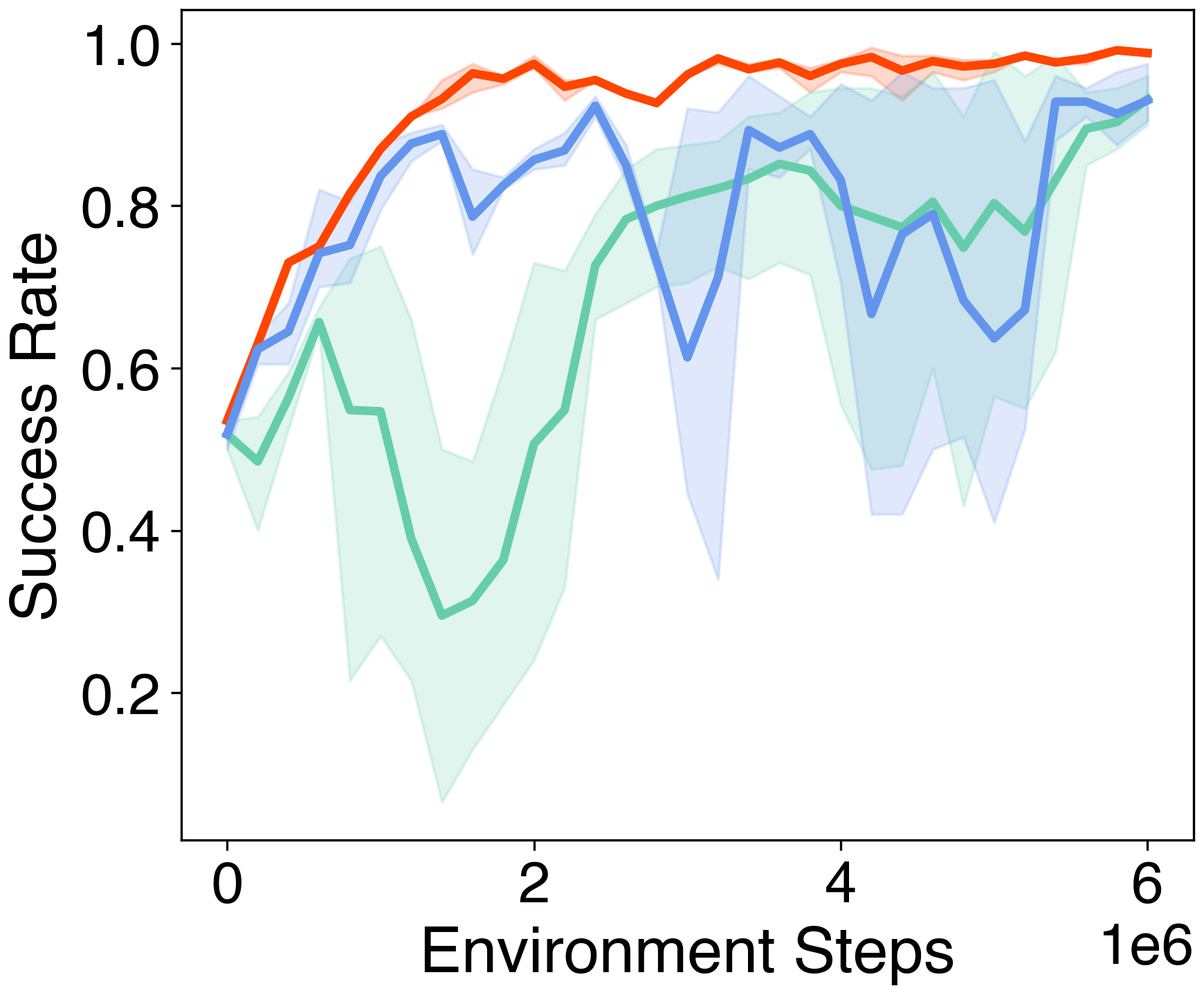}%
    \label{fig:abl-square-sr-3stage} 
  }
  \subfloat[Kitchen-complete-v0]{%
    \includegraphics[width=0.492\linewidth]{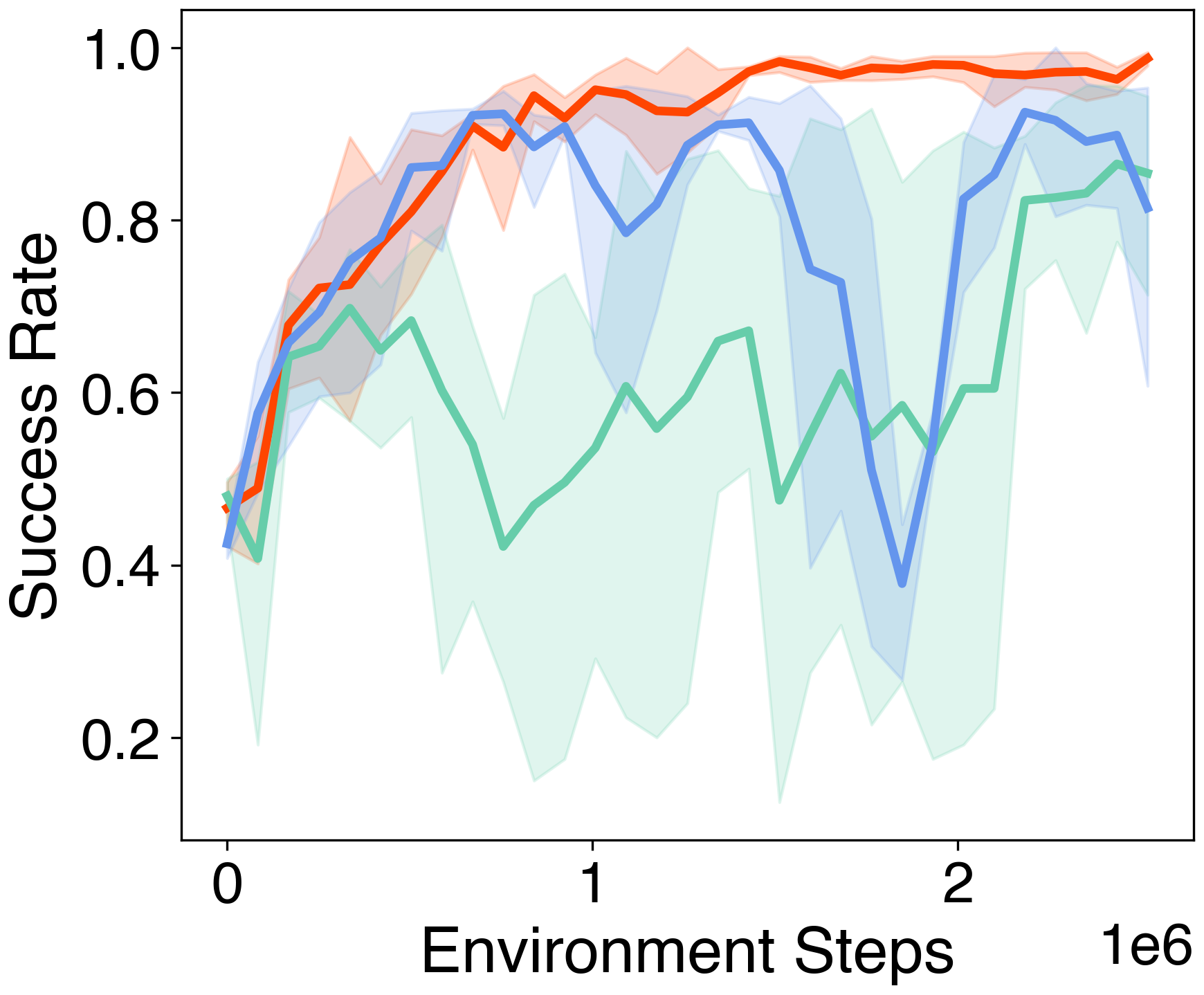}%
    \label{fig:abl-square-se-rew} 
  }
  \\
  \subfloat{
    \includegraphics[width = 0.95\linewidth]{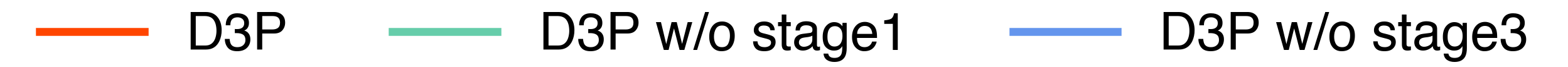}
  }
  \caption{Ablation of the three-stage training strategy. Removing Stage 1 impairs initial learning, while removing Stage 3 destabilizes final convergence. The full strategy is critical for achieving rapid and stable performance.}
  \label{fig:abl-3stage}
\end{figure}

\Cref{fig:abl-3stage} validates the importance of our three-stage training strategy. Removing stage 1 leads to a significant performance drop at the start of the training, while removing stage 3 causes unstable curves during later training. Our full three-stage approach effectively warms up the policy and then stabilizes the fine-tuning process, proving crucial for guiding a robust, optimal convergence.

\begin{figure}
  \centering 
  \subfloat[Square]{%
    \includegraphics[width=0.492\linewidth]{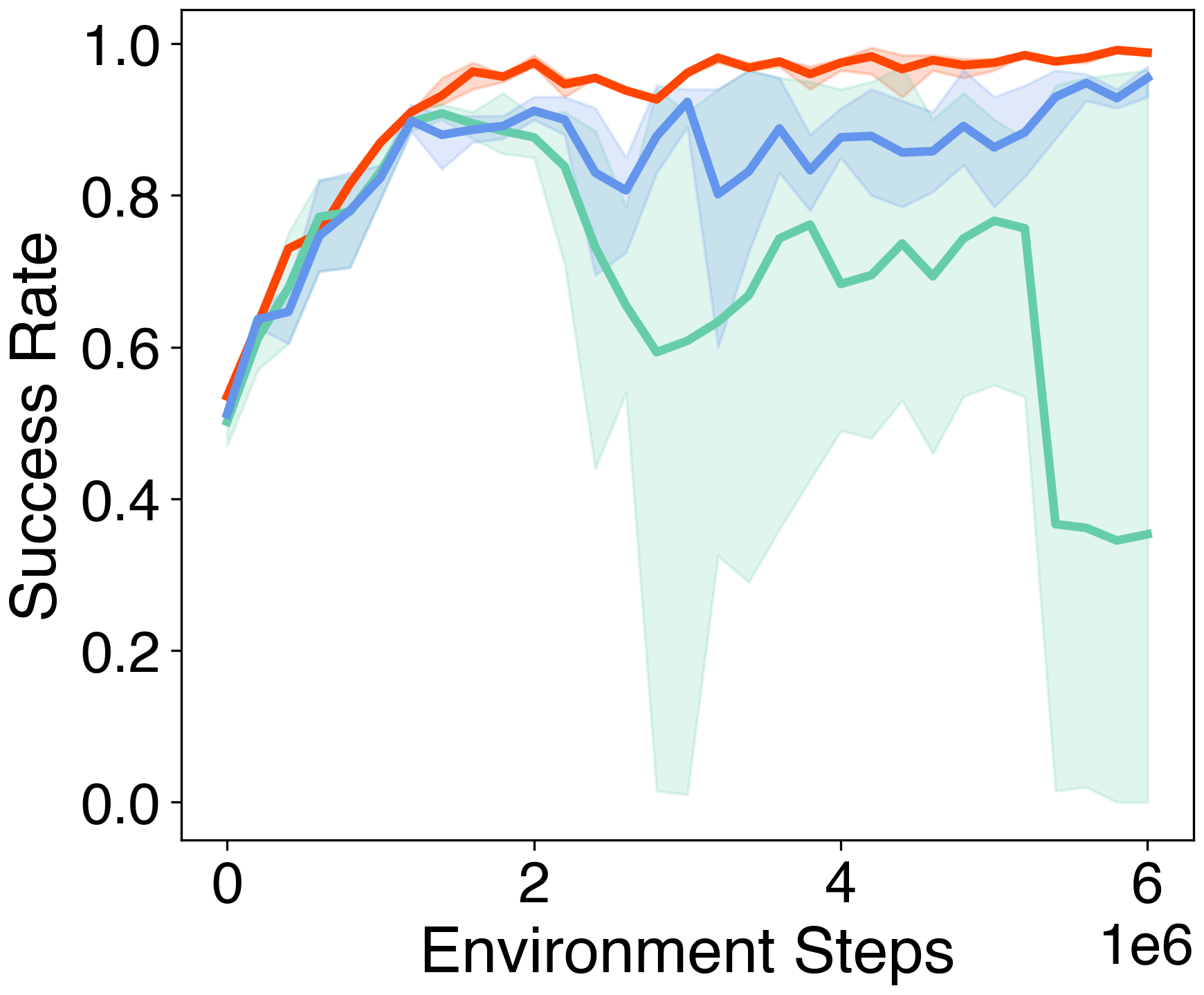}%
    \label{fig:abl-square-sr-rew} 
  }
  \subfloat[Kitchen-complete-v0]{%
    \includegraphics[width=0.492\linewidth]{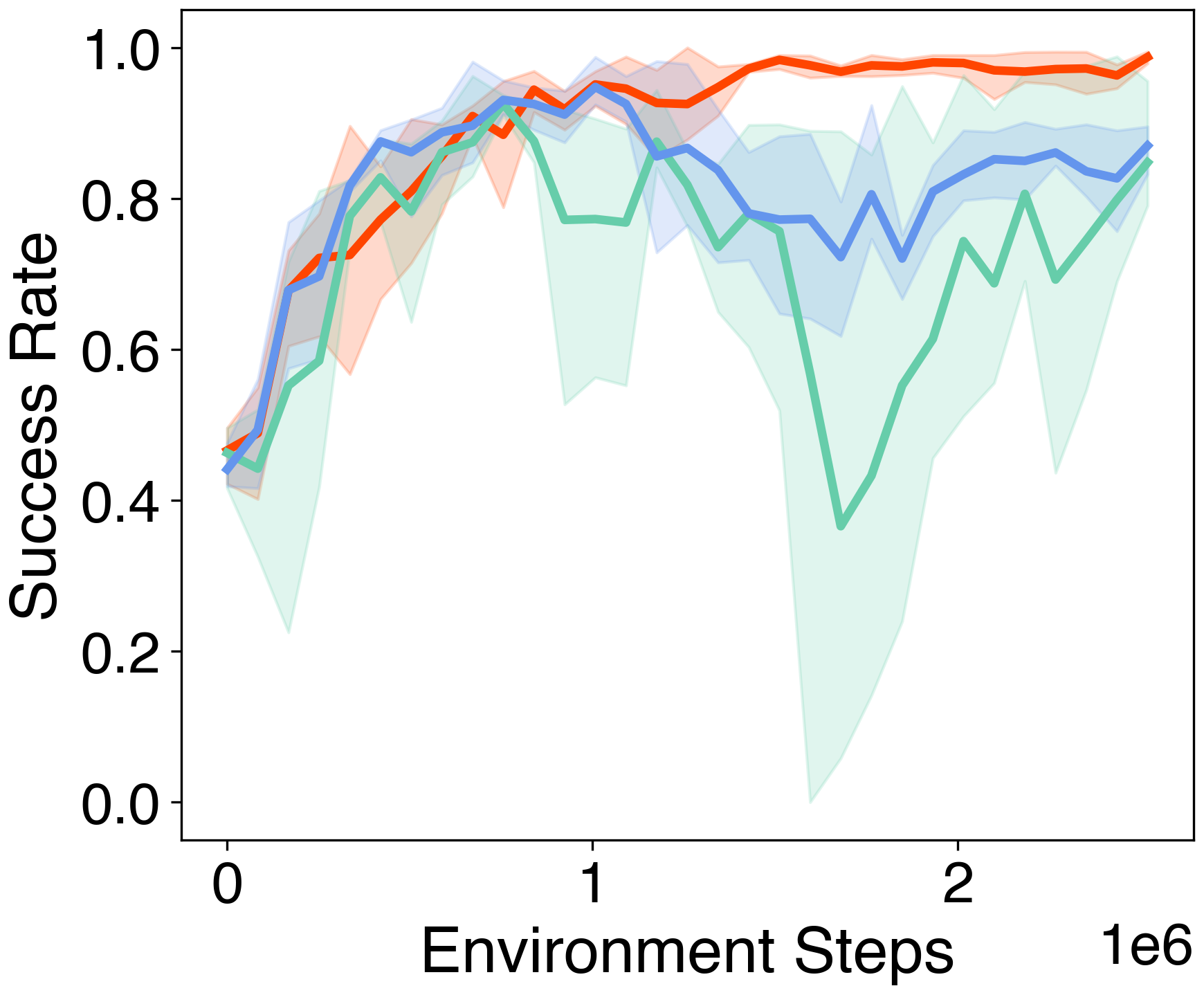}%
    \label{fig:abl-square-sr-rew} 
  }
  \\
  \subfloat{
    \includegraphics[width = 0.816\linewidth]{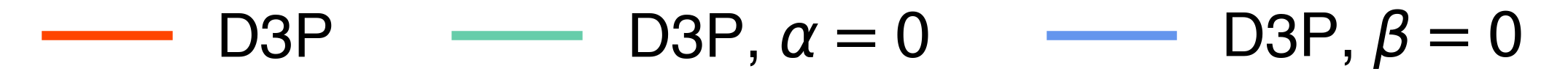}
  }
  \caption{Ablation of our reward formulation. Relying solely on the success reward ($\alpha=0$) causes training instability, while using only the advantage term ($\beta=0$) leads to suboptimal convergence. Both components are essential for stable training towards an optimal policy.}
  \label{fig:abl-rew}
\end{figure}

\Cref{fig:abl-rew} shows the analysis of our reward formulation in \Cref{eq:rew1}. Setting $\alpha = 0$ leaves only the unbiased but delayed success reward. This high-variance signal makes training unstable and prone to failure. Setting $\beta = 0$ leaves only the low-variance but biased advantage term. While the advantage term stabilizes training, the policy is more likely to converge to a suboptimal solution.

\subsection{Real-world Deployment}
To demonstrate D3P's effectiveness in the physical world, we deploy the policy on a Franka robot arm. All inference was performed on a consumer-grade desktop (i7-12900K CPU, RTX 2080 GPU). We mitigate the visual sim-to-real gap with a latent diffusion model~\cite{rombach2021highresolution} that aligns real-world images simulated ones. As illustrated in \Cref{fig:real}, D3P successfully performs the \texttt{Square} task. For crucial actions such as grasping and aligning, D3P increases its denoising steps to 8 and 6, to generate accurate actions. Conversely, for simpler motions, it reduces the step count to as low as 3. D3P achieves the control frequency of 33.68 Hz, a 1.92$\times$ speedup over the 17.59 Hz of a fixed 10-step diffusion policy. Additional deployment details are provided in \textbf{Appendix D}.

\begin{figure}
    \centering
    \includegraphics[width=1.0\linewidth]{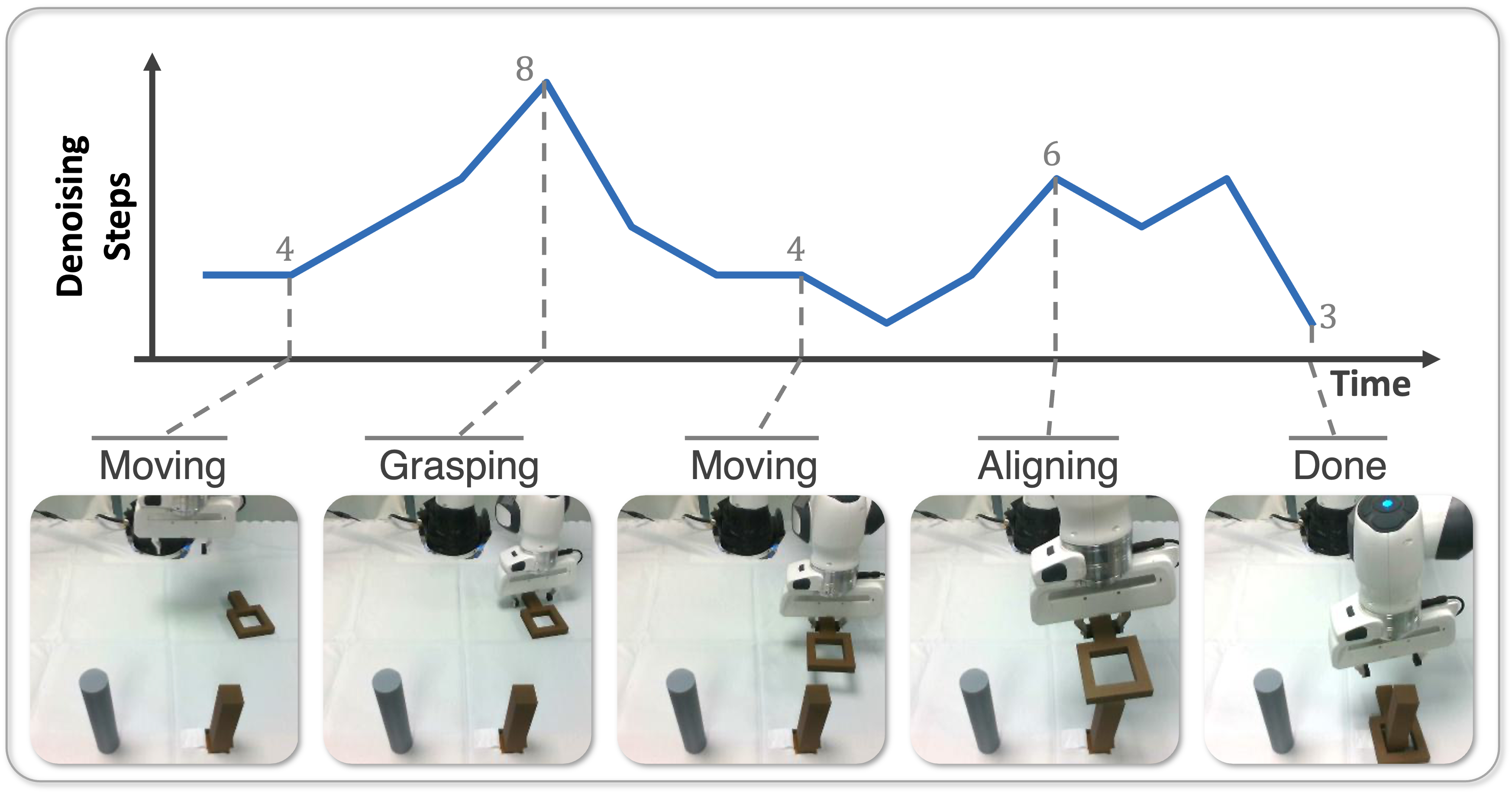}
    \caption{Real-world demonstration of D3P performing the \texttt{Square} task. The plot shows that D3P dynamically adjusts the number of denoising steps during task execution. It allocates more steps for crucial actions, such as grasping and aligning, and fewer for routine movements.}
    \label{fig:real}
\end{figure}

\section{Related Work}
\subsubsection{Optimizing Diffusion Policies via RL}
To overcome the data dependency of imitation learning~\cite{chi2023diffusion, pearce2023imitatinghumanbehaviourdiffusion, wang2024one, prasad2024consistency}, many methods use RL to optimize DPs. In offline RL, methods adapt DPs using techniques derived from Q-learning~\cite{wang2022diffusion, hansen2023idql} or policy gradients~\cite{kang2023efficient}. In the online setting, DPs are often trained within actor-critic frameworks~\cite{wang2024diffusion, yang2023policy, ren2024dppo, li2024learning} or with action-gradients from a learned Q-function~\cite{psenka2023learning}. However, these methods apply a fixed number of denoising steps for all actions, leaving the critical issue of slow inference speed unresolved.

\subsubsection{Accelerating Diffusion Policy Inference}
To improve the inference speed of DPs, a straightforward way is reduce the number of denoising steps. Prior work primarily use policy distillation~\cite{prasad2024consistency, wang2024one} or streaming denoising~\cite{hoeg2024streaming, chen2025falcon}. A key limitation is that these methods treat all actions as equally important. In contrast, our method, D3P, adaptively adjusts the computational effort for each action.

While the concept of adaptive denoising exists for single-image generation~\cite{ye2025schedule}, the sequential decision-making of robotics presents distinct challenges with long-term rewards and temporal dependencies. We address this by formulating the dynamic denoising problem as a two-layer POMDP and jointly optimizing the base diffusion policy and the adaptor. The training process is stabilized by a specialized reward and a three-stage training strategy.
\section{Conclusion}
In this work, we introduced Dynamic Denoising Diffusion Policy (D3P), a diffusion policy capitalizing on the varying action criticalities in robotic tasks. D3P employs a lightweight adaptor to dynamically adjust denoising steps, assigning more steps to crucial actions and fewer to routine ones. We use RL to joint optimize the base policy and the adaptor with a carefully-designed reward and a three-stage training strategy. Our simulation experiments demonstrate that D3P achieves an averaged 2.2$\times$ inference speed-up over baselines without compromising task success. Furthermore, D3P is deployed on a physical robot, achieving a 1.9$\times$ inference acceleration against a fixed-step diffusion policy. These results underscore the potential of adaptive inference in robot learning, developing more efficient policies for real-time applications.

\newpage
\bibliography{ref}

\setcounter{secnumdepth}{2}
\clearpage
\onecolumn
\appendix
\makeatletter
\def\section{\@startsection {section}{1}{\z@}{-2.0ex plus
-0.5ex minus -.2ex}{3pt plus 2pt minus 1pt}{\Large\bf\raggedright}}
\makeatother
\section{Implementation Details of Empirical Study}
Our empirical study reveals that not all actions in a robotic task contribute equally to the results. This section details the specifics of our empirical study.

As introduced in the \textbf{Empirical Study}, we train a return predictor, denoted as $D_\phi:\mathcal{O}_\text{ENV} \times \mathcal{A}_\text{ENV} \to \mathbb{R}$, to predict the subsequent return based on current observation $o_t$ and the unperturbed action $a_t$. This predictor is trained via supervised learning on a dataset collected from Monte-Carlo rollouts. The detailed procedure for data collection and training is presented in \Cref{alg:criticality_training}, with all notations consistent with the main text.

\begin{algorithm}
    \caption{Train a Return Predictor $D_\phi$}
    \label{alg:criticality_training}
    \begin{algorithmic}[1]
        \State \textbf{Input:} Task environment $\mathcal{M}_\text{ENV}$, expert policy $\pi_{\text{expert}}$, number of episodes $L$, discount factor $\gamma_{\text{ENV}}$, episode horizon $T$, update interval $l_{\text{int}}$, update epoch $Ep$, max buffer length $M$, variance $v$.
        \State \textbf{Initialize:} Network $D_\phi$, Empty buffer $\mathcal{B}$ with max length $M$.
        \For{$l=0, 1, \dots, L-1$}
            \State Reset environment to get $o_0$
            \State Sample $t_l \sim \text{Uniform}\{0, 1, \dots, T-1\}$
            \For{$t=0, 1, \dots, T-1$}
                \State Get expert action $a_t\sim \pi_{\text{expert}}(\cdot \mid o_t)$
                \If{$t==t_l$}
                    \State $\xi\sim\mathcal{N}\left(0, v^2\right)$, $a'_t \leftarrow a_t + \xi$
                    \State Execute $a'_t$ to get $o_{t+1}$ and $r_t$
                \Else
                    \State Execute $a_t$ to get $o_{t+1}$ and $r_t$
                \EndIf
            \EndFor
            \State Calculate $J_l \leftarrow \sum_{\tau=0}^T \gamma_\text{ENV}^{\tau-t_l} r_\tau$
            \State Add $(o_{t_l}, a_{t_l}, J_l)$ to $\mathcal{B}$
            \If{$l\ \%\ l_{\text{int}} == 0$}
                \For{$e=0, 1, \dots, Ep-1$}
                    \State Update $D_\phi$ on $\mathcal{B}$ by minimizing the loss: $\mathcal{L}(\phi) = \frac{1}{L} \sum_{(o, a, J) \in \mathcal{B}} (D_\phi(o, a) - J)^2$
                \EndFor
            \EndIf
        \EndFor
        \State \textbf{Return:} Trained predictor $D_\phi$.
    \end{algorithmic}
\end{algorithm}

We parameterize $D_\phi$ as a 5-layer MLP with hidden layer sizes of $[256, 512, 1024, 512, 256]$. For our experiments on the \texttt{Square} and \texttt{Transport} tasks, we use $L=600$ and $L=350$, respectively. The full training hyperparameters are provided in \Cref{tab:emp_hp}, with training curves shown in \Cref{fig:emp_loss}. Detailed settings for both tasks are available in \textbf{Appendix B}.

\begin{figure}[hbp]
    \centering
    \subfloat[Square]{
        \includegraphics[width = 0.3\textwidth]{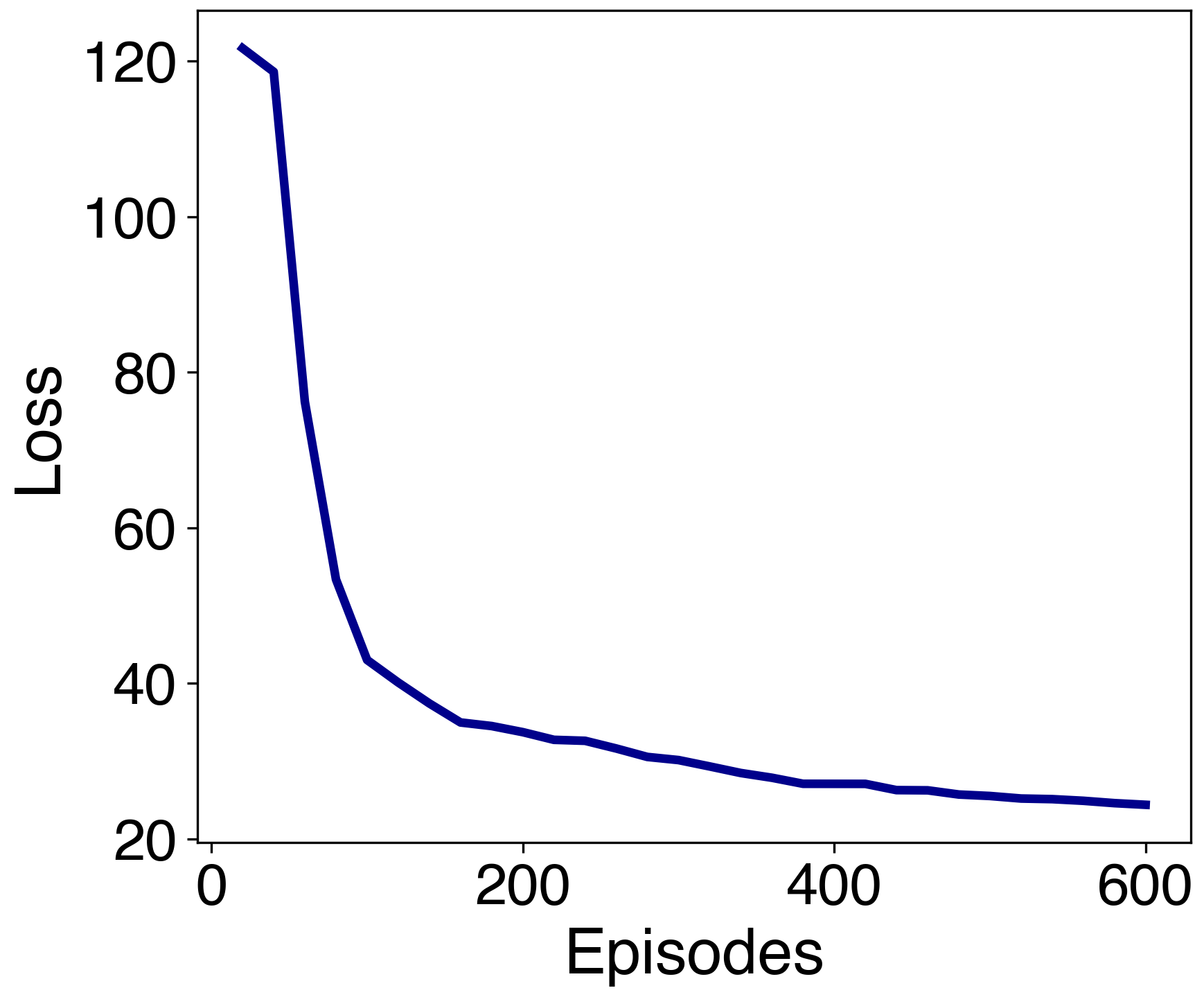}
    }
    \subfloat[Transport]{
        \includegraphics[width = 0.3\textwidth]{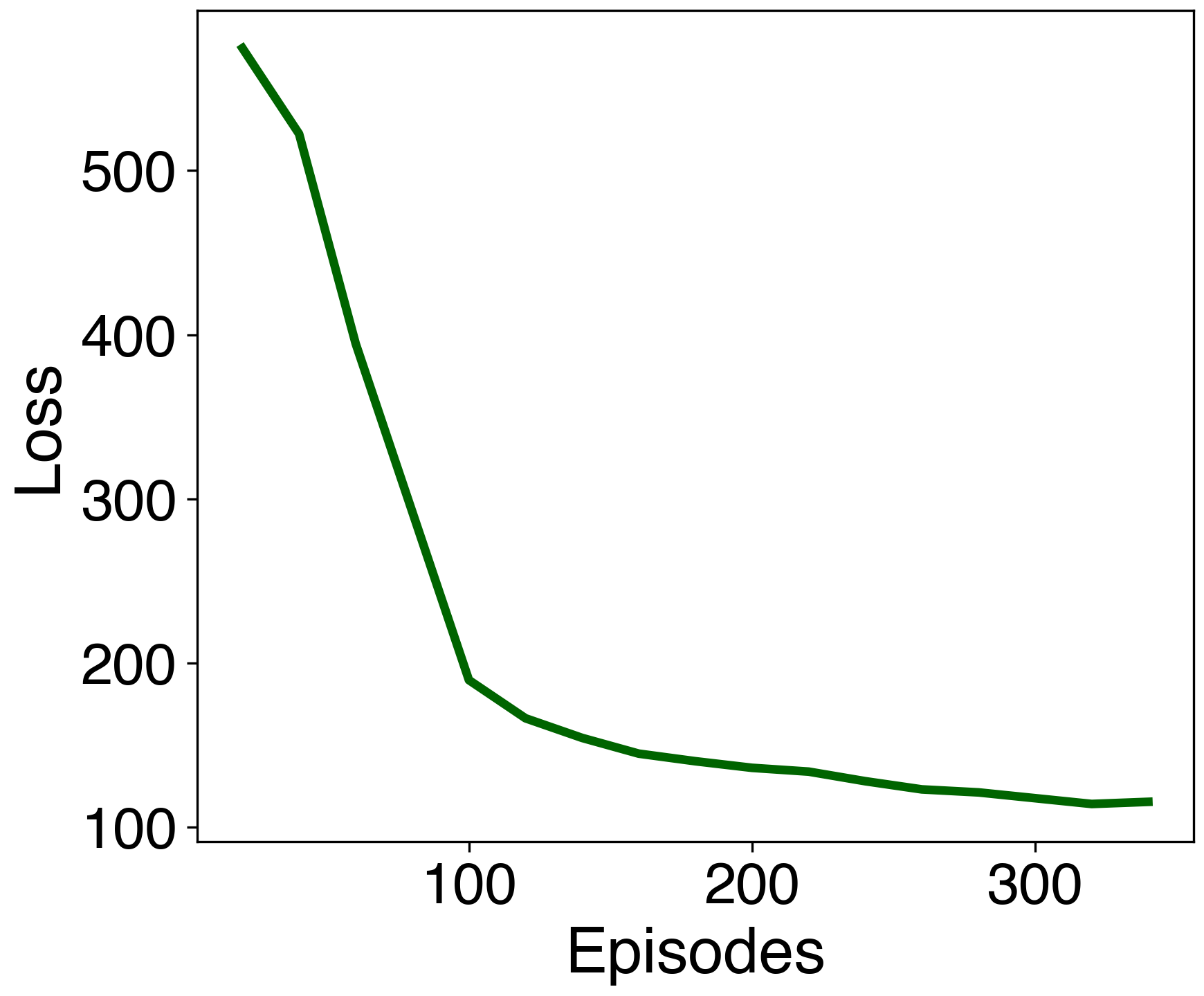}
    }
    \caption{Training loss of the return predictor $D_\phi$ versus training episodes for the (a) \texttt{Square} and (b) \texttt{Transport} tasks}
    \label{fig:emp_loss}
\end{figure}

\begin{table}
    \centering
    \begin{tabular}{cc}
        \toprule
        Hyperparameter & Value \\
        \midrule
        max buffer size $M$ & 100000\\
        number of parallel environments & 10\\
        discount factor $\gamma_{\text{ENV}}$ & 0.99\\
        noise variance $v$ & 0.1 \\
        update interval $l_{\text{int}}$ & 20 \\
        update epoch $Ep$ & 6\\
        learning rate & 0.0003\\
        weight decay & 0.0001\\
        \bottomrule
    \end{tabular}
    \caption{Hyperparameters for training the return predictor $D_\phi$.}
    \label{tab:emp_hp}
\end{table}

\newpage
\section{Additional details of simulation experiments}
\subsection{Environment and Dataset}
\subsubsection{Environment}

\begin{figure}
    \centering
    \subfloat[Lift]{
        \includegraphics[width = 0.23\textwidth]{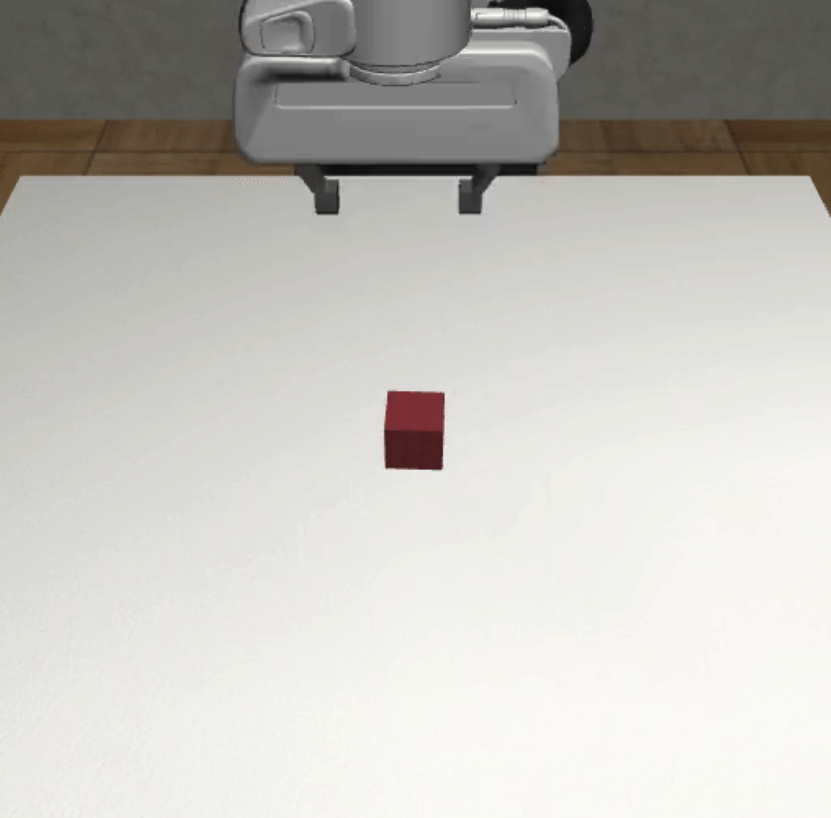}
    }
    \subfloat[Can]{
        \includegraphics[width = 0.23\textwidth]{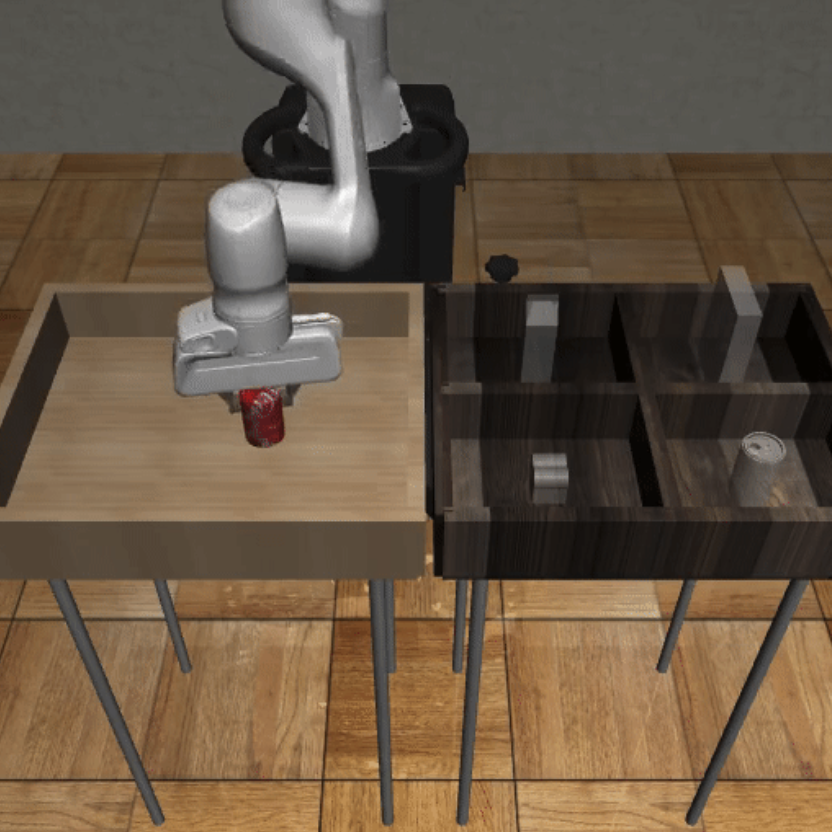}
    }
    \subfloat[Square]{
        \includegraphics[width = 0.23\textwidth]{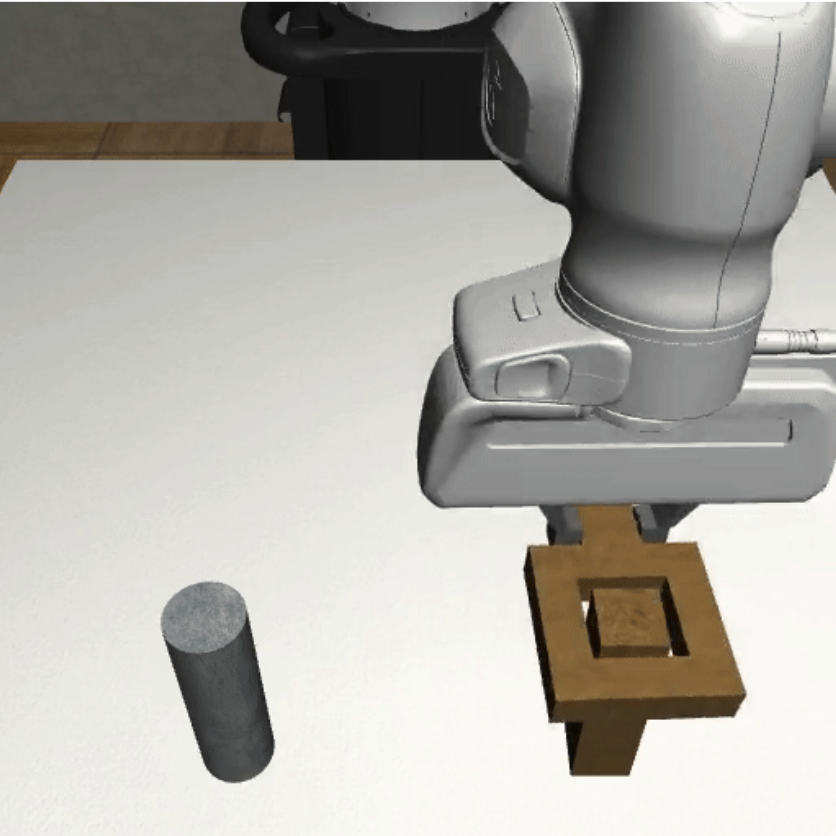}
    }
    \subfloat[Transport]{
        \includegraphics[width = 0.23\textwidth]{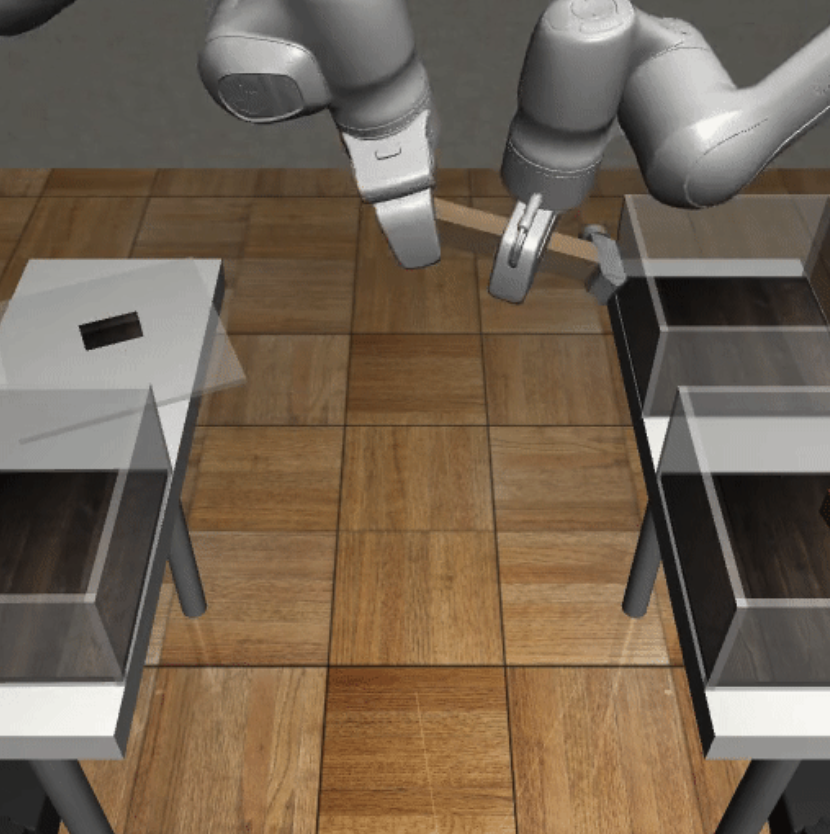}
    }
    \caption{Four manipulation tasks in Robomimic}
    \label{fig:robomimic}
\end{figure}

\begin{figure}
    \centering
    \includegraphics[width = 0.25\textwidth]{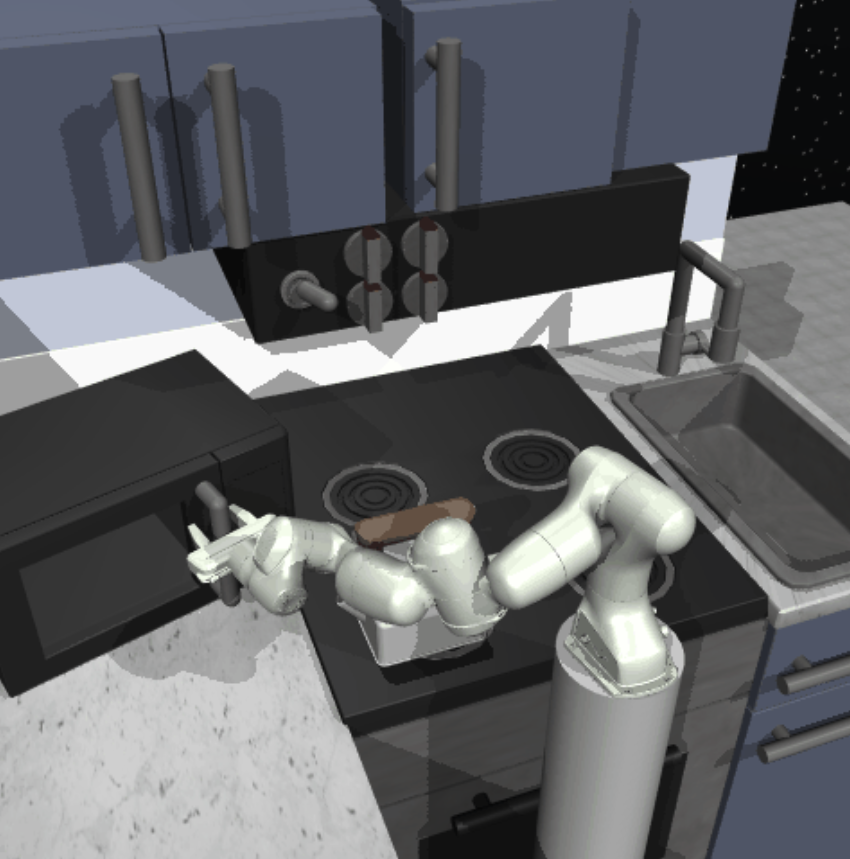}
    \caption{Franka Kitchen environment. In the environment, the robot need to complete 4 subtasks in sequence: open the microwave, move the kettle, flip the light switch, and slide open the cabinet door.}
    \label{fig:kitchen}
\end{figure}

We evaluate D3P in two benchmarks: 
\begin{enumerate}
    \item \textbf{Robomimic}~\cite{robomimic2021}. We evaluate our method on manipulation tasks from the Robomimic suite, including the simple pick-and-place tasks \texttt{Lift} and \texttt{Can}, the assembly task \texttt{Square}, and the bimanual handover task \texttt{Transport}. \Cref{fig:robomimic} demonstrate the four manipulation tasks.
    \item \textbf{Franka Kitchen}~\cite{gupta2019relay}. We also use the Franka Kitchen environment, a benchmark for challenging long-horizon, multi-stage manipulation. As illustrated in \Cref{fig:kitchen}, the robot need to complete 4 subtasks in sequence: open the microwave, move the kettle, flip the light switch, and slide open the cabinet door. We use two settings in this environmrnt: (1) \texttt{Kitchen-complete-v0}, where the policy is pretrained on a dataset of successful demonstrations, and (2) \texttt{Kitchen-mixed-v0}, where the pretraining dataset contains contains various subtasks being performed, but the 4 target subtasks are never completed in sequence together.
\end{enumerate}

We list the task configurations in \Cref{tab:tasks}. In this table, ``Obs dim (State)'' indicates the dimension of the state observation, ``Obs dim (Image)'' is the dimension of the pixel observation, Act dim $\times T_a$ shows the shape of an action chunk, and $T$ is the episode horizon. 

All tasks employ a sparse reward. In Robomimic tasks, the agent receives a reward of 0 for every timestep prior to task completion. Upon successful completion, the agent is awarded a +1 reward for all subsequent timesteps until the episode concludes. Therefore, the episodic return directly reflects the agent's quality. A higher return indicates faster task completion, whereas a lower return suggests the task was only finished near the end of the episode. If the task is not completed within the maximum episode length of $T$ steps, the total episodic return is 0. For the Franka Kitchen environment, the agent receives a +1 reward upon the completion of each sub-task. As there are 4 sub-tasks, the maximum possible episodic return is 4.

\begin{table}
    \centering
    \begin{tabular}{ccccccc}
        \toprule
        Benchmark & Task & Obs dim (State) & Obs dim (Pixel) & Act dim $\times\ T_a$  & $T$ & Sparse reward \\
        \midrule
        \multirow{2}{*}{Franka Kitchen} & {kitchen-complete-v0} & 60 & - & 9$\times$4 & 280 & Yes \\
        ~ & {kitchen-mixed-v0} & 60 & - & 9$\times$4 & 280 & Yes \\
        \midrule
        \multirow{6}{*}{Robomimic} & {Lift} & 19 & - & 7$\times$4 & 300 & Yes \\
        ~ & {Can} & 23 & - & 7$\times$4 & 300 & Yes \\
        ~ & {Square} & 23 & - & 7$\times$4 & 400 & Yes \\
        ~ & {Transport} & 59 & - & 14$\times$8 & 800 & Yes \\
        ~ & {Square} (Pixel) & 9 & $(3, 96, 96)\times 1$ & 7$\times$4 & 400 & Yes \\
        ~ & {Transport} (Pixel) & 18 & $(3, 96, 96)\times 2$ & 14$\times$8 & 800 & Yes \\
        \bottomrule
    \end{tabular}
    \caption{We detail the task configurations in this table, where``Obs dim (State)'' is the dimension of the state observation, ``Obs dim (Image)'' is the dimension of the pixel observation, ``Act dim'' is the dimension of a single action, $T_a$ is the action chunck horizon, and $T$ is the episode horizon.}
    \label{tab:tasks}
\end{table}

\subsubsection{Dataset}
We pre-train the policies on the dataset provided by one of our baseline, DPPO~\cite{ren2024dppo}. As acknowledged by the authors
of DPPO the dataset includes suboptimal data.

\subsection{Baseline}
We compare D3P against three representative baselines that cover different paradigms for improving or acclerating diffusion policies.

\subsubsection{DPPO}
Diffusion policy policy optimization (DPPO)~\cite{ren2024dppo} is a state-of-the-art algorithm for \textit{online fine-tuning} diffusion policies that operate with a fixed number of denoising steps. DPPO models the problem as a two-layer POMDP to leverage the sequential nature of the diffusion denoising process. It directly optimizes the entire denoising chain using an actor-critic framework. At each iteration, the diffusion policy $\pi_\theta$ is updated with the following PPO-style loss function:
\begin{equation}
    \mathcal{L}_{\theta} = \mathbb{E} \left[\min \left( \hat{A}^{\pi_{\theta_{\text{old}}}} (\bar{o}_{\bar{t}}, a_{\bar{t}}) \frac{\pi_{\theta} (\bar{o}_{\bar{t}}, a_{\bar{t}})}{\pi_{\theta_{\text{old}}} (\bar{o}_{\bar{t}}, a_{\bar{t}})}, \hat{A}^{\pi_{\theta_{\text{old}}}} (\bar{o}_{\bar{t}}, a_{\bar{t}}) \ \text{clip} \left( \frac{\pi_{\theta} (\bar{o}_{\bar{t}}, a_{\bar{t}})}{\pi_{\theta_{\text{old}}} (\bar{o}_{\bar{t}}, a_{\bar{t}})}, 1 - \epsilon_{\text{clip}}, 1 + \epsilon_{\text{clip}} \right) \right) \right],
    \label{eq:dppo}
\end{equation}
where $\bar{t}(t, i) = tN + (N-i-1)$ is the time index in the two-layer POMDP, $\bar{o}_{\bar{t}}$ is the observation, and $a_{\bar{t}}$ is the action. The advantage $\hat{A}^{\pi_\theta}$ is estimated as shown in \Cref{eq:dppo-adv}, using a discount factor of $\gamma^i_{\text{DENOISE}}$.
\begin{equation}
    \hat{A}^{\pi_\theta}(\bar{o}_{\bar{t}}, a_{\bar{t}}) = \gamma^i_{\text{DENOISE}}\left(J_{\pi_\theta}(\bar{o}_{\bar{t}}, a_{\bar{t}})-V_\Theta(\bar{o}_{\bar{t}})\right).
    \label{eq:dppo-adv}
\end{equation}
Notably, DPPO employs a dynamic clipping parameter $\epsilon_{\text{clip}}$, whose value is determined by the denoising progress, $t=1-\frac{i}{N}$, as defined in \Cref{eq:eps}.
\begin{equation}
    \epsilon_{\text{clip}} = \epsilon_{\text{base}} + (\epsilon_{\text{coef}} - \epsilon_{\text{base}}) \cdot \frac{e^{\epsilon_{\text{rate}}\ t} - 1}{e^{\epsilon_{\text{rate}}} - 1}.
    \label{eq:eps}
\end{equation}
Here, $\epsilon_{\text{base}}$, $\epsilon_{\text{coef}}$, and $\epsilon_{\text{rate}}$ are hyperparameters. DPPO provides structured online exploration and enhances training stability. In experiments, we use the official hyperparameter settings from the original implementation, detailed in \Cref{tab:dppo_hp}.

\begin{table}
    \centering
    \begin{tabular}{lc}
    \toprule
        Hyperparameter & Value \\
        \midrule
        Number of parallel environments & 40 \\
        Condition horizon $T_o$ & 1 \\
        Reward discount factor $\gamma_{\text{ENV}}$ & 0.999 \\
        Advantage discount factor $\gamma_{\text{DENOISE}}$ & 0.99 \\
        GAE $\lambda$ & 0.95 \\
        Optimizer & AdamW \\
        Actor learning rate & 1e-4 \\
        Actor weight decay & 0 \\
        Critic learning rate & 1e-3 \\
        Critic weight decay & 0 \\
        Batch size & 10000 \\
        Value loss coefficient & 0.5 \\ 
        Entropy loss coefficient & 0 \\
        Clip coefficient base $\epsilon_{\text{base}}$ & 0.001\\
        Clip coefficient $\epsilon_{\text{coef}}$ & 0.01 \\
        Clip coefficient rate $\epsilon_{\text{rate}}$ & 3 \\
        Max gradient norm & 10.0\\
        Rollout steps & 400 \\
        Update epoch & 10 \\
        Denoising $\eta$ & 1.0 \\
        \bottomrule
    \end{tabular}
    \caption{Hyperparameters of DPPO}
    \label{tab:dppo_hp}
\end{table}

\subsubsection{Consistency Policy}
A Consistency Policy (CP)~\cite{prasad2024consistency} is created by distilling a pretrained diffusion policy, which enforces self-consistency along the teacher model's learned trajectories. This distillation enables CP to sample actions using very few denoising steps. We denote the CP as $g_\theta(o_t, X_t^i, i, s)$, which is conditioned on the current observation $o_t$, the current action chunk $X_t^i$, the current noise level $i$, and a target noise level $s < i$. The distillation loss combines a Denoising Score Matching (DSM) loss and a Consistency Trajectory Model (CTM) loss:
\begin{equation}
    \begin{aligned}
        & \mathcal{L}_{DSM} = \mathbb{E}_{x_0,\ i} \left[d(x_0, g_\theta (o_t, x_i, i, 0)) \right], \\
        & \mathcal{L}_{CTM} = \mathbb{E}_{x_0, i, s} \left[ d\left(g_\theta \left(o_t, g_\theta(o, X_t^i, i, s), s, 0 \right),\ g_\theta \left(o_t, g_\theta(o, X_t^{i-1}, i-1, s), s, 0 \right)\right) \right].
    \end{aligned}
    \label{eq:cp_loss}
\end{equation}
In these equations, $X_t^{i-1}$ is generated from $X_t^i$ by the teacher policy, and $d(x, y)$ is a pseudo-Huber loss function measuring the distance between $x$ and $y$, as shown in \Cref{eq:hbloss}
\begin{equation}
    d(x, y) = \sqrt{\|x-y\|_2^2 + c^2} - c.
    \label{eq:hbloss}
\end{equation}
During training, the final loss is a weighted sum of $\mathcal{L}_{DSM}$ and $\mathcal{L}_{CTM}$ with coefficients $w_{DSM}$ and $w_{CTM}$, respectively. For our experiments, we distill CPs from our DPPO-finetuned policies. The hyperparameters for distillation are listed in \Cref{tab:cp_hp}.

\begin{table}
    \centering
    \begin{tabular}{lc}
        \toprule
        Hyperparameter & Value \\
        \midrule
        Weight of DSM loss $w_{DSM}$ & 1.0 \\
        Weight of CTM loss $w_{CTM}$ & 1.0 \\
        Optimizer & AdamW \\
        Batch size & 512 \\
        Training epoch & 1500 \\
        Learning rate & 1e-4 \\
        Weight decay & 1e-6 \\
        Learning rate scheduler & \makecell{CosineAnnealingWarmupRestart \\ \cite{loshchilov2016sgdr}} \\
        First cycle steps & 1500 \\
        Warmup steps & 100 \\
        Minimum learning rate & 1e-5 \\
        \bottomrule
    \end{tabular}
    \caption{Hyperparameters of consistency policy}
    \label{tab:cp_hp}
\end{table}

\subsubsection{Falcon} Falcon~\cite{chen2025falcon} is a training-free \textit{streaming} method that accelerates diffusion policies through partial denoising. The core insight of Falcon is to leverage the sequential dependency inherent in such tasks. Instead of initiating the denoising process from a standard Gaussian distribution for every action, Falcon reuses a partially denoised action from a latent buffer of historical actions, thereby significantly reducing the required number of sampling steps.

The selection of this action prior is managed by a thresholding mechanism. Falcon first uses the unexecuted action sequence from the previous timestep as a reference. Then, for each partially denoised action a $X_\tau^i$ in the latent buffer, Falcon computes a one-step estimation $\hat{X}_t^0$ conditioned on the current observation $o_t$ with Tweedie's formula (\Cref{eq:falcon}), where $\epsilon_\theta$ is the noise-predicting network defined in \text{Preliminaries}, and $\bar{\alpha}_i$ is a set of parameters determinated by a fixed noise schedule.
\begin{equation}
    \hat{X}_t^0 = \mathbb{E}[\hat{X}_t^0|o_t, X_t^i, i] = \frac{X_t^t-\sqrt{1-\bar{\alpha}_i}\epsilon_{\theta}(o_{t},X_{t}^{i},i)}{\sqrt{\bar{\alpha}_i}}.
\label{eq:falcon}
\end{equation}
Then we form a candidate set $\mathcal{S}$, consists of actions whose one-step estimations are within a certain distance $\epsilon_{\text{Falcon}}$ of the reference action. The final prior action $X_t^i$ is then sampled from this set with a probability that favors lower noise levels, modulated by a temperature parameter $\kappa$. This thresholding mechanism allows Falcon to find the best action priors.

\subsection{Hyperparameters of D3P}
The D3P training hyperparameters are provided in \Cref{tab:d3p_hpshare,tab:d3p_hpsep}. The settings for the base policy are based on those from DPPO. In contrast, for the adaptor, we keep most hyperparameters unchanged but specifically tune several key parameters, such as $\zeta_1, \zeta_2$ and the reward weight $\beta$.

\begin{table}[H]
    \centering
    \begin{tabular}{lc|lc}
    \toprule
        \multicolumn{2}{c|}{Fine-tuning Base policy $\pi_\theta$} & \multicolumn{2}{c}{Training Adaptor $K_\omega$}\\
        Hyperparameter & Value & Hyperparameter & Value \\
        \midrule
        Number of parallel environments & 40 & Condition horizon $T_o$ & 1 \\
        Condition horizon $T_o$ & 1 & Denoising step discount $\gamma_s$ & 0.95\\
        Reward discount factor $\gamma_{\text{ENV}}$ & 0.999 & reward discount factor $\gamma$ & 0.99\\
        Advantage discount factor $\gamma_{\text{DENOISE}}$ & 0.99 & GAE $\lambda$ & 0.95\\
        GAE $\lambda$ & 0.95 & Optimizer & AdamW\\
        Optimizer & AdamW & Adaptor weight decay & 1e-3\\
        Actor learning rate & 1e-4 & Batch size & 40000\\
        Actor weight decay & 0 & Value loss coefficient & 1.0\\
        Critic learning rate & 1e-3 & Entropy loss coefficient & 0.01\\
        Critic weight decay & 0 & Clip coefficient & 0.01\\
        Batch size & 10000 & Max gradient norm & 10.0\\
        Value loss coefficient & 0.5 & Update epoch & 10\\ 
        Entropy loss coefficient & 0 & Rollout steps & 400\\
        Clip coefficient base $\epsilon_{\text{base}}$ & 0.001 & Reward weight $\alpha$ & 1.0\\
        Clip coefficient $\epsilon_{\text{coef}}$ & 0.01 \\
        Clip coefficient rate $\epsilon_{\text{rate}}$ & 3 \\
        Max gradient norm & 10.0\\
        Rollout steps & 400 \\
        Denoising $\eta$ & 1.0 \\
        \bottomrule
    \end{tabular}
    \caption{Shared hyperparameters of D3P}
    \label{tab:d3p_hpshare}
\end{table}

\begin{table}[H]
    \centering
    \begin{tabular}{l|cccccc}
        \toprule
        Hyperparameter & Lift & Can & \makecell{Square \\ (State \& Pixel)} & \makecell{Transport \\ (State \& Pixel)} & kitchen-complete-v0 & kitchen-mixed-v0 \\
        Update epoch & 10 & 10 & 10 & 6 & 10 & 10 \\
        Adaptor learning rate & 1e-4 & 1e-4 & 1e-4 & 3e-5 & 1e-4 & 1e-4 \\
        Threshold $\zeta_1$ & 100.0 & 170.0 & 210.0 & 310.0 & 3.3 & 3.3\\
        Threshold $\zeta_2$ & 4.0 & 4.0 & 5.0 & 7.5 & 5.0 & 5.0 \\
        Reward weight $\beta$ & 0.2 & 0.2 & 0.06 & 0.1 & 0.4 & 0.4 \\
        \bottomrule
    \end{tabular}
    \caption{Task-specific hyperparameters of D3P}
    \label{tab:d3p_hpsep}
\end{table}

\subsection{Computation}
All experiments are conducted on our server cluster running Ubuntu 22.04. The training environment is built on Python 3.8, with specific dependencies listed in \Cref{lst:dep}. The hardware resources utilized for training are detailed in \Cref{tab:comp_infra}. As the Robomimic and Franka Kitchen simulation environments are primarily CPU-intensive, our training process consumes significant CPU and memory resources. Each training run is performed using a single GPU.

\begin{table}[H]
    \centering
    \begin{tabular}{cc|ccc}
        \toprule
        ~ & Environment & CPU & Memory & GPU \\
        \midrule
        \multirow{3}{*}{Training} & Robomimic (State) & 45 Core & 128G & RTX 3090 24G \\
        ~ & Robomimic (Pixel) & 45 Core & 256G & A800 40G \\
        ~ & Franka Kitchen & 10 Core & 64G & RTX 3090 24G \\
        Evaluation & All environment & 16 Core & 32G & RTX 3090 24G\\
        \bottomrule
    \end{tabular}
    \caption{Computing resources for training D3P}
    \label{tab:comp_infra}
\end{table}

\vspace{3mm}
\begin{lstlisting}[language=Bash, caption={Dependencies for training D3P\label{lst:dep}}]
av==12.3.0
einops==0.8.0
gdown==5.2.0
gym==0.22.0
hydra-core==1.3.2
imageio==2.35.1
matplotlib==3.7.5
omegaconf==2.3.0
pretty_errors==1.2.25
torch==2.4.0
tqdm==4.66.5
wandb==0.17.7

# for robomimic environment
cython<3
d4rl
patchelf
mujoco==3.1.6
robomimic
robosuite @ v1.4.1

# for franka kitchen
cython<3
d4rl
dm_control==1.0.16
mujoco==3.1.6
patchelf
\end{lstlisting}

\newpage
\section{Additional Experiment Results}
\subsection{Performance Comparison}
\begin{figure}
  \centering 
  \subfloat[Lift (State) - Success Rate]{%
    \includegraphics[width=0.245\textwidth]{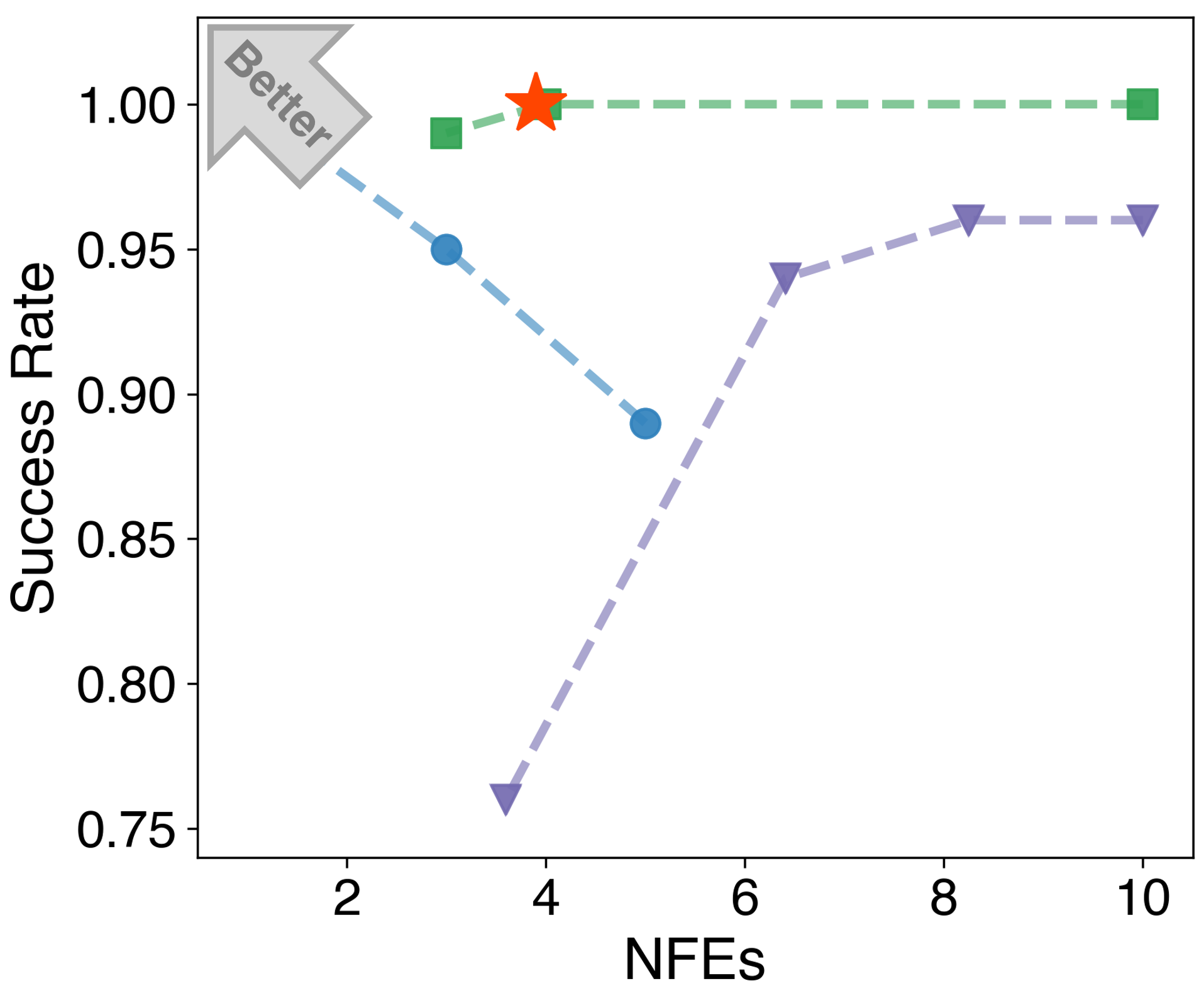}%
    \label{fig:lift} 
  }
  \subfloat[Can (State) - Success Rate]{%
    \includegraphics[width=0.245\textwidth]{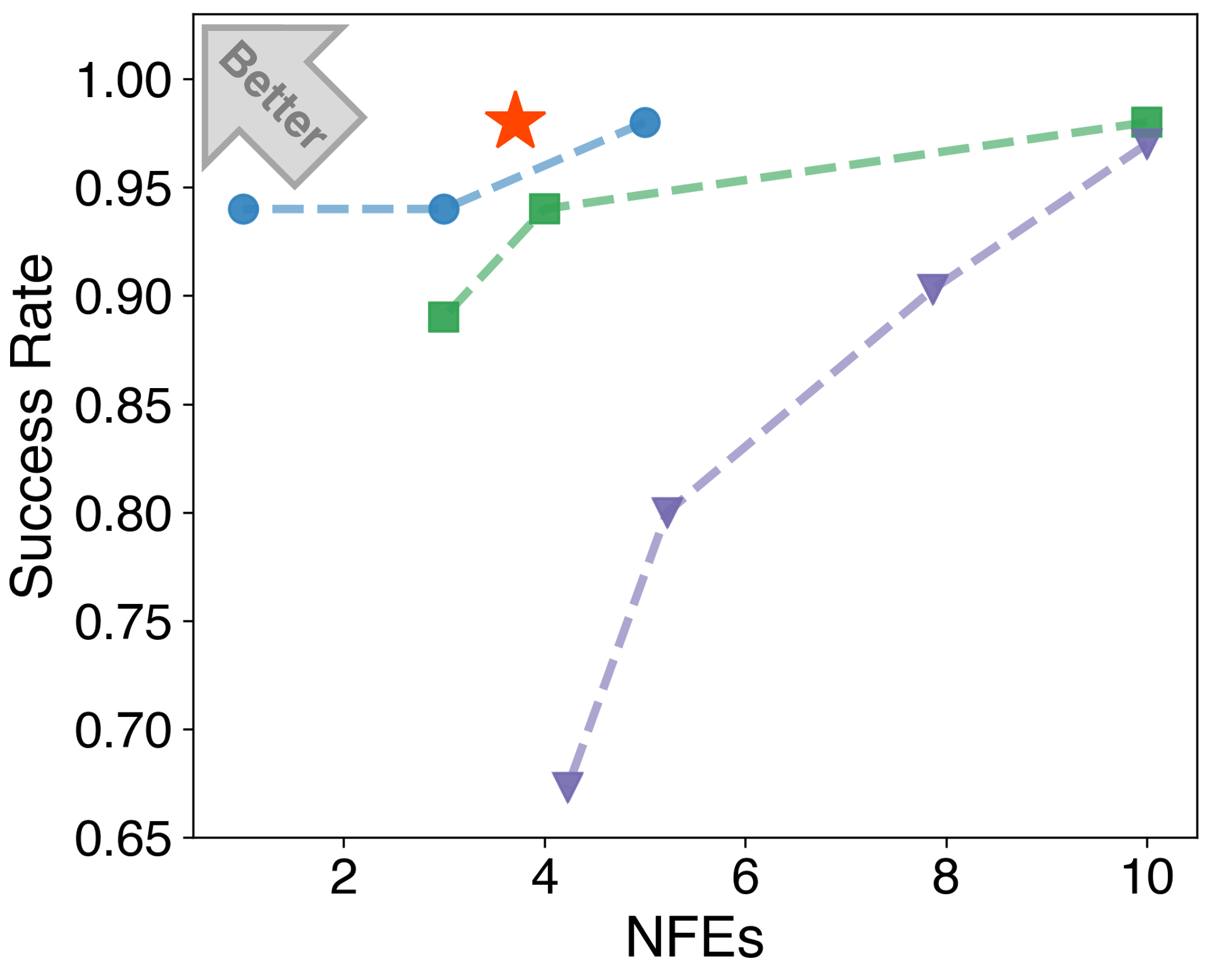}%
    \label{fig:can} 
  }
  \subfloat[Square (State) - Success Rate]{%
    \includegraphics[width=0.245\textwidth]{figs/square_sr.png}%
    \label{fig:square} 
  }
  \subfloat[Square (Pixel) - Success Rate]{%
    \includegraphics[width=0.245\textwidth]{figs/square_img_sr.png}%
    \label{fig:square-img} 
  }

  \subfloat[Transport (State) \\ - Success Rate]{%
    \includegraphics[width=0.245\textwidth]{figs/transport_sr.png}%
    \label{fig:transport} 
  }
  \subfloat[Transport (Pixel) \\ - Success Rate]{%
    \includegraphics[width=0.245\textwidth]{figs/transport_img_sr.png}%
    \label{fig:transport-img} 
  }
  \subfloat[Kitchen-complete-v0\\ - Success Rate]{%
    \includegraphics[width=0.245\textwidth]{figs/kitchen_complete_sr.png}%
    \label{fig:kitchen-complete} 
  }
  \subfloat[Kitchen-mixed-v0\\ - Success Rate]{%
    \includegraphics[width=0.245\textwidth]{figs/kitchen_mixed_sr.png}%
    \label{fig:kitchen-mixed} 
  }

  \subfloat[Lift (State) - Return]{%
    \includegraphics[width=0.245\textwidth]{figs/lift.png}%
    \label{fig:lift} 
  }
  \subfloat[Can (State) - Return]{%
    \includegraphics[width=0.245\textwidth]{figs/can.png}%
    \label{fig:can} 
  }
  \subfloat[Square (State) - Return]{%
    \includegraphics[width=0.245\textwidth]{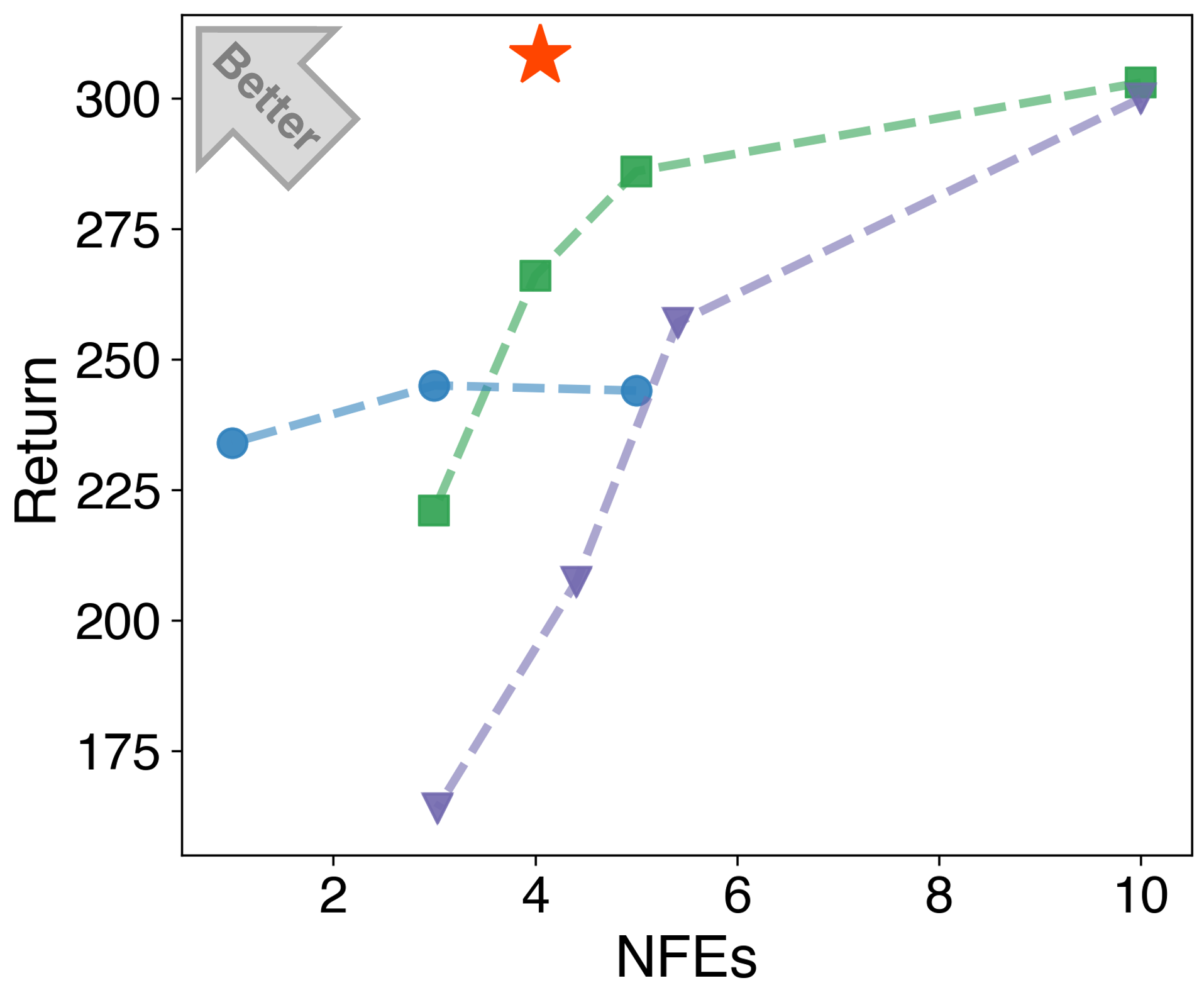}%
    \label{fig:square} 
  }
  \subfloat[Square (Pixel) - Return]{%
    \includegraphics[width=0.245\textwidth]{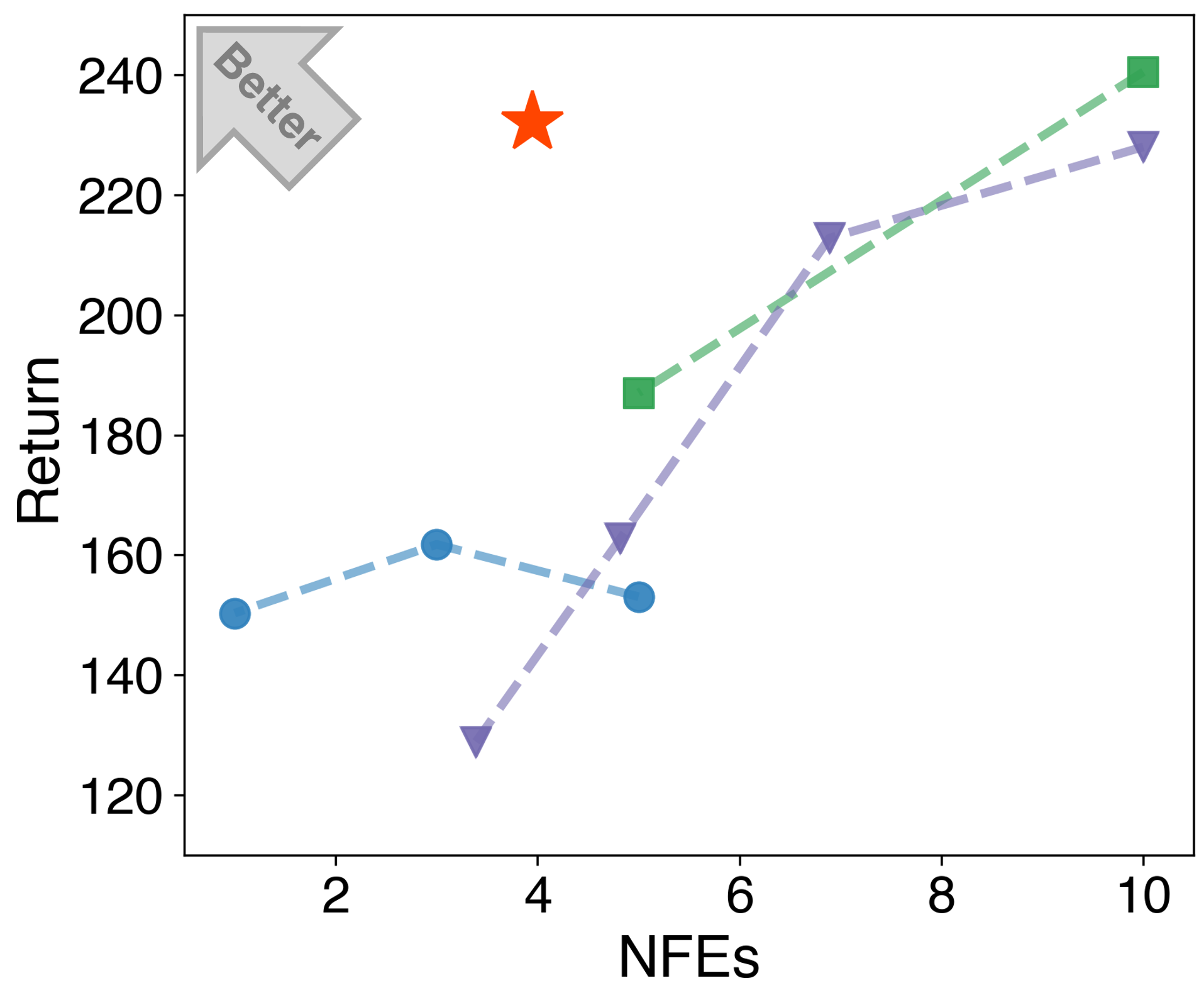}%
    \label{fig:square-img} 
  }

  \subfloat[Transport (State) \\ - Return]{%
    \includegraphics[width=0.245\textwidth]{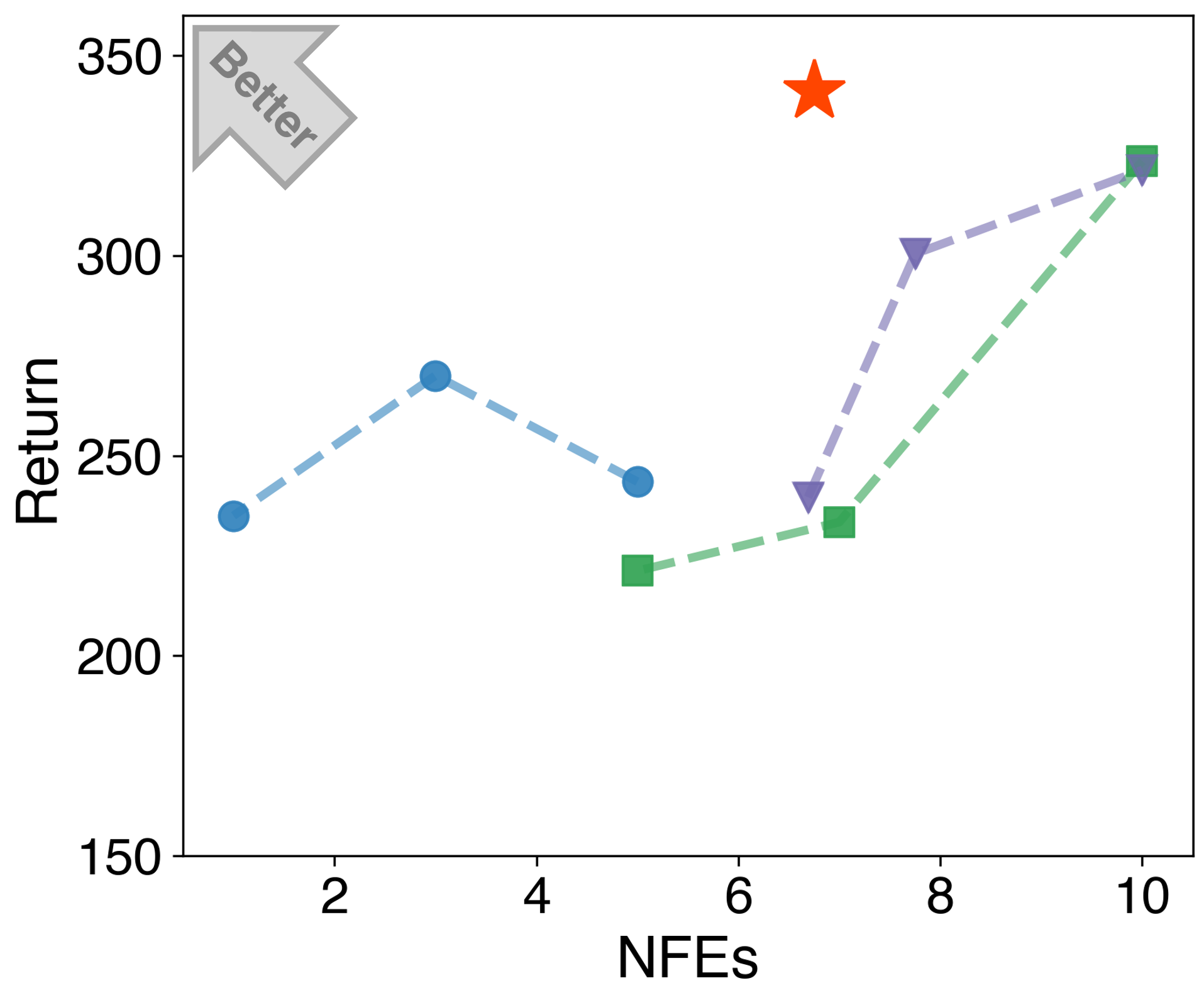}%
    \label{fig:transport} 
  }
  \subfloat[Transport (Pixel) \\ - Return]{%
    \includegraphics[width=0.245\textwidth]{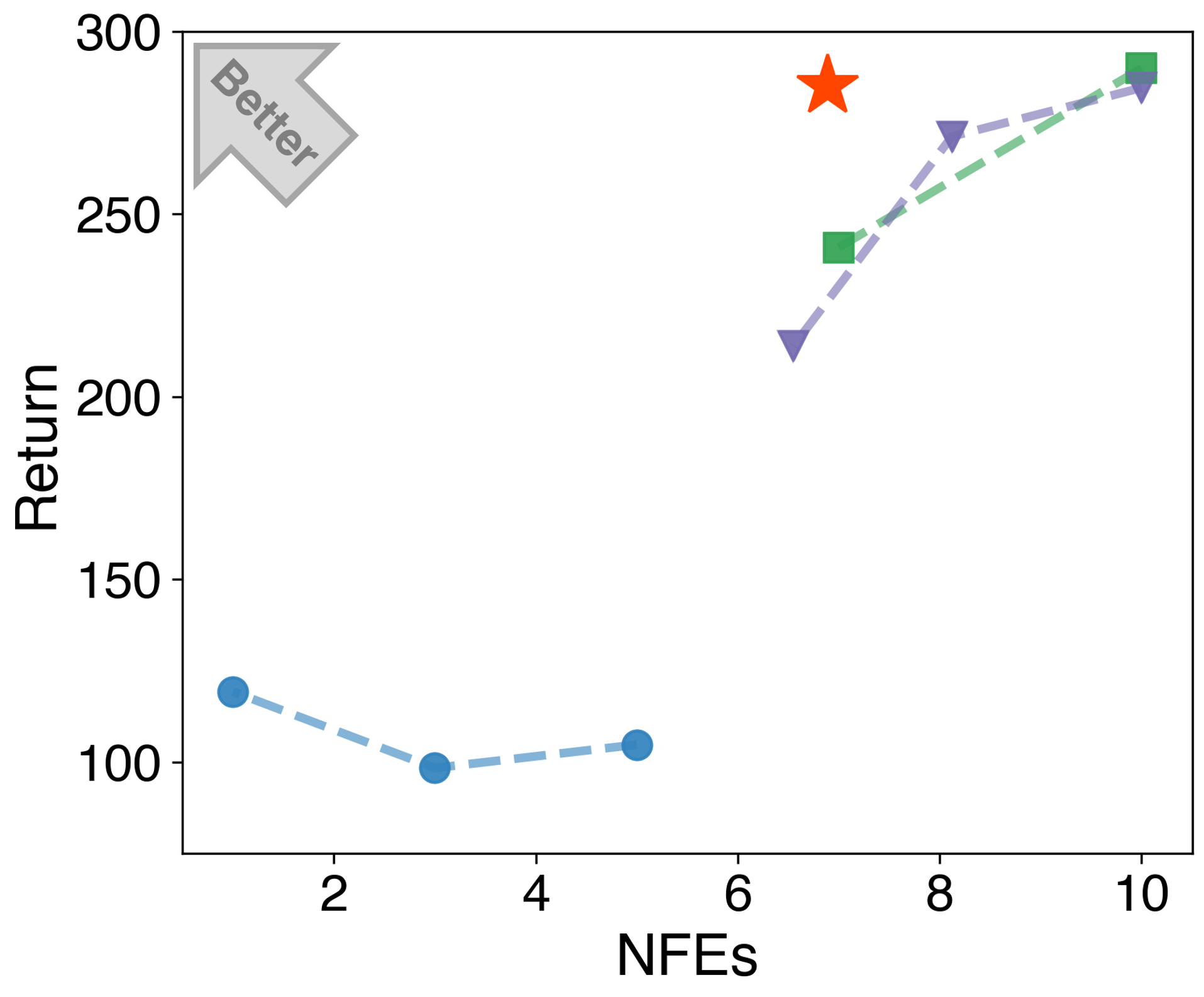}%
    \label{fig:transport-img} 
  }
  \subfloat[Kitchen-complete-v0\\ - Return]{%
    \includegraphics[width=0.245\textwidth]{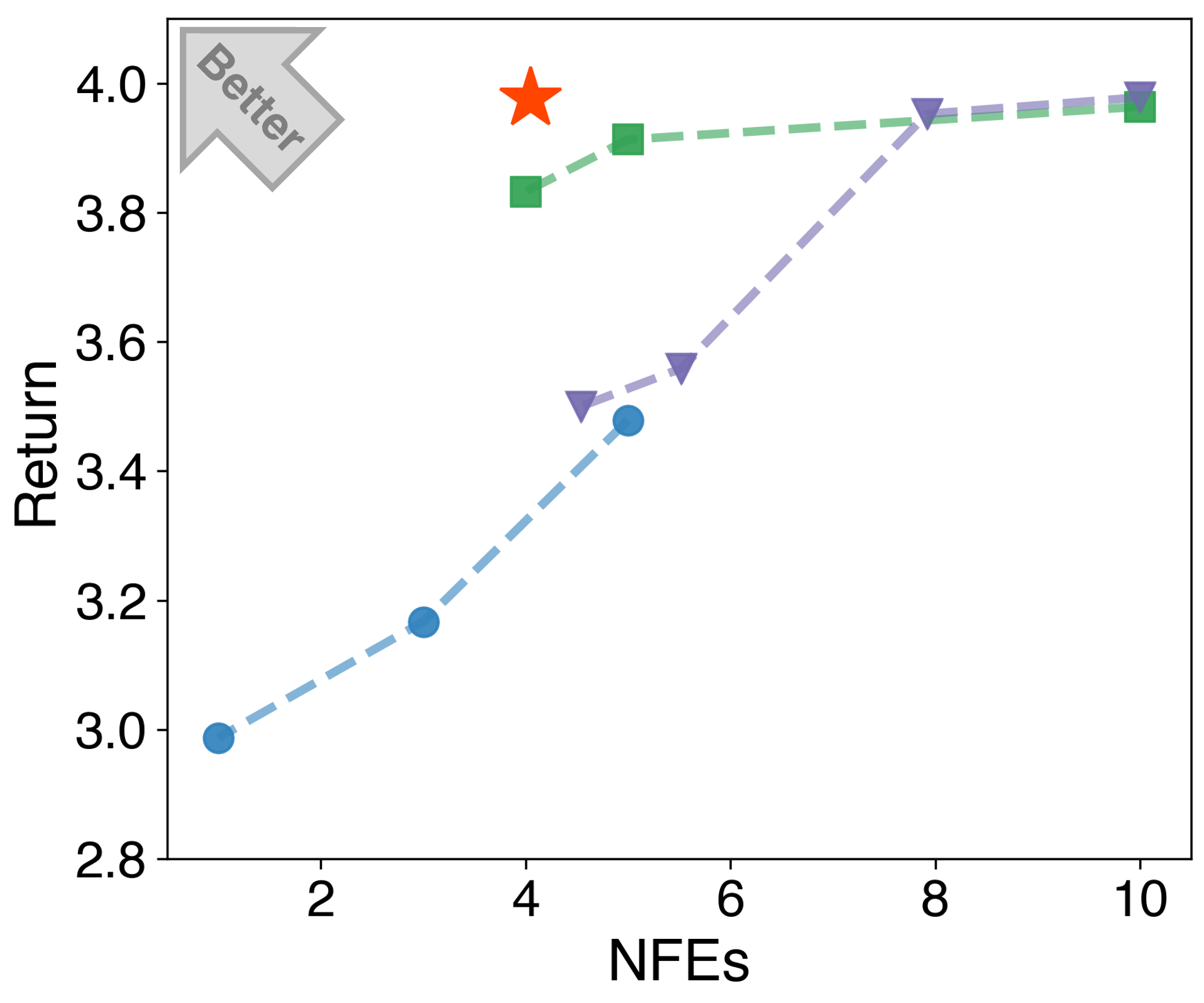}%
    \label{fig:kitchen-complete} 
  }
  \subfloat[Kitchen-mixed-v0\\ - Return]{%
    \includegraphics[width=0.245\textwidth]{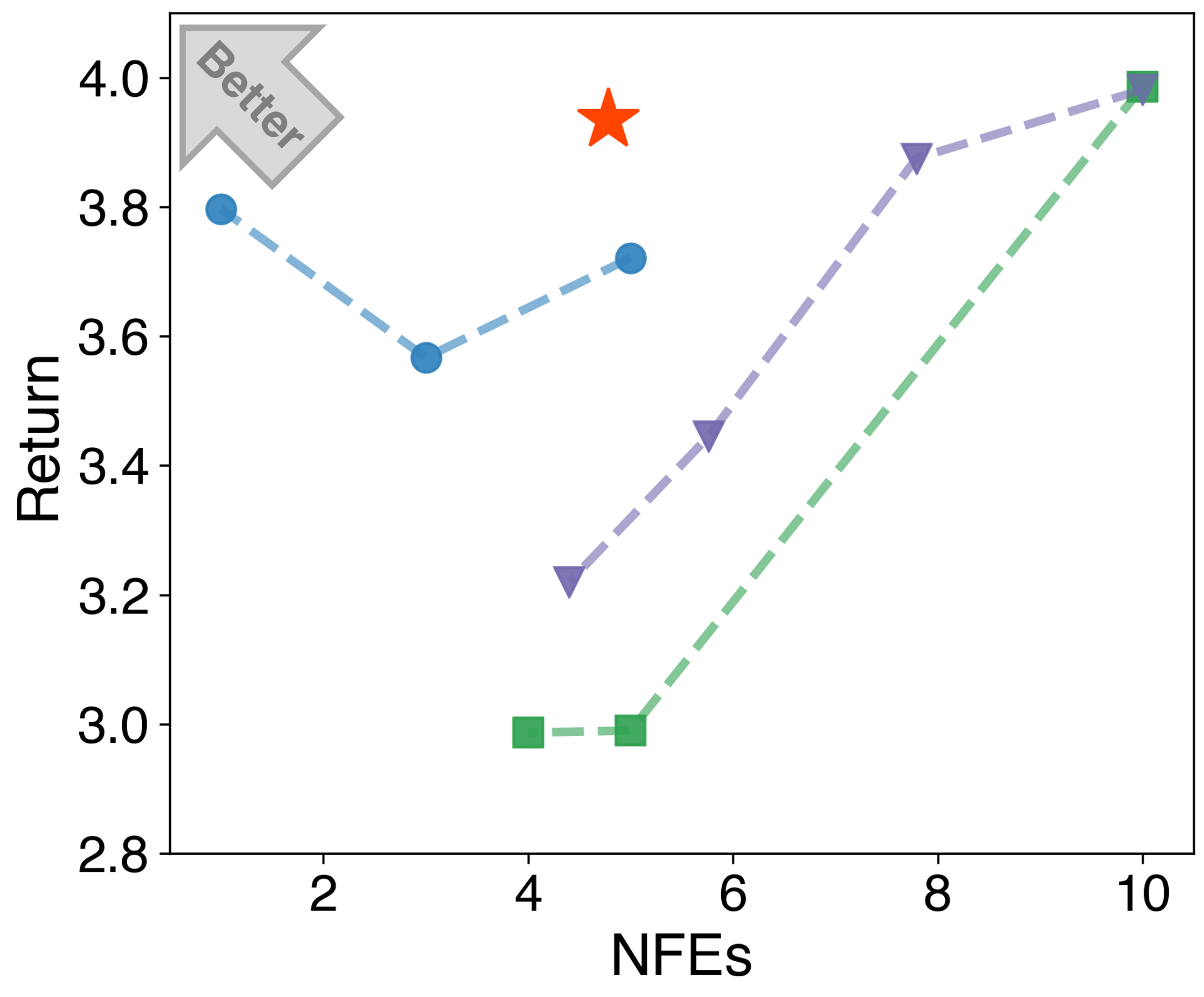}%
    \label{fig:kitchen-mixed} 
  }

  \subfloat{
  \includegraphics[width = 0.5\textwidth]{figs/legend.png}
  }
  \caption{We plot success rate and episodic return against the NFE. An ideal method occupies the upper-left corner, representing high performance at a low inference cost. D3P matches or surpasses the peak performance of the 10-step DPPO baseline while achieving an average 2.2$\times$ speed-up. All results are averaged over 5 seeds.}
  \label{fig:full_res}
\end{figure}

\Cref{fig:full_res} presents the success rate and return versus Number of Function Evaluations (NFE) for all methods. In these plots, the upper-left corner represents the ideal trade-off: high performance with low inference cost. With an equivalent training budget, D3P consistently matches or surpasses the fixed 10-step diffusion policy baseline across both metrics.

The performance of the DPPO baseline correlates directly with its inference cost, dropping significantly as NFE is reduced. The two acceleration baselines, CP and Falcon, exhibit distinct features. CP excels at few-step inference, outperforming Falcon on tasks like \texttt{Lift}, \texttt{Can}, and \texttt{Square} (State). In contrast, Falcon, a training-free accelerator, achieves a better performance-efficiency trade-off on more complex tasks such as \texttt{Square} (Pixel) and \texttt{Transport} (Pixel).

\begin{figure}
  \centering 
  \subfloat[Lift (State) - Success Rate]{%
    \includegraphics[width=0.245\textwidth]{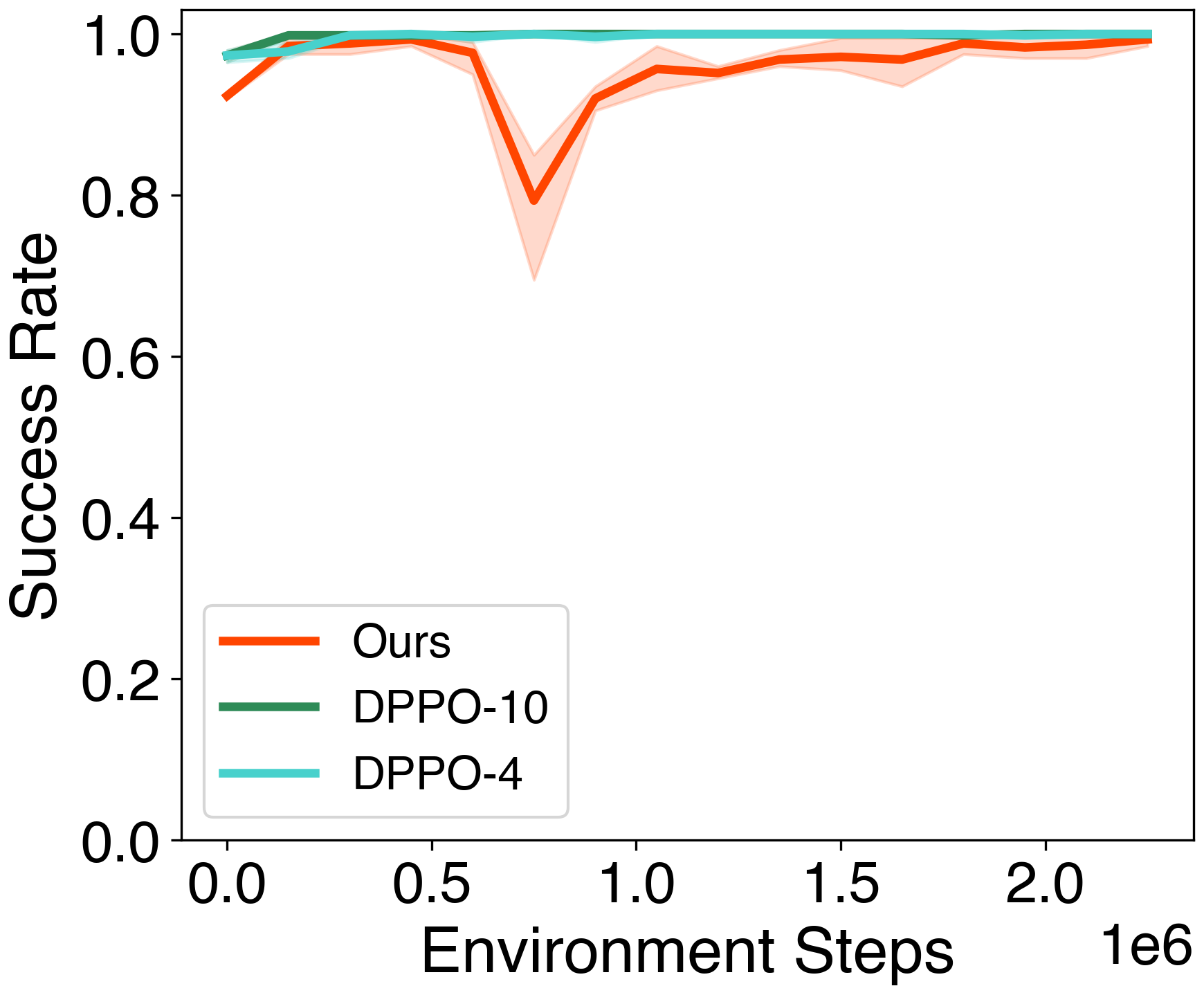}%
    \label{fig:train-lift} 
  }
  \subfloat[Can (State) - Success Rate]{%
    \includegraphics[width=0.245\textwidth]{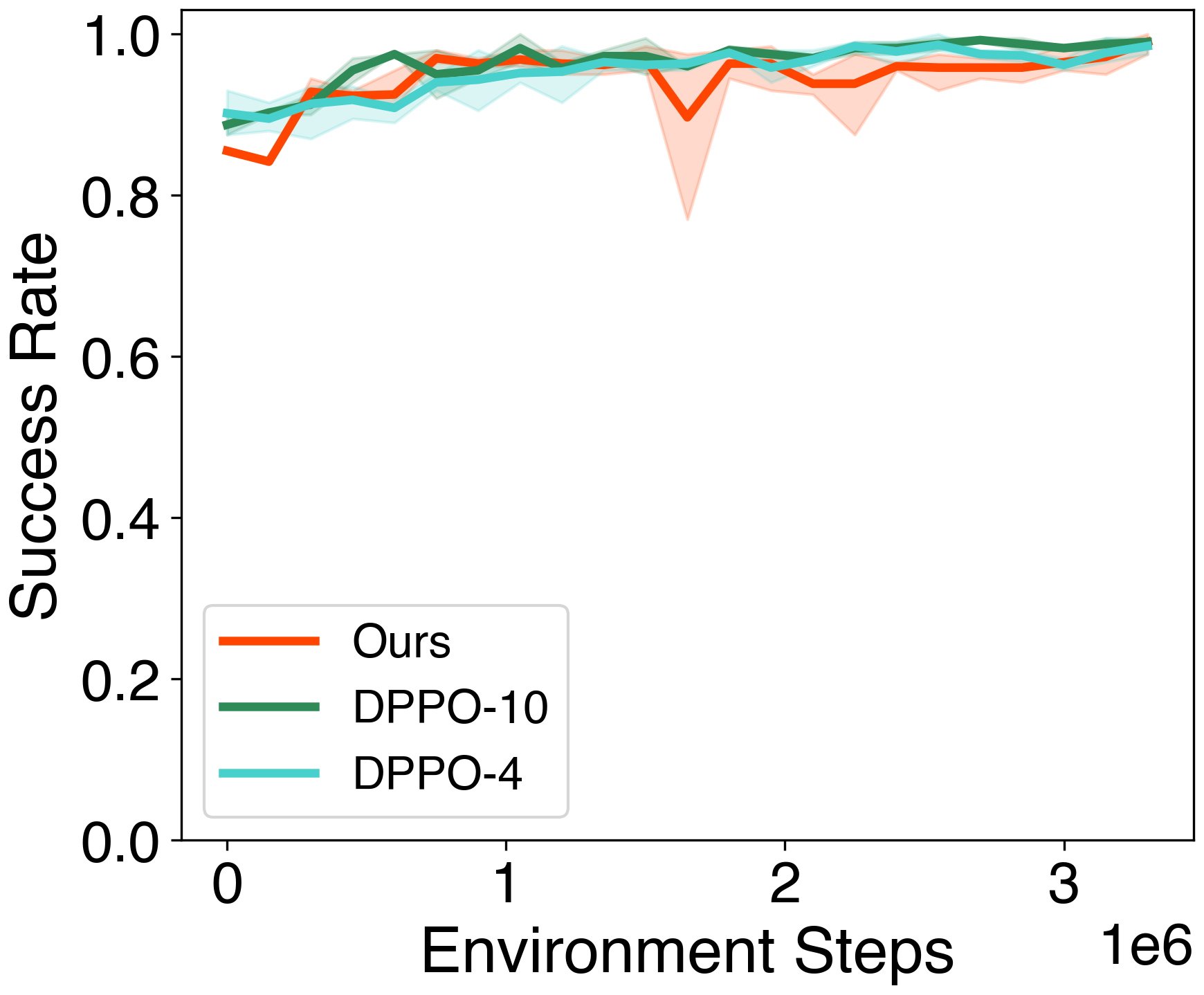}%
    \label{fig:train-can} 
  }
  \subfloat[Square (State) - Success Rate]{%
    \includegraphics[width=0.245\textwidth]{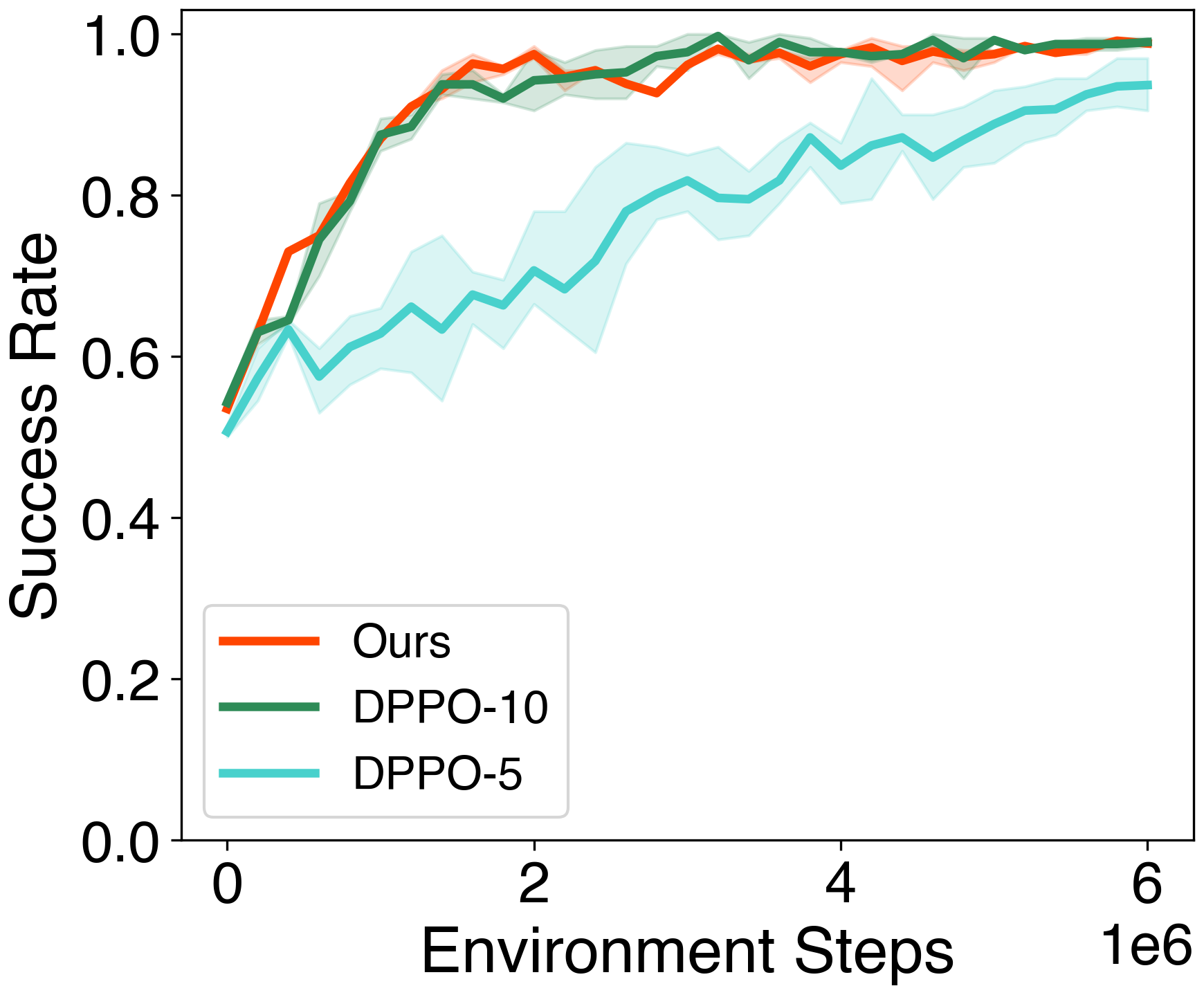}%
    \label{fig:train-square} 
  }
  \subfloat[Square (Pixel) - Success Rate]{%
    \includegraphics[width=0.245\textwidth]{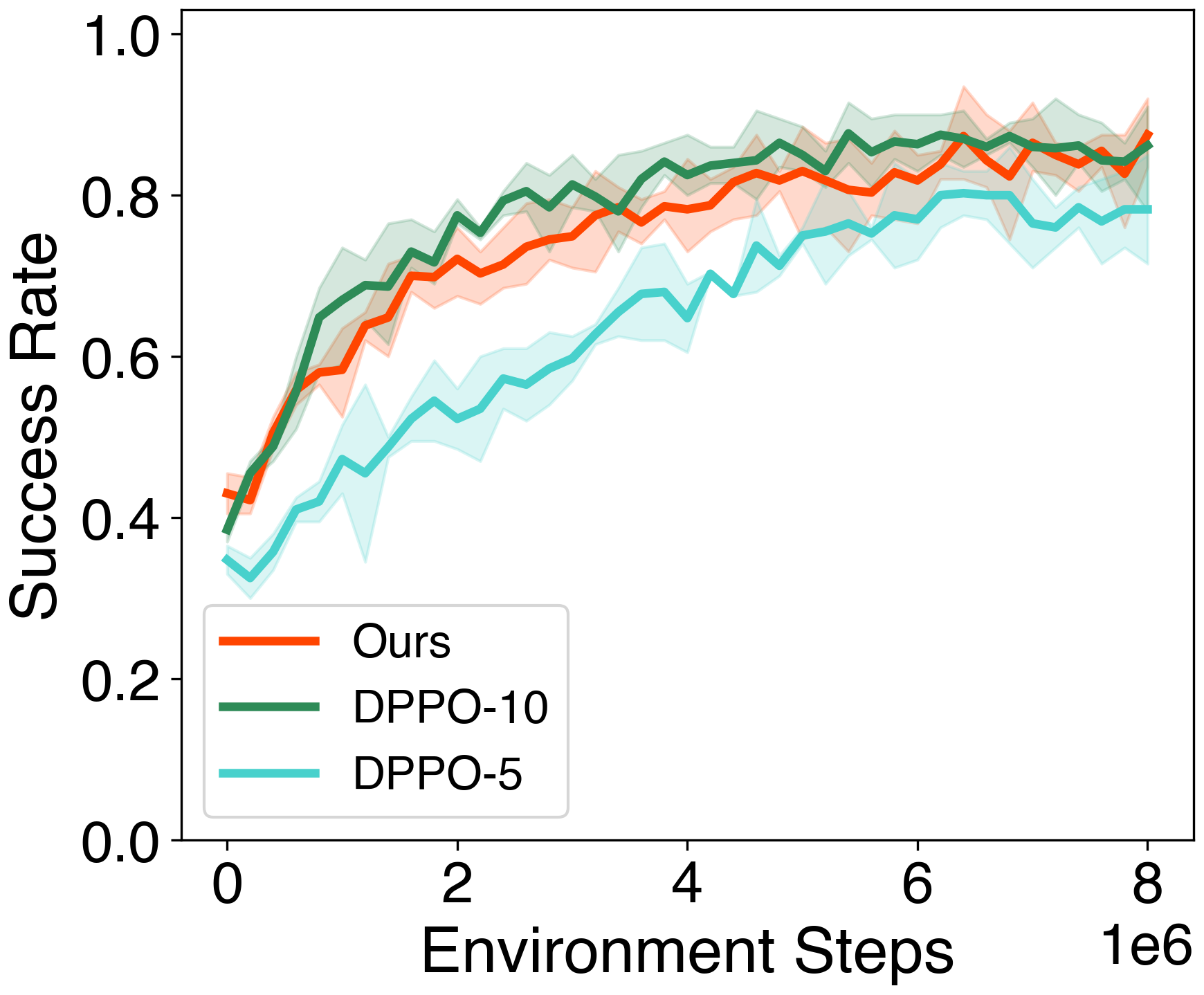}%
    \label{fig:train-square-img} 
  }

  \subfloat[Transport (State) \\ - Success Rate]{%
    \includegraphics[width=0.245\textwidth]{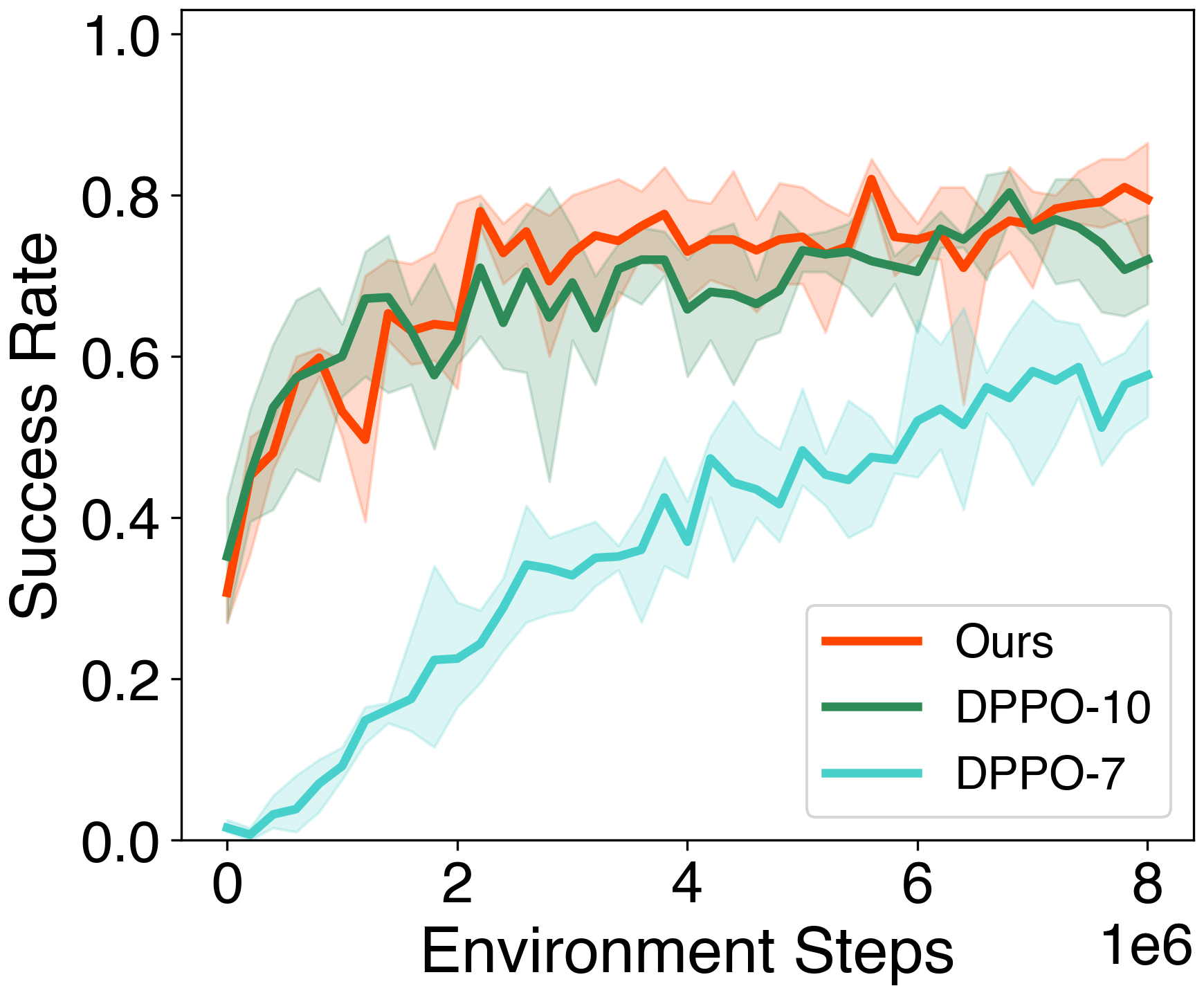}%
    \label{fig:train-transport} 
  }
  \subfloat[Transport (Pixel) \\ - Success Rate]{%
    \includegraphics[width=0.245\textwidth]{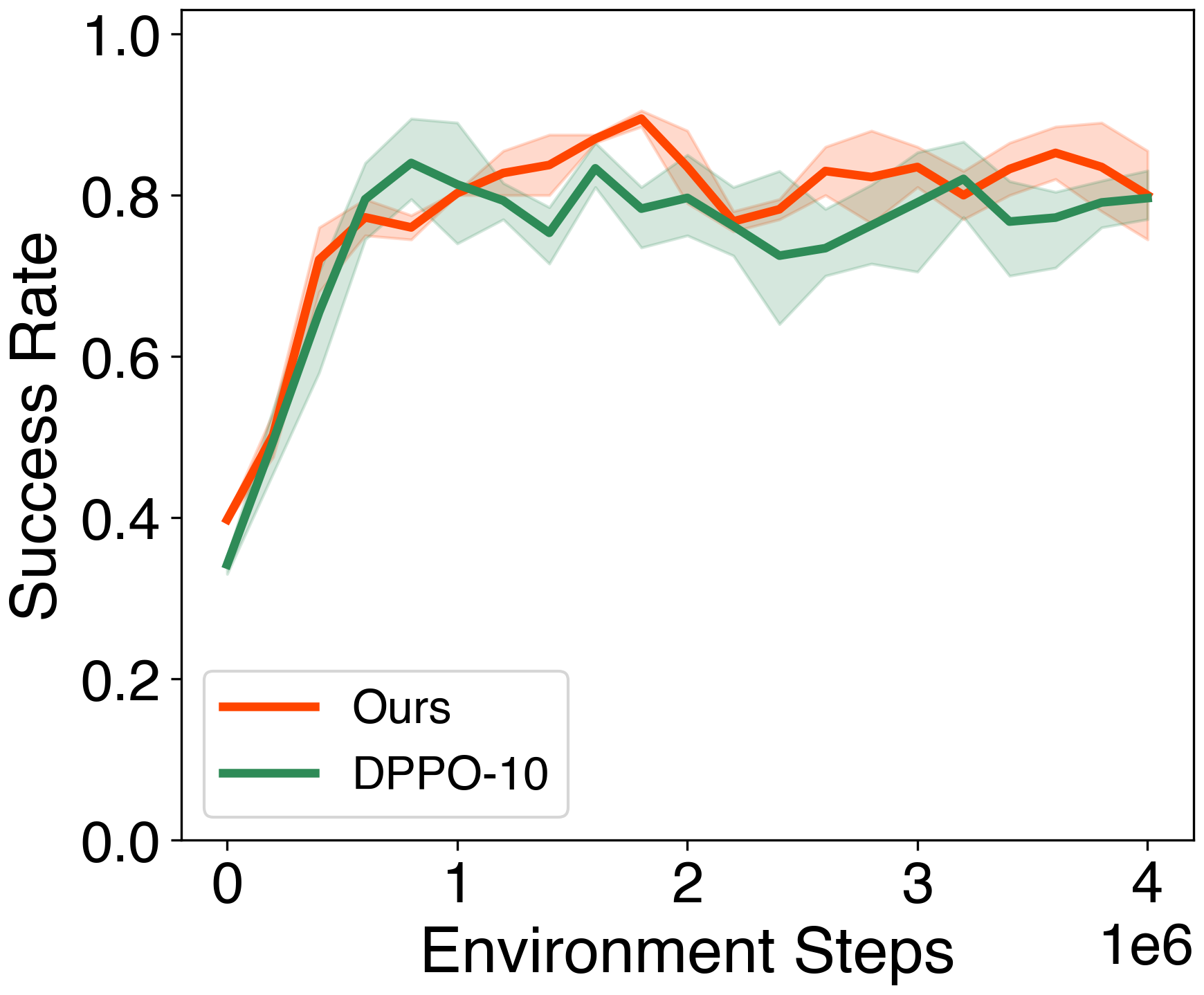}%
    \label{fig:train-transport-img} 
  }
  \subfloat[Kitchen-complete-v0\\ - Success Rate]{%
    \includegraphics[width=0.245\textwidth]{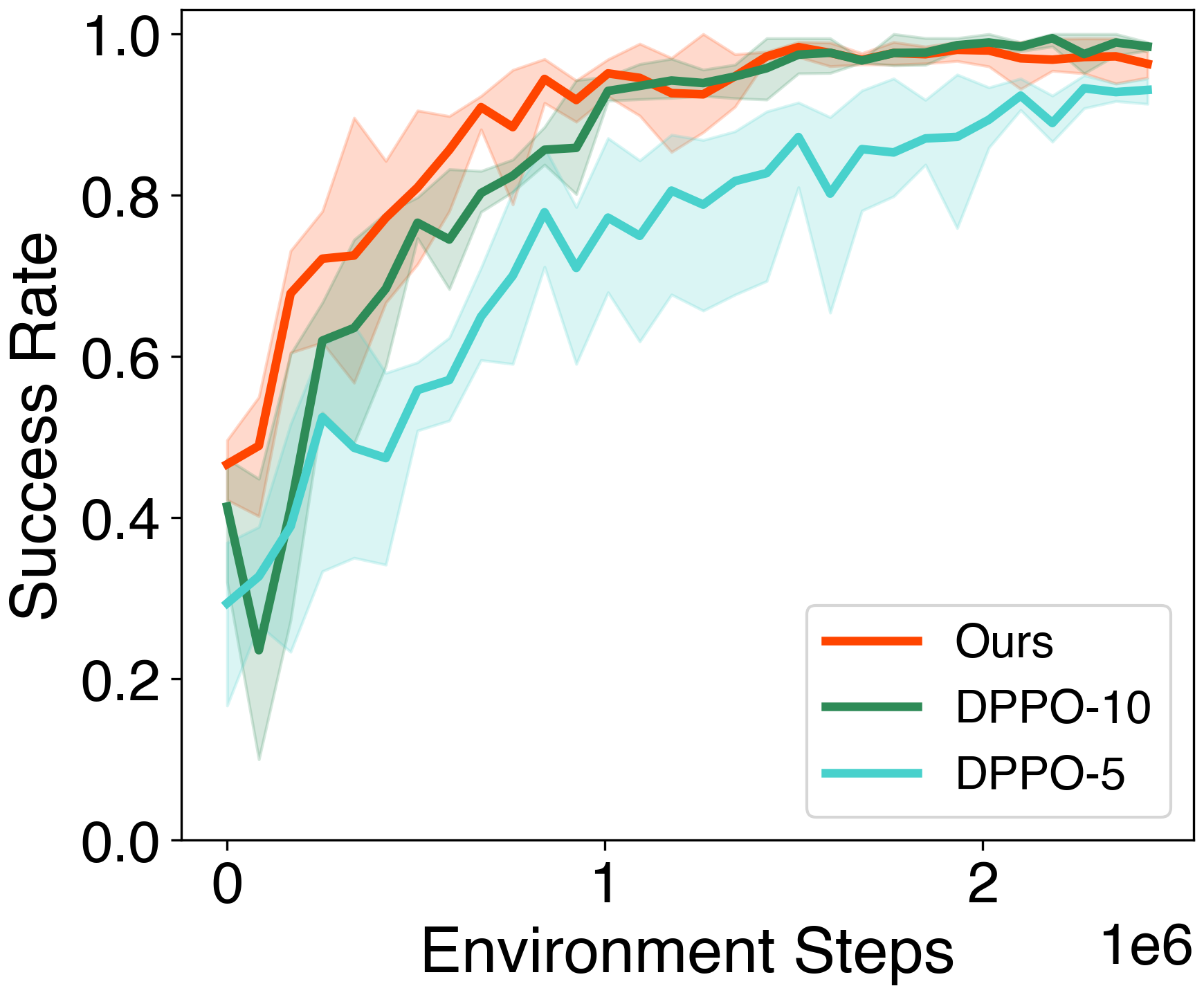}%
    \label{fig:train-kitchen-complete} 
  }
  \subfloat[Kitchen-mixed-v0\\ - Success Rate]{%
    \includegraphics[width=0.245\textwidth]{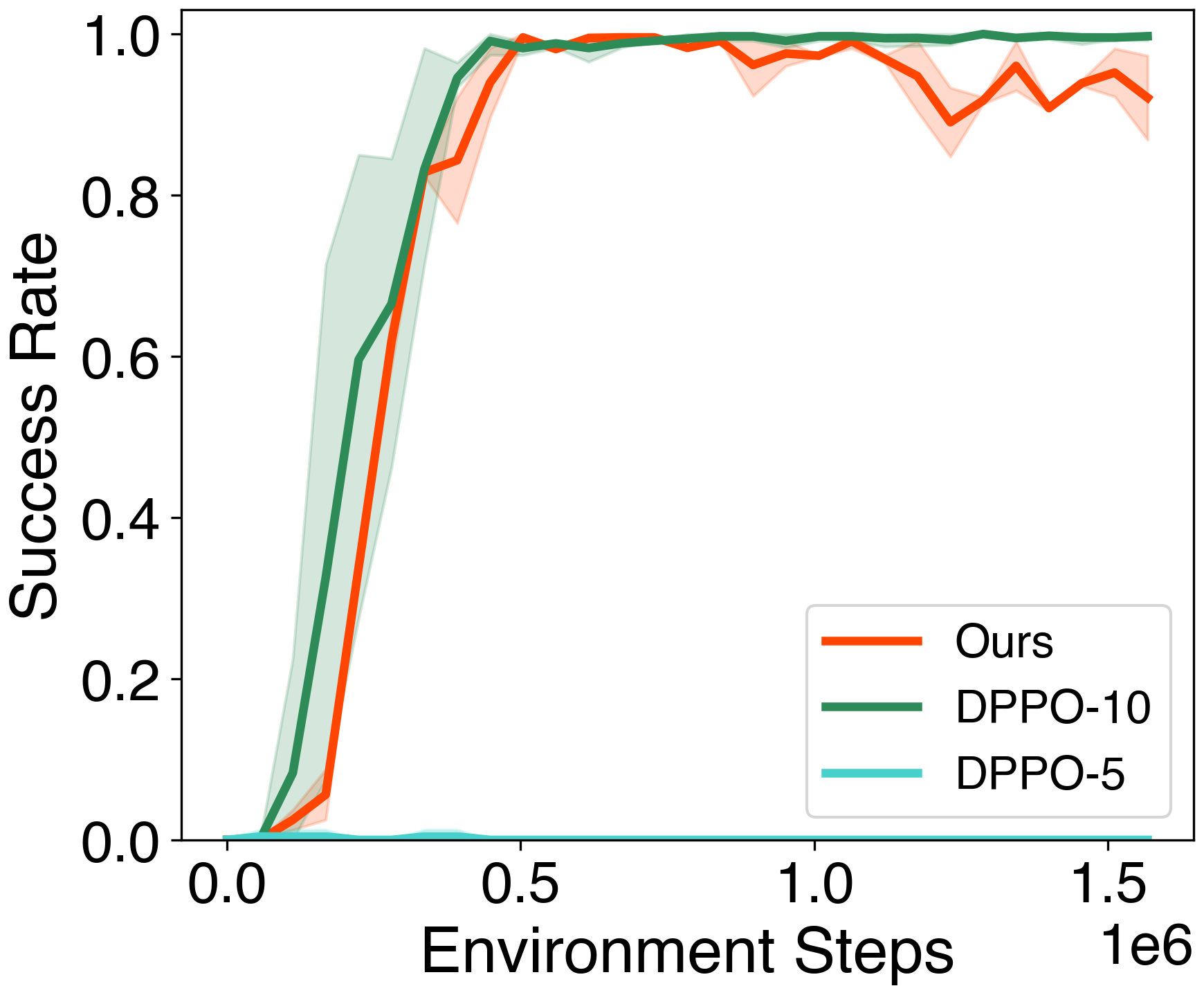}%
    \label{fig:train-kitchen-mixed} 
  }

  \subfloat[Lift (State) - Return]{%
    \includegraphics[width=0.245\textwidth]{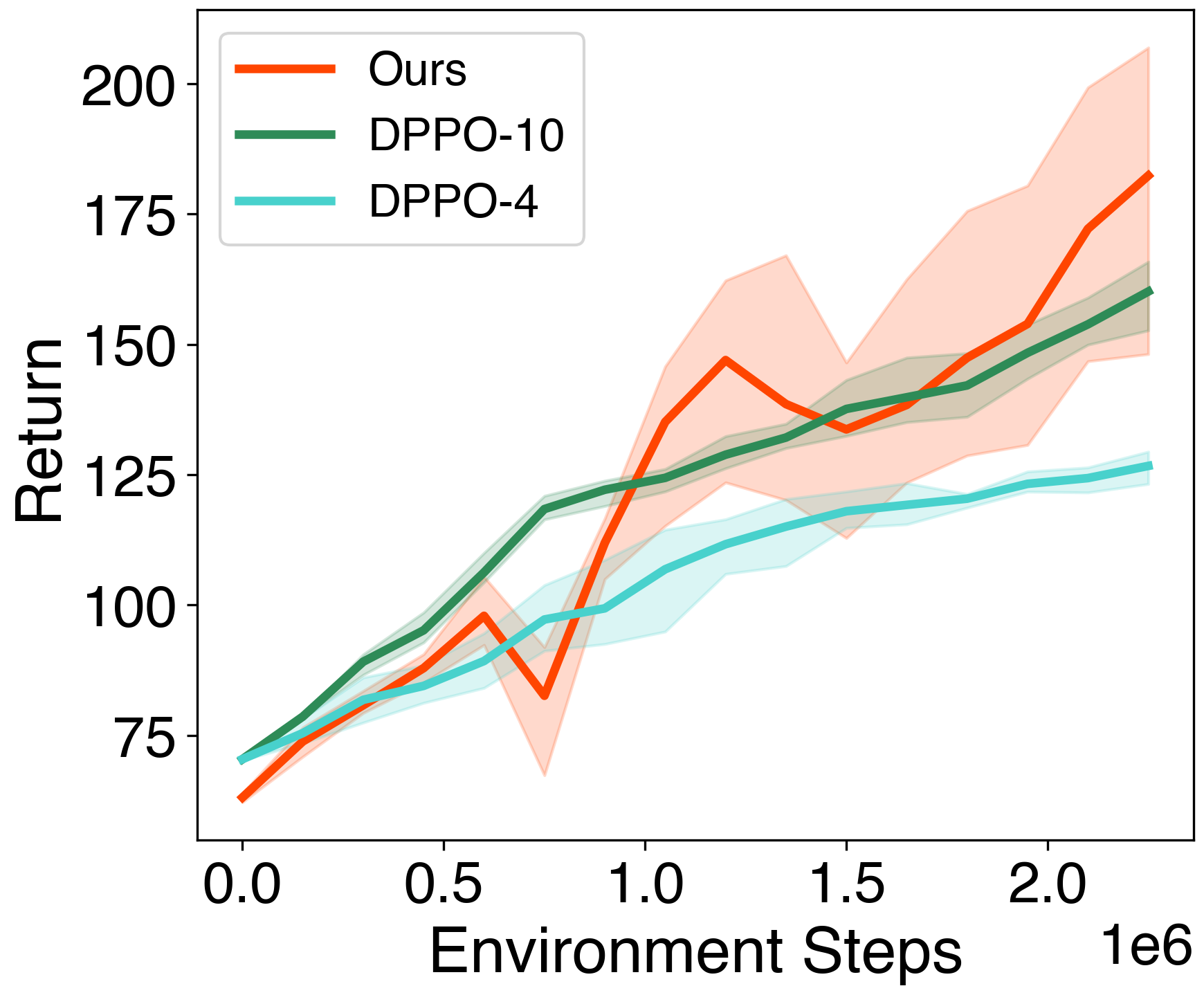}%
    \label{fig:train-lift} 
  }
  \subfloat[Can (State) - Return]{%
    \includegraphics[width=0.245\textwidth]{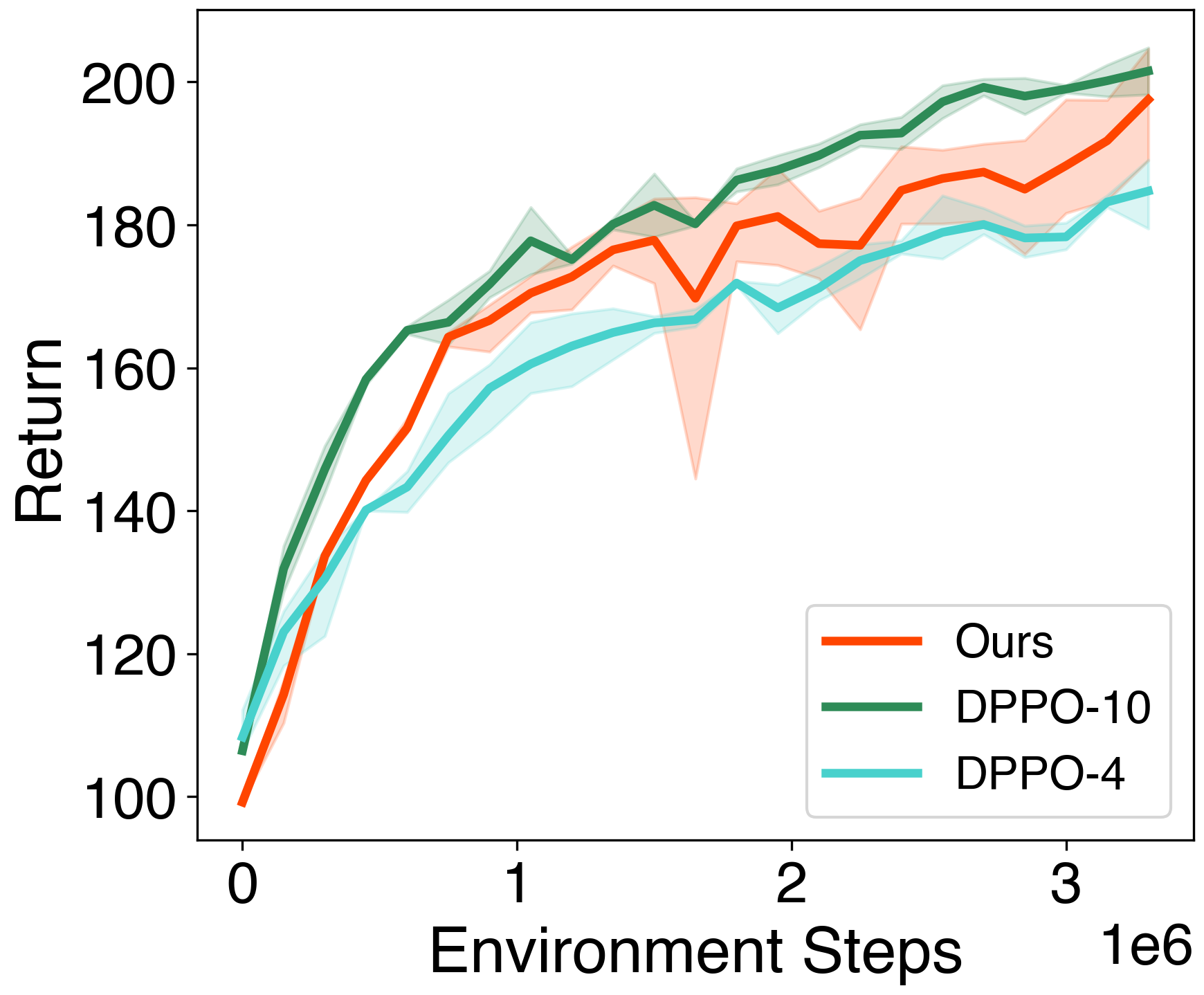}%
    \label{fig:train-can} 
  }
  \subfloat[Square (State) - Return]{%
    \includegraphics[width=0.245\textwidth]{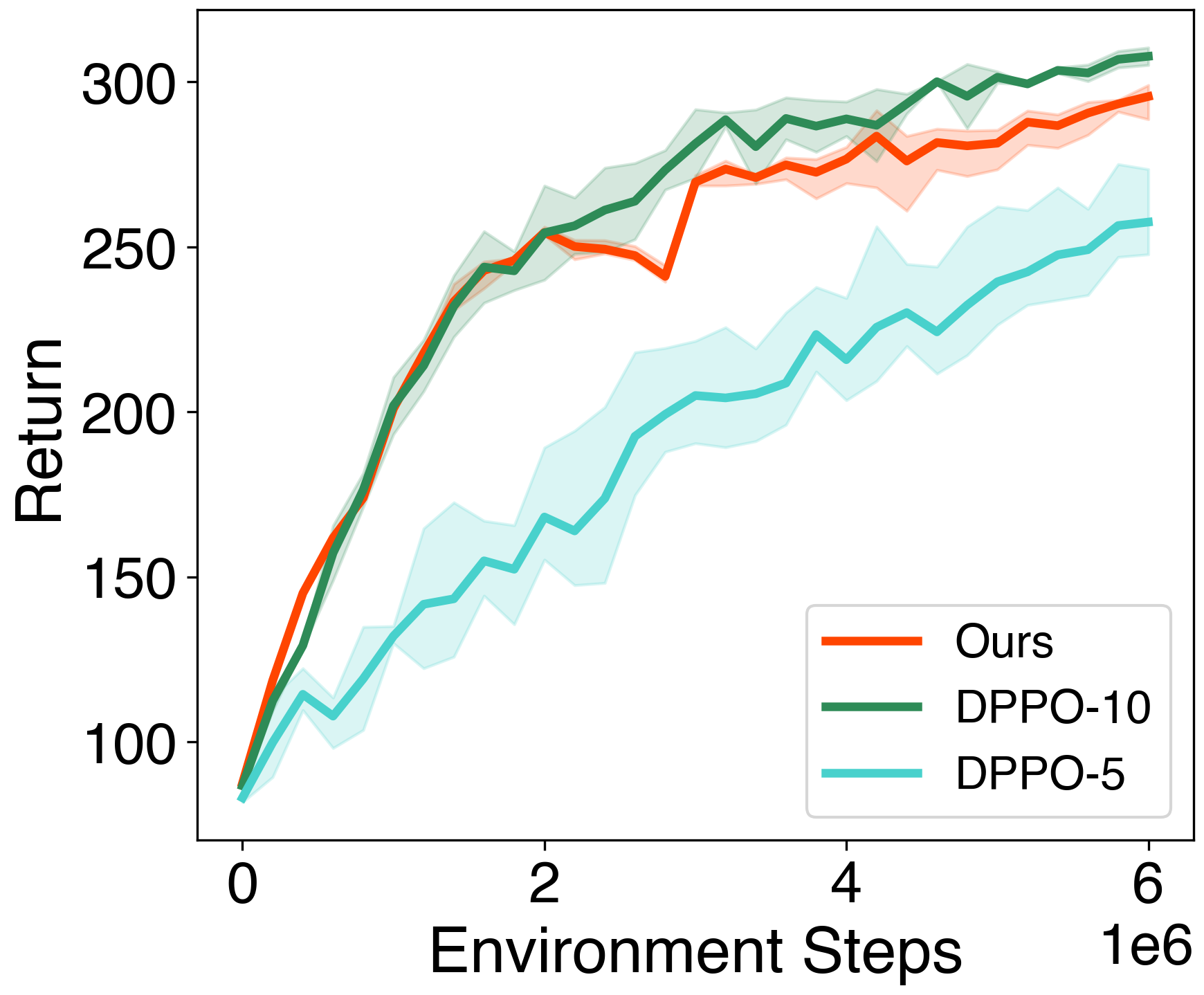}%
    \label{fig:train-square} 
  }
  \subfloat[Square (Pixel) - Return]{%
    \includegraphics[width=0.245\textwidth]{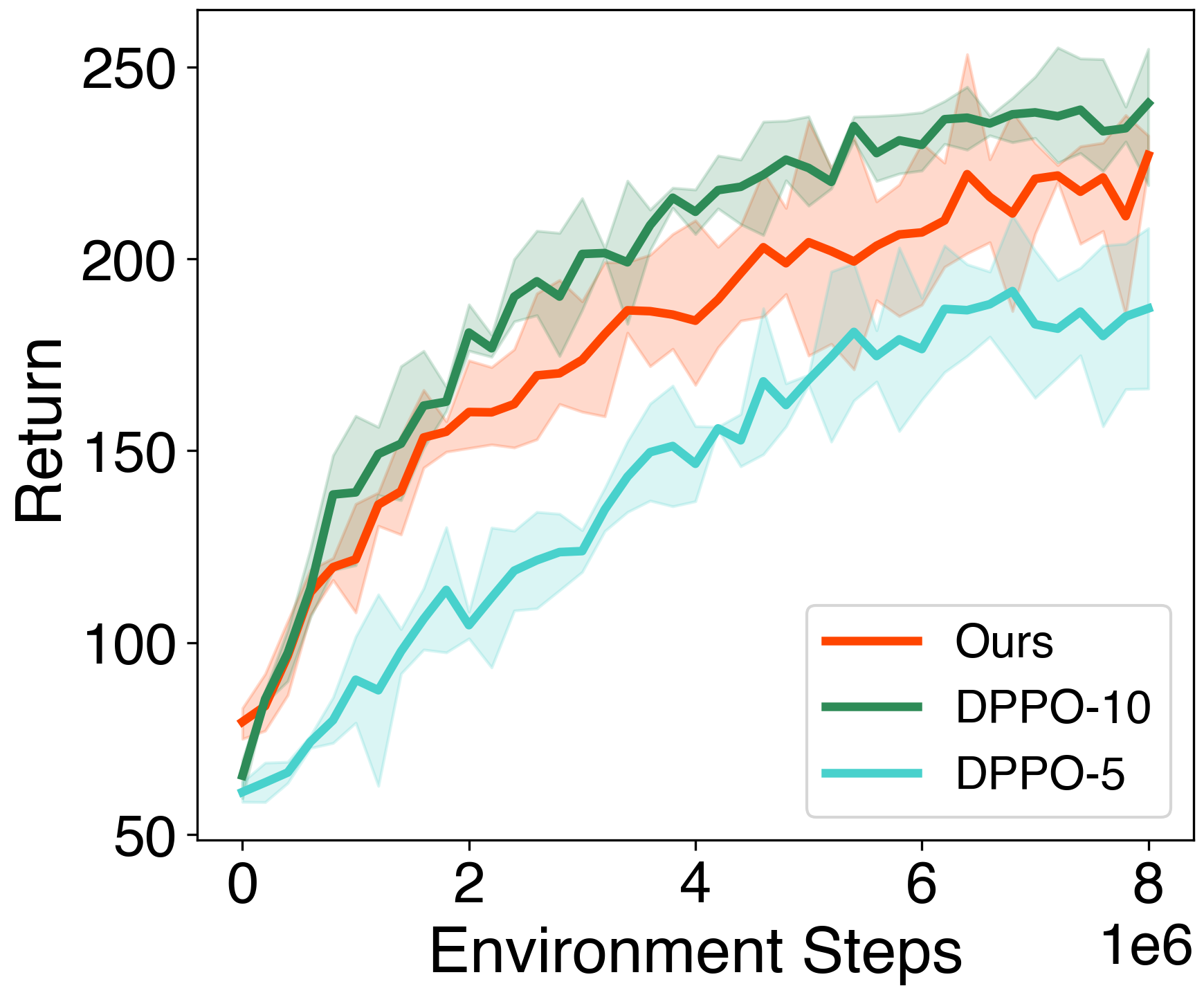}%
    \label{fig:train-square-img} 
  }

  \subfloat[Transport (State) \\ - Return]{%
    \includegraphics[width=0.245\textwidth]{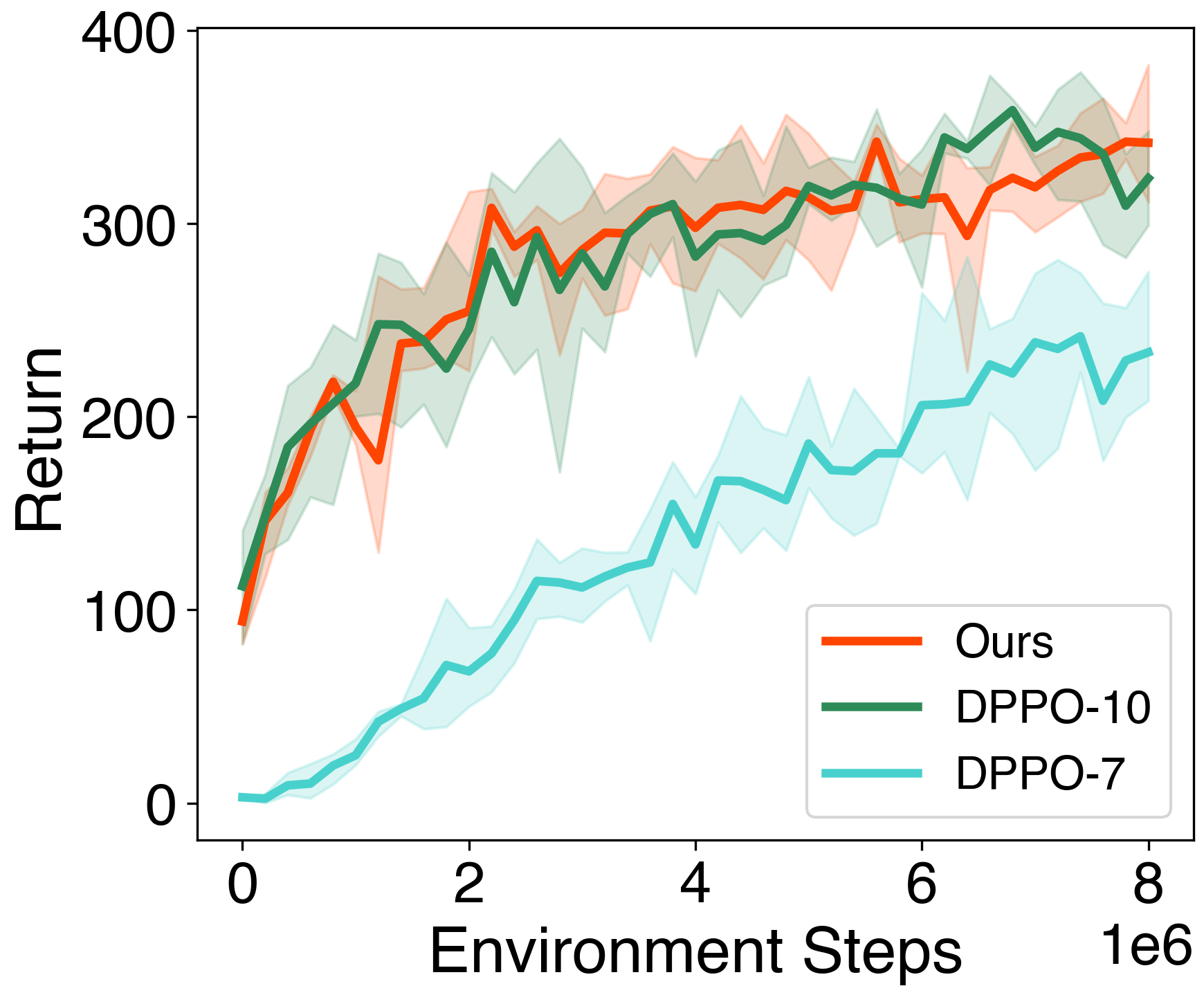}%
    \label{fig:train-transport} 
  }
  \subfloat[Transport (Pixel) \\ - Return]{%
    \includegraphics[width=0.245\textwidth]{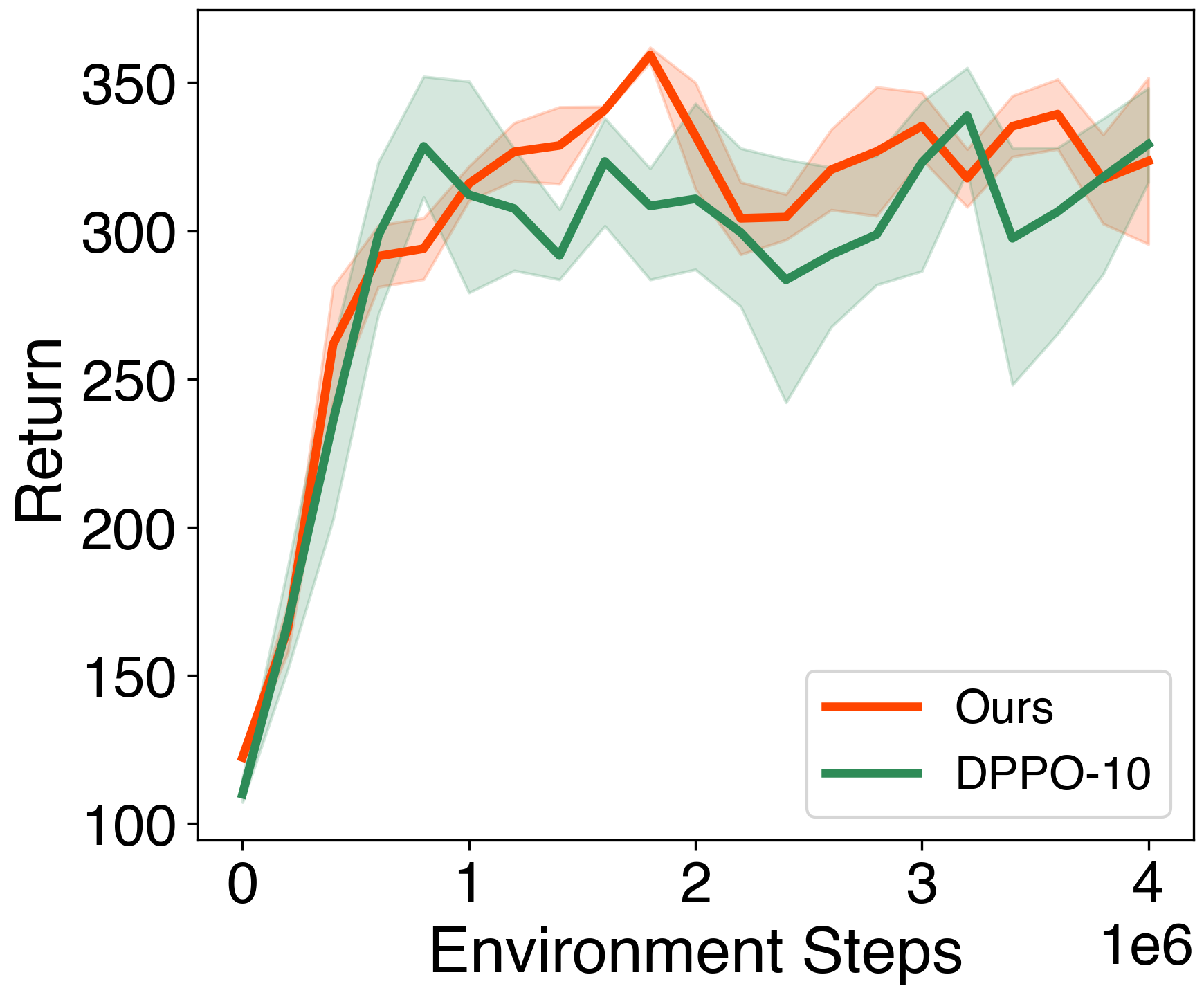}%
    \label{fig:train-transport-img} 
  }
  \subfloat[Kitchen-complete-v0\\ - Return]{%
    \includegraphics[width=0.245\textwidth]{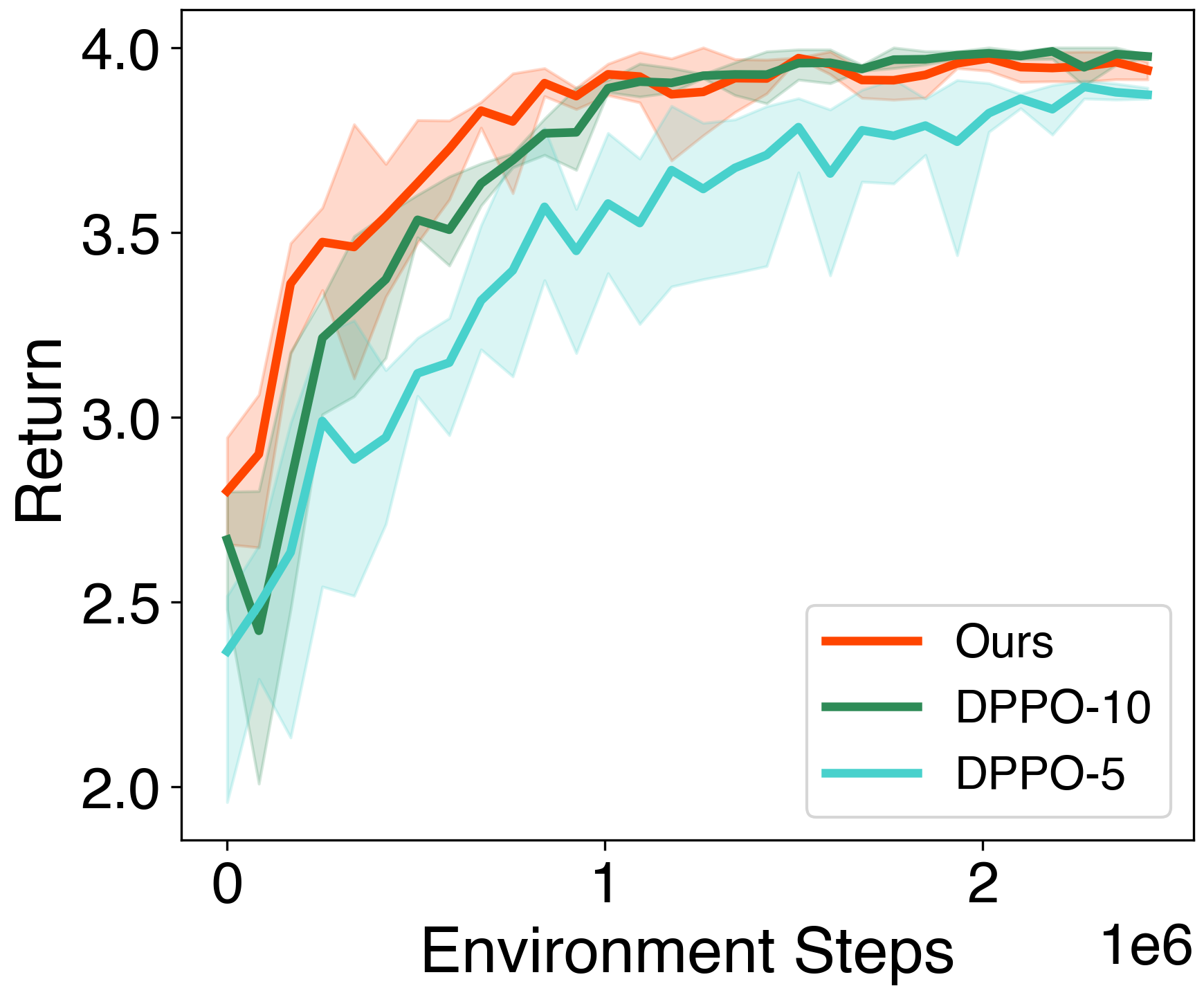}%
    \label{fig:train-kitchen-complete} 
  }
  \subfloat[Kitchen-mixed-v0\\ - Success Rate]{%
    \includegraphics[width=0.245\textwidth]{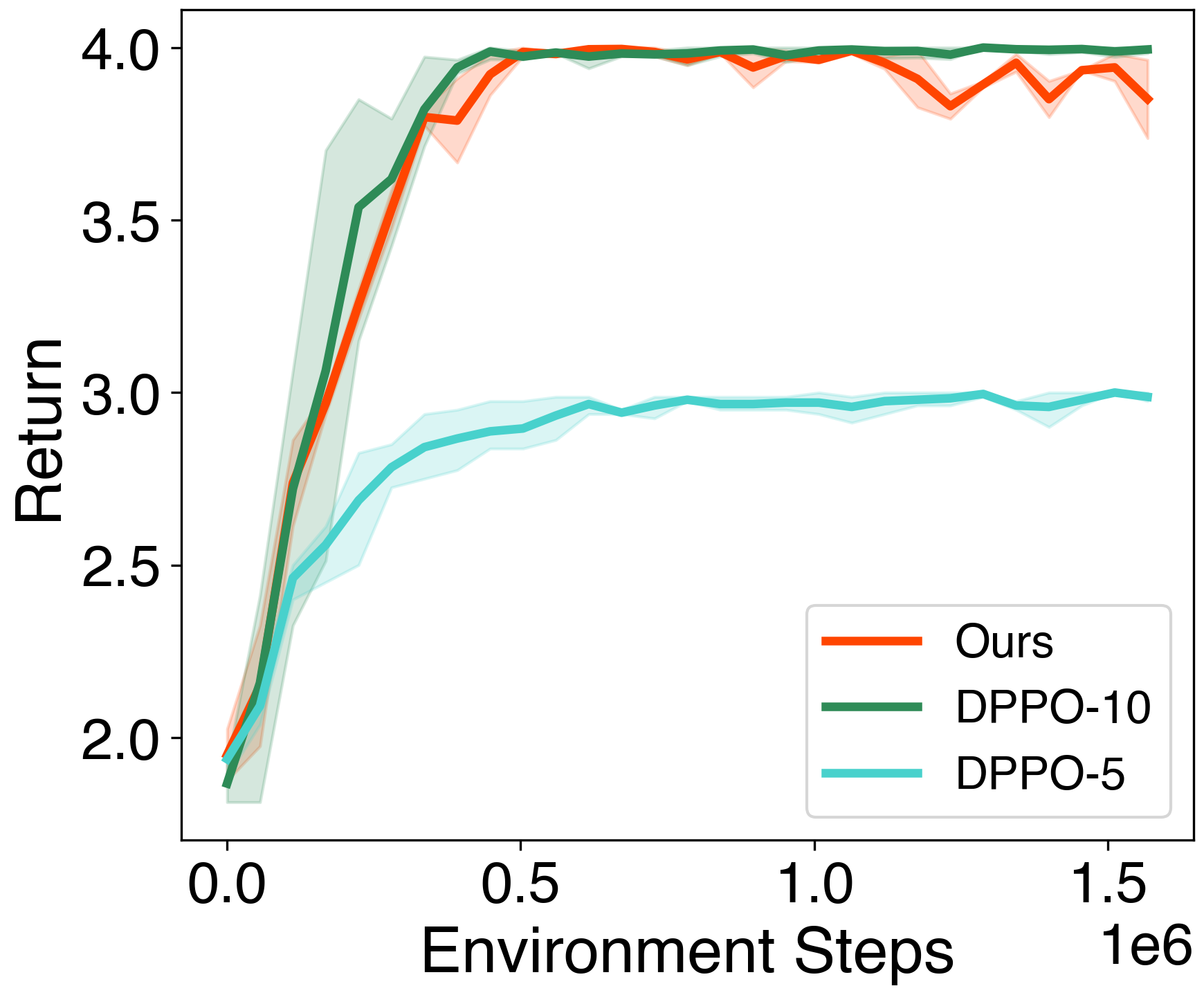}%
    \label{fig:train-kitchen-mixed} 
  }
  \caption{Training curves of eight tasks. Each curve is averaged over 5 random seeds.}
  \label{fig:curves}
\end{figure}

We present training curves in \Cref{fig:curves}, and summarize quantitative results in \Cref{tab:full_res}. Across eight tasks, D3P achieves an average success rate of 0.917, nearly matching the 0.918 of the 10-step DPPO baseline. Meanwhile, D3P achieves a 2.2$\times$ mean speed-up. We calculate this per-task acceleration ratio, $r_{acc}$, using \Cref{eq:acc_ratio}.

\begin{equation}
    r_{acc} = \frac{\mathbb{E}_{\text{all episodes}}\left[\sum_{t=0}^T \text{stp}_{\text{DPPO}, t}\right]}{\mathbb{E}_{\text{all episodes}}\left[\sum_{t=0}^T \text{stp}_{\text{D3P}, t}\right]}
    \label{eq:acc_ratio}
\end{equation}

\begin{table}[!hbp]
\vspace{2mm}
\setlength{\tabcolsep}{4pt}
\centering
\scalebox{0.76}{
\begin{tabular}{c|ccc|ccc|ccc|cccc}
    \toprule
     \multirow{2}{*}{Task} & \multicolumn{3}{c|}{DPPO} & \multicolumn{3}{c|}{Falcon} & \multicolumn{3}{c|}{CP} & \multicolumn{4}{c}{D3P (Ours)} \\
    ~ & NFE & SR  & Return & NFE & SR & Return & NFE & SR & Return & NFE & SR & Return & Acc. Ratio\\
    \midrule
    \multirow{4}{*}{\makecell{Lift \\ (State)}} 
    & 10.0 & 1.00$\pm$0.00 & 160.1$\pm$4.4 & 10.0$\pm$0.0 & 0.96$\pm$0.00 & 154.6$\pm$4.2 & 5.00 & 0.89$\pm$0.01 & 84.9$\pm$2.4 & \multirow{4}{*}{3.90$\pm$0.12} & \multirow{4}{*}{1.00$\pm$0.01} & \multirow{4}{*}{182.4$\pm$23.2} & \multirow{4}{*}{2.56} \\
    ~ & 4.00 & 1.00$\pm$0.00 & 126.7$\pm$2.7 & 8.25$\pm$0.25 & 0.96$\pm$0.01 & 158.4$\pm$3.5 & 3.00 & 0.95$\pm$0.01 & 89.3$\pm$2.8 & ~ & ~ & ~ & ~ \\
    ~ & 3.00 & 0.99$\pm$0.01 & 120.1$\pm$2.7 & 6.41$\pm$0.21 & 0.94$\pm$0.01 & 145.9$\pm$5.9 & 1.00 & 1.00$\pm$0.00 & 143.3$\pm$2.4 & ~ & ~ & ~ & ~ \\
    ~ & ~ & ~ & ~ & 3.60$\pm$0.44 & 0.76$\pm$0.03 & 121.8$\pm$9.2 & ~ & ~ & ~ & ~ & ~ & ~ & ~ \\
    \midrule
    \multirow{4}{*}{\makecell{Can \\ (State)}} & 10.0 & 0.98$\pm$0.00 & 204.4$\pm$3.4 & 10.0$\pm$0.0 & 0.97$\pm$0.01 & 194.0$\pm$5.8 & 5.00 & 0.98$\pm$0.01 & 186.8$\pm$3.7 &\multirow{4}{*}{3.71$\pm$0.37} & \multirow{4}{*}{0.98$\pm$0.02} & \multirow{4}{*}{201.1$\pm$8.0} & \multirow{4}{*}{2.70} \\
    ~ & 4.00 & 0.94$\pm$0.01 & 187.9$\pm$3.5 & 7.87$\pm$0.12 & 0.90$\pm$0.02 & 183.3$\pm$5.8 & 3.00 & 0.94$\pm$0.01 & 162.6$\pm$3.3 & ~ & ~ & ~ \\
    ~ & 3.00 & 0.89$\pm$0.01 & 173.3$\pm$3.7 & 5.22$\pm$0.35 & 0.80$\pm$0.03 & 117.1$\pm$6.4 & 1.00 & 0.94$\pm$0.00 & 167.7$\pm$3.0 & ~ & ~ & ~ \\
    ~ & ~ & ~ & ~ & 4.23$\pm$0.28 & 0.67$\pm$0.03 & 96.3$\pm$8.1 & ~ & ~ & ~ & ~ & ~ & ~ \\
    \midrule
    \multirow{4}{*}{\makecell{Square \\ (State)}} & 10.0 & 0.99$\pm$0.00 & 303.1$\pm$2.2 & 10.0$\pm$0.0 & 0.98$\pm$0.00 & 300.4$\pm$6.3 & 5.0 & 0.88$\pm$0.01 & 244.3$\pm$8.4 & \multirow{4}{*}{4.05$\pm$0.20} & \multirow{4}{*}{0.99$\pm$0.01} & \multirow{4}{*}{308.7$\pm$5.5} & \multirow{4}{*}{2.47} \\
    ~ & 5.00 & 0.95$\pm$0.03 & 285.9$\pm$10.0 & 5.41$\pm$0.22 & 0.94$\pm$0.04 & 256.7$\pm$7.3 & 3.0 & 0.88$\pm$0.02 & 244.6$\pm$2.6 & ~ & ~ & ~ & ~ \\
    ~ & 4.00 & 0.90$\pm$0.04 & 265.8$\pm$12.3 & 4.41$\pm$0.30 & 0.80$\pm$0.04 & 207.4$\pm$8.8 & 1.0 & 0.86$\pm$0.01 & 234.0$\pm$2.4 & ~ & ~ & ~ & ~ \\
    ~ & 3.00 & 0.88$\pm$0.04 & 221.2$\pm$14.9 & 3.03$\pm$0.34 & 0.74$\pm$0.06 & 164.1$\pm$8.6 & ~ & ~ & ~ & ~ & ~ & ~ & ~ \\
    \midrule
    \multirow{3}{*}{\makecell{Transport \\ (State)}} & 10.0 & 0.80$\pm$0.05 & 323.6$\pm$24.3 & 10.0$\pm$0.0 & 0.76$\pm$0.04 & 321.4$\pm$28.4 & 5.00 & 0.71$\pm$0.06 & 243.5$\pm$24.8 & \multirow{3}{*}{6.75$\pm$0.14} & \multirow{3}{*}{0.80$\pm$0.07} & \multirow{3}{*}{341.0$\pm$30.8} & \multirow{3}{*}{1.48} \\
    ~ & 7.00 & 0.58$\pm$0.04 & 233.4$\pm$25.0 & 7.75$\pm$0.50 & 0.76$\pm$0.04 & 300.4$\pm$31.1 & 3.00 & 0.74$\pm$0.03 & 270.0$\pm$19.9 & ~ & ~ & ~ & ~ \\
    ~ & 5.00 & 0.58$\pm$0.05 & 221.2$\pm$22.3 & 6.69$\pm$0.89 & 0.56$\pm$0.06 & 239.6$\pm$31.5 & 1.00 & 0.64$\pm$0.03 & 235.0$\pm$22.8 & ~ & ~ & ~ & ~ \\
    \midrule
    \multirow{4}{*}{\makecell{Square \\ (Image)}} & 10.0 & 0.86$\pm$0.05 & 240.5$\pm$14.2 & 10.0$\pm$0.0 & 0.87$\pm$0.04 & 228.2$\pm$14.9 & 5.00 & 0.63$\pm$0.06 & 153.1$\pm$7.4 & \multirow{4}{*}{3.95$\pm$0.74} & \multirow{4}{*}{0.89$\pm$0.03} & \multirow{4}{*}{232.0$\pm$3.2} & \multirow{4}{*}{2.53} \\
    ~ & 5.00 & 0.78$\pm$0.07 & 187.0$\pm$20.9 & 6.89$\pm$0.39 & 0.87$\pm$0.06 & 212.8$\pm$21.4 & 3.00 & 0.65$\pm$0.03 & 161.8$\pm$9.6 & ~ & ~ & ~ & ~ \\
    ~ & ~ & ~ & ~ & 4.82$\pm$0.51 & 0.70$\pm$0.06 & 162.8$\pm$28.1 & 1.00 & 0.60$\pm$0.03 & 150.3$\pm$4.8 & ~ & ~ & ~ & ~ \\
    ~ & ~ & ~ & ~ & 3.39$\pm$0.36 & 0.64$\pm$0.07 & 128.9$\pm$8.4 & ~ & ~ & ~ & ~ & ~ & ~ & ~ \\
    \midrule
    \multirow{3}{*}{\makecell{Transport \\ (Image)}} & 10.0 & 0.75$\pm$0.02 & 289.9$\pm$13.5 & 10.0$\pm$0.0 & 0.75$\pm$0.02 & 284.6$\pm$9.7 & 5.00 & 0.39$\pm$0.07 & 104.8$\pm$8.4 & \multirow{3}{*}{6.89$\pm$0.52} & \multirow{3}{*}{0.75$\pm$0.03} & \multirow{3}{*}{285.0$\pm$9.8} & \multirow{3}{*}{1.45} \\
    ~ & 7.00 & 0.63$\pm$0.03 & 240.9$\pm$12.9 & 8.12$\pm$0.16 & 0.74$\pm$0.04 & 271.3$\pm$10.2 & 3.00 & 0.36$\pm$0.08 & 98.4$\pm$9.3 & ~ & ~ & ~ & ~ \\
    ~ & ~ & ~ & ~ & 6.55$\pm$0.71 & 0.59$\pm$0.09 & 213.9$\pm$16.1 & 1.00 & 0.45$\pm$0.04 & 119.3$\pm$7.8 & ~ & ~ & ~ & ~ \\
    \midrule
    \multirow{4}{*}{\makecell{Kitchen \\ (Complete)}} & 10.0 & 0.98$\pm$0.00 & 3.96$\pm$0.02 & 10.0$\pm$0.0 & 0.99$\pm$0.01 & 3.98$\pm$0.03 & 5.0 & 0.76$\pm$0.03 & 3.48$\pm$0.09 & \multirow{4}{*}{4.15$\pm$0.37} & \multirow{4}{*}{0.98$\pm$0.01} & \multirow{4}{*}{3.98$\pm$0.02} & \multirow{4}{*}{2.41} \\
    ~ & 5.00 & 0.94$\pm$0.01 & 3.91$\pm$0.03 & 7.92$\pm$0.18 & 0.98$\pm$0.01 & 3.95$\pm$0.07 & 3.0 & 0.67$\pm$0.01 & 3.17$\pm$0.02 & ~ & ~ & ~ & ~ \\
    ~ & 4.00 & 0.90$\pm$0.01 & 3.83$\pm$0.03 & 5.52$\pm$0.39 & 0.82$\pm$0.05 & 3.56$\pm$0.08 & 1.0 & 0.65$\pm$0.01 & 2.99$\pm$0.02 & ~ & ~ & ~ & ~ \\
    ~ & ~ & ~ & ~ & 4.54$\pm$0.31 & 0.77$\pm$0.05 & 3.50$\pm$0.07 & ~ & ~ & ~ & ~ & ~ & ~ & ~ \\
    \midrule
    \multirow{4}{*}{\makecell{Kitchen \\ (Mixed)}} & 10.0 & 0.99$\pm$0.00 & 3.99$\pm$0.00 & 10.0$\pm$0.0 & 0.99$\pm$0.01 & 3.98$\pm$0.05 & 5.0 & 0.85$\pm$0.06 & 3.72$\pm$0.10 & \multirow{4}{*}{4.78$\pm$0.02} & \multirow{4}{*}{0.97$\pm$0.02} & \multirow{4}{*}{3.93$\pm$0.05} & \multirow{4}{*}{2.09} \\
    ~ & 5.00 & 0.0$\pm$0.0 & 2.99$\pm$0.01 & 7.79$\pm$0.15 & 0.99$\pm$0.01 & 3.88$\pm$0.04 & 3.0 & 0.81$\pm$0.06 & 3.57$\pm$0.09 & ~ & ~ & ~ & ~ \\
    ~ & 4.00 & 0.0$\pm$0.0 & 2.99$\pm$0.00 & 5.76$\pm$0.21 & 0.90$\pm$0.05 & 3.44$\pm$0.09 & 1.0 & 0.90$\pm$0.05 & 3.80$\pm$0.07 & ~ & ~ & ~ & ~ \\
    ~ & ~ & ~ & ~ & 4.40$\pm$0.37 & 0.70$\pm$0.07 & 3.22$\pm$0.10 & ~ & ~ & ~ & ~ & ~ & ~ & ~ \\
    \bottomrule
\end{tabular}}

\caption{Detailed results of all methods, fomulated as mean $\pm$ std. All results are averaged over 5 seeds.}
\label{tab:full_res}
\setlength{\tabcolsep}{6pt}
\end{table}

\newpage
\section{Real-world Deployment}

We deploy D3P on a Franka robot arm to complete the \texttt{Square} task. 

\subsubsection{Setup}
\begin{figure}[!hbp]
    \centering
    \subfloat[\texttt{Sqaure} Task Setting]{
        \includegraphics[width = 0.4\textwidth]{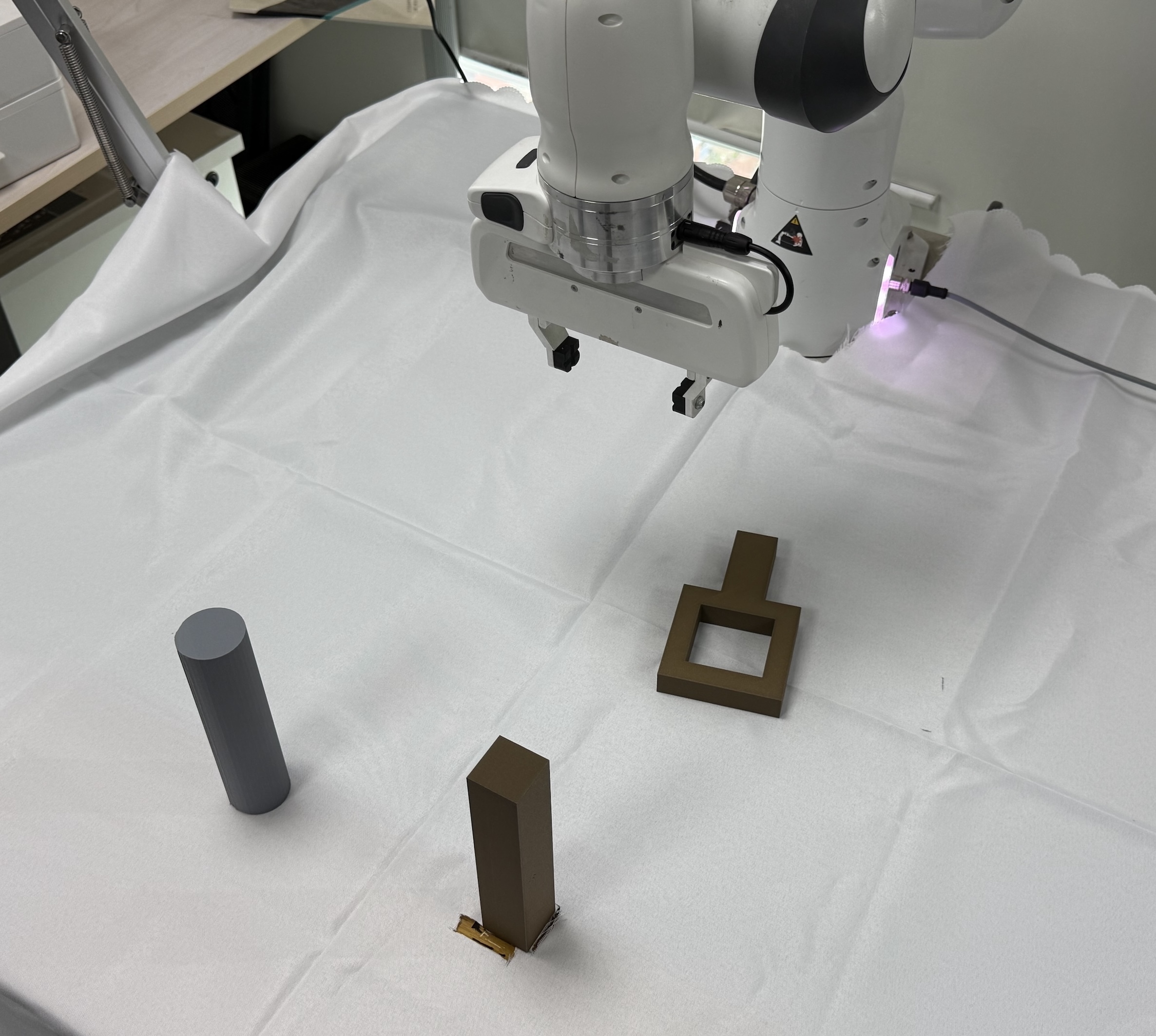}
        \label{fig:real-env}
    }
    \subfloat[We use 3D print to manufacture the objects]{
        \includegraphics[width = 0.4\textwidth]{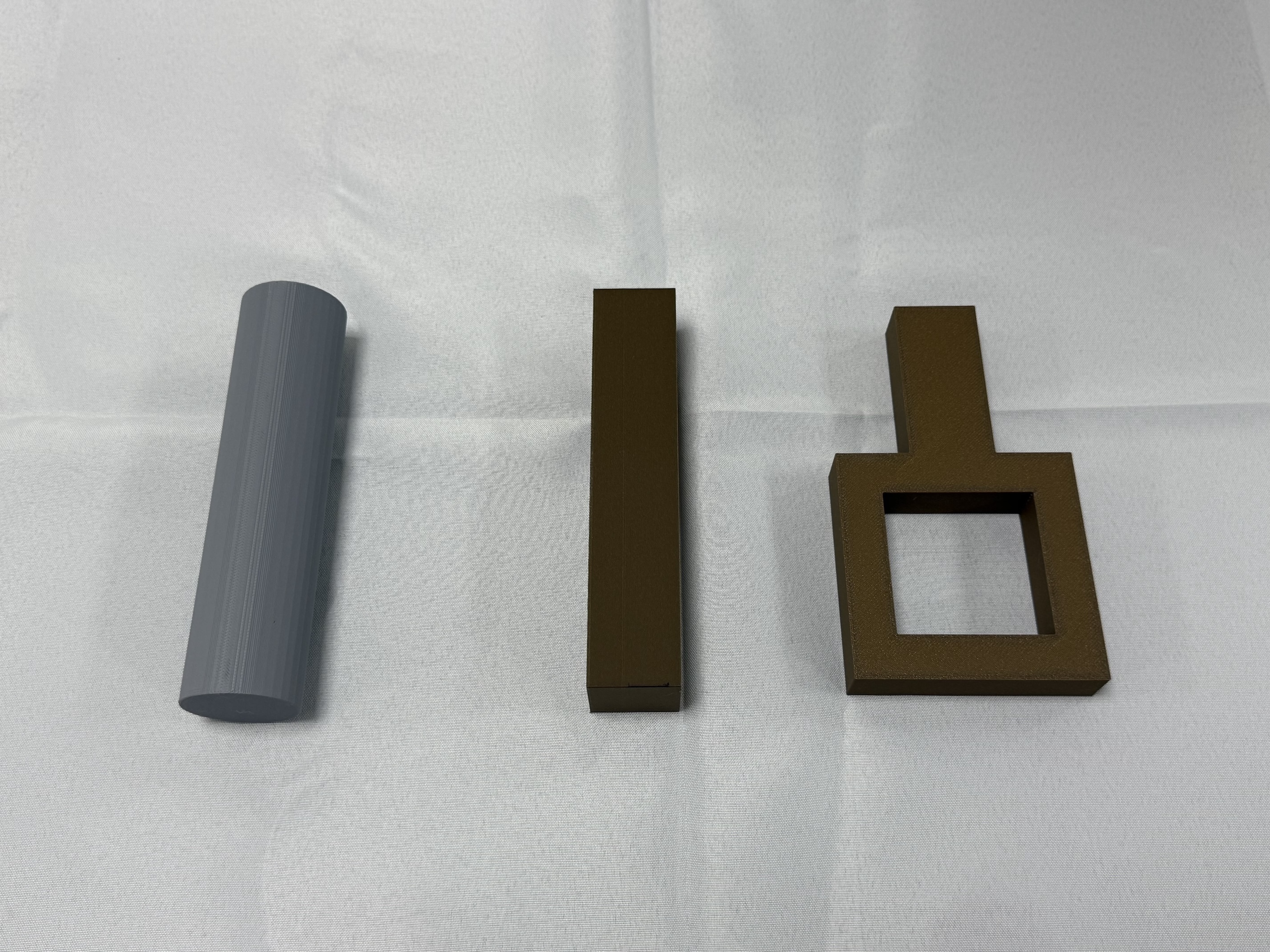}
        \label{fig:real-obj}
    }
    \caption{The \texttt{Square} task involves: (1) grasping the handle of the square, (2) inserting the square onto corresponding peg.}
    \label{fig:real}
\end{figure}

We define the task environment as illustrated in \Cref{fig:real-env}. In the \texttt{Square} task, the agent is required to successfully grasp the handle of the square nut and precisely mate it with the corresponding square peg. We use a Franka robot arm for executing, and a consumer-grade desktop (i7-12900K CPU, RTX 2080 GPU) for computing. We fabricated the physical objects required for the task using 3D printing, ensuring that their color, shape, and dimensions are consistent with simulated objects, as shown in \Cref{fig:real-obj}. For this task, D3P utilizes images and joint positions as input. The RGB images are captured by an Intel RealSense D435i camera positioned 0.92m in front of and 0.53m above the robot's base, providing a 45 degree downward viewing angle. \Cref{fig:real-progress} shows the progress of this task, and we include a video in the supplementary materials that showcases the agent executing this task.

\begin{figure}[!hbp]
    \centering
    \subfloat{
        \includegraphics[width = 0.16\textwidth]{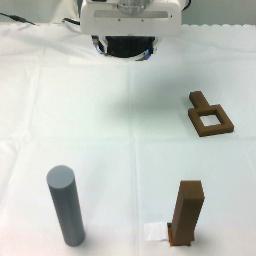}
    }
    \subfloat{
        \includegraphics[width = 0.16\textwidth]{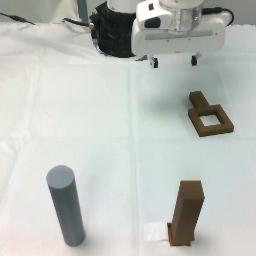}
    }
    \subfloat{
        \includegraphics[width = 0.16\textwidth]{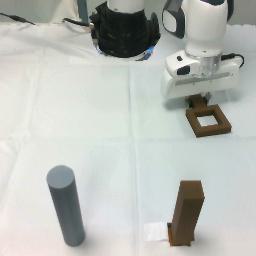}
    }
    \subfloat{
        \includegraphics[width = 0.16\textwidth]{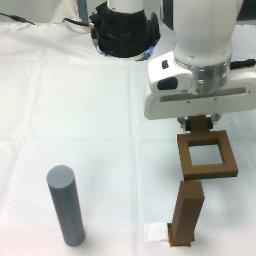}
    }
    \subfloat{
        \includegraphics[width = 0.16\textwidth]{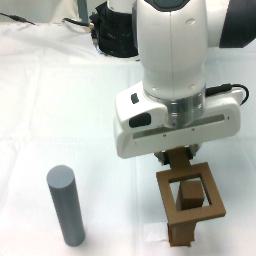}
    }
    \subfloat{
        \includegraphics[width = 0.16\textwidth]{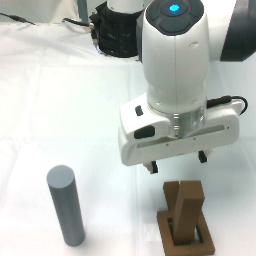}
    }
    \caption{Demonstration of the \texttt{Square} task}
    \label{fig:real-progress}
\end{figure}

\subsubsection{Sim-to-real transfer}
\begin{figure}
    \centering
    \subfloat[Real-world Image]{
        \includegraphics[width = 0.28\textwidth]{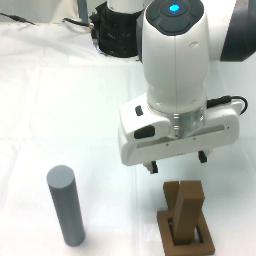}
    }
    \subfloat[Converted Image]{
        \includegraphics[width = 0.28\textwidth]{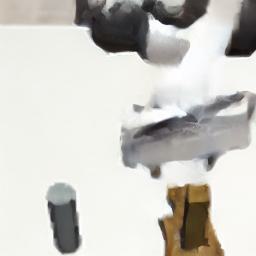}
    }
    \caption{We use LDM to convert real-world images into the simulation style}
    \label{fig:convert}
\end{figure}

To achieve sim-to-real transfer without real-world data, we employ a latent diffusion model (LDM)~\cite{rombach2021highresolution} to convert real-world images into a style that approximates the simulated domain, as illustrated in \Cref{fig:convert}. The process begins with a pretrained Variational Autoencoder (VAE) that encodes the input image into a compact latent feature. Subsequently, a diffusion model operates on this latent feature to perform the translation. Finally, the VAE's decoder reconstructs the processed feature into the converted output image.

\begin{figure}
    \centering
    \subfloat[Canonical Image]{
        \includegraphics[width = 0.28\textwidth]{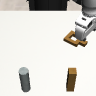}
    }
    \subfloat[Randomized Image]{
        \includegraphics[width = 0.28\textwidth]{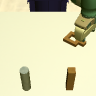}
    }
    \caption{Images pair for training LDM}
    \label{fig:ldmdata}
\end{figure}

We train the LDM with paires of images collected in simulation by domain radomization. As illustrated in \Cref{fig:ldmdata}, we employ domain randomization in the simulation by varying the object materials and lighting conditions, from which we collect paired images of the canonical and randomized scenes.

Finally, to bridge the gap in camera parameters between the simulated and real-world environments, we introduce a curriculum learning strategy during RL training. This curriculum systematically adjusts the camera's intrinsic and extrinsic parameters, progressively transitioning them from the initial simulation settings to the physical hardware setting.

\end{document}